\documentclass[11pt]{article}
\usepackage[numbers]{natbib}
\usepackage{macros/packages}
\usepackage{macros/editing-macros}
\usepackage{macros/formatting}
\usepackage{macros/statistics-macros}
\usepackage[utf8]{inputenc} 
\usepackage[T1]{fontenc}    
\usepackage{hyperref}       
\usepackage{url}            
\usepackage{booktabs}       
\usepackage{amsfonts}       
\usepackage{nicefrac}       
\usepackage{microtype}      
\usepackage{xcolor}         
\usepackage{wrapfig}
\usepackage{subcaption}
\usepackage{tikz}
\usepackage{algorithm}
\usepackage{algorithmic}

\usetikzlibrary{%
  calc,
  fit,
  arrows,
  arrows.meta,
  positioning,
  decorations.pathreplacing,
  decorations.shapes,
}
\theoremstyle{definition}
\newtheorem{myexample}{Example}

\usepackage{ulem}

\renewcommand{\emph}[1]{\textit{#1}}

\begin{document}

\abovedisplayskip=8pt plus0pt minus3pt
\belowdisplayskip=8pt plus0pt minus3pt


\begin{center}
  {\huge Architectural and Inferential Inductive Biases  For \vspace{0.3cm}\\ Exchangeable Sequence Modeling} \\
  \vspace{.5cm} {\large Daksh Mittal$^{*}$ ~~~~ Ang Li$^{*}$ ~~~~ Tzu-Ching Yen$^{*}$  ~~~~ Daniel Guetta ~~~~  Hongseok Namkoong   } \\
  \vspace{.2cm}
  {\large Decision, Risk, and Operations Division, Columbia Business School} \\
     \vspace{.2cm}
   \texttt{\{dm3766, al4263, ty2531, crg2133, hongseok.namkoong\}@columbia.edu}
  \vspace{.2cm}
  \texttt{}
\end{center}


\begin{abstract}%
  Autoregressive models have emerged as a powerful framework for modeling exchangeable sequences---i.i.d. observations when conditioned on some latent factor---enabling direct modeling of uncertainty from missing data (rather than a latent). 
Motivated by the critical role posterior inference plays as a subroutine in decision-making (e.g., active learning, bandits), we study the inferential and architectural inductive biases that are most effective  for exchangeable sequence modeling.
For the inference stage, we highlight a fundamental limitation of the prevalent single-step generation approach:  inability to distinguish between epistemic and aleatoric uncertainty.
Instead, a long line of works in Bayesian statistics advocates for multi-step autoregressive generation; we demonstrate this "correct approach" enables superior uncertainty quantification that translates into better performance on downstream decision-making tasks.
This naturally leads to the next question: which architectures are best suited for multi-step inference? We identify a subtle yet important gap between recently proposed Transformer architectures for exchangeable sequences~\cite{MullerHoArGrHu22, NguyenGr22, YeNa24}, and prove that they in fact cannot guarantee exchangeability despite introducing significant computational overhead.  We illustrate our findings using controlled synthetic settings, demonstrating how custom architectures can significantly underperform standard causal masks, underscoring the need for new architectural innovations.


\end{abstract}
\def\thefootnote{*}\footnotetext{Equal contribution}
\section{Introduction}
\label{section:introduction}

Intelligent agents must be able to articulate their own uncertainty about the underlying environment, and sharpen its beliefs as it gathers more information. We consider an sequence of observations $Y_{1:\infty}$ gathered from an unseen environment $\theta$, e.g., noisy answers in a math quiz, generated by a student's current proficiency level $\theta$.
When marginalizing over the latent  $\theta$, the joint distribution of the sequence $Y_{1:\infty}$ is permutation invariant, and thus \emph{exchangeable sequences} form a basic unit of study in uncertainty quantification of latents that govern data generation.

Autoregressive sequence modeling has recently gained significant attention as a powerful approach for modeling exchangeable sequences $Y_{1:\infty}$~\cite{MullerHoArGrHu22, NguyenGr22, ZhangCaNaRu24, YeNa24, Hollmannetal25}. Unlike conventional Bayesian modeling—which requires specifying a prior over an unobserved latent variable $\theta$ and a likelihood for the observed data, often a challenging task—autoregressive sequence modeling builds on De Finneti's predictive view of Bayesian inference~\cite{  CifarelliRe96, DeFinetti33, DeFinetti37,
DeFinetti17}. This view directly models the observables $Y_{1:\infty}$: for an exchangeable sequence, epistemic uncertainty in the latent variable $\theta$ stems from the unobserved future data~\cite{BertiPrRi04, BertiDrLePrRi22, FongHoWa23}.  Viewing future observations as the sole source of epistemic uncertainty in $\theta$ for exchangeable sequences ~\cite{BertiPrRi04, BertiDrLePrRi22, FongHoWa23}, autoregressive sequence modeling enables direct prediction of the observables $Y_{1:\infty}$, offering a conceptually elegant and practical alternative to conventional Bayesian approaches.

\begin{figure}[t]
\includegraphics[width=\linewidth,height=8cm]{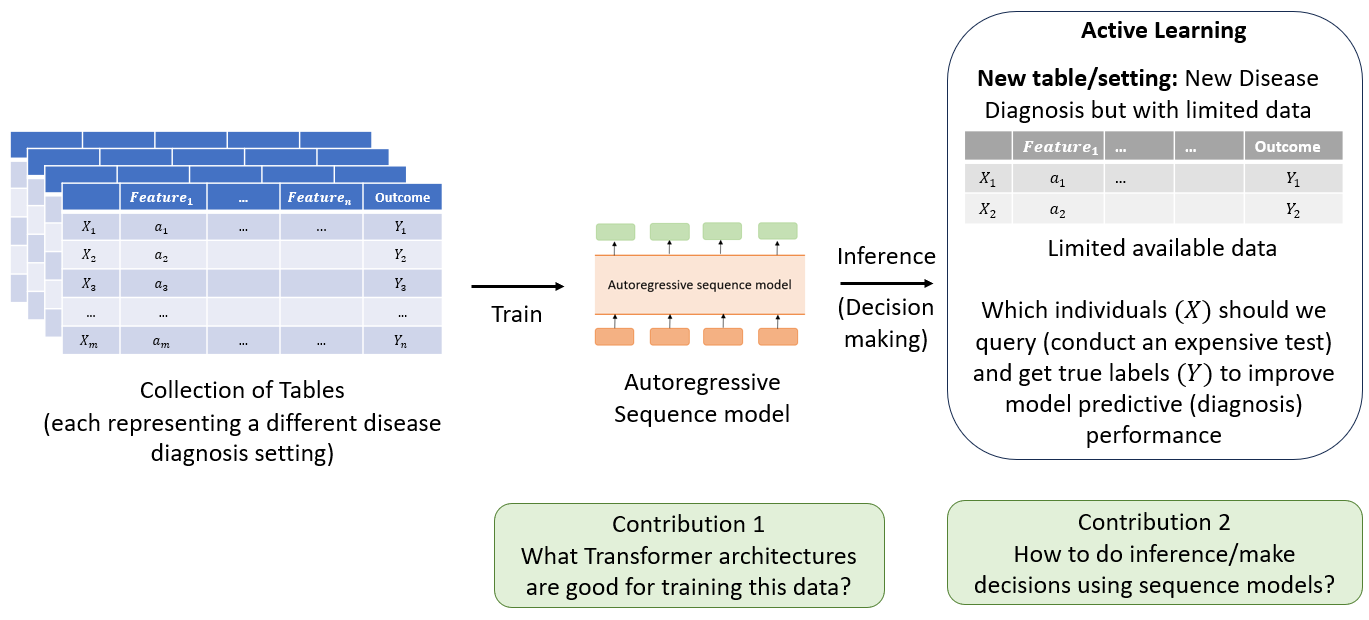}
    \caption{\textbf{Meta-learned sequence models can be used for decision making}}
    \label{fig:motivating_example}
\end{figure}

Transformers have emerged as the dominant architecture for autoregressive sequence modeling~\cite{Hollmannetal25, MullerHoArGrHu22, HanYoArPf24, HegselmannBuLaAgJiSo23, GardnerPeSc24, YanZhXuZhChSuWuCh24}, owing to their remarkable performance in natural language and vision applications~\cite{BrownEtAl20, Dosovitskiyetal21}. 
As Transformers are increasingly employed to meta-learn probabilistic models for large-scale tabular datasets~\cite{zhaoBiCh23, Hollmannetal25}, they offer a unique opportunity to move beyond traditional prediction tasks or merely replicating supervised algorithms—an area that has been the primary focus so far.  Instead, following De Finetti’s perspective, when meta-trained on tabular datasets, these models can effectively quantify epistemic uncertainty, which powers decision-making and active exploration across diverse domains, including recommendation systems, adaptive experimentation, and active learning~\cite{ZhangCaNaRu24}. For instance, we can train sequence models on a collection of tables, each representing a different disease diagnosis setting---akin to applying meta-learning in the context of  disease diagnosis. These models can then be leveraged to actively gather additional data in a previously unseen disease diagnosis setting to enhance model's predictive performance (see Figure \ref{fig:motivating_example} for illustratation).

\begin{figure}[t]
    \centering
  \includegraphics[width=0.8\linewidth,height=9.5cm]{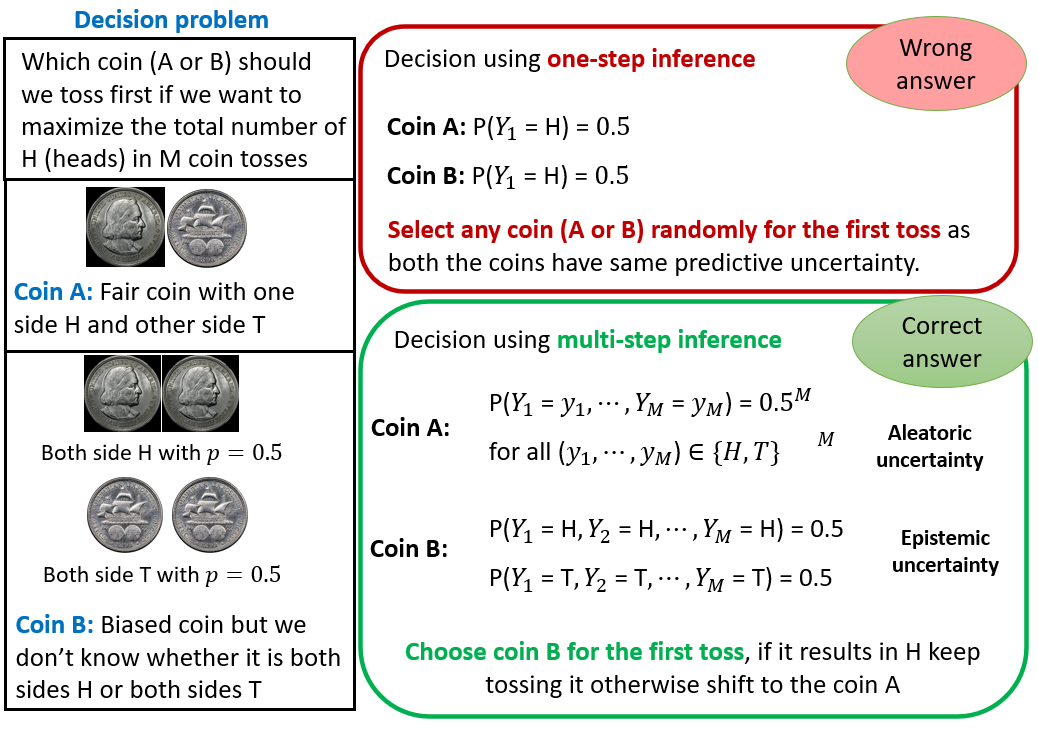}
    \caption{\textbf{[Illustration of Multi-step inference v/s One-step inference in decision making]} Coins A and B are considered identical by single-step inference because both have the same level of predictive uncertainty in their rewards. However, multi-step inference highlights a key difference: for Coin B, the uncertainty can be reduced (epistemic uncertainty) by performing a single toss, whereas for Coin A, all the uncertainty is irreducible (aleatoric) and arises from the inherent randomness of a fair coin toss. Consequently, multi-step inference prioritizes tossing Coin B first to reduce epistemic uncertainty.}
    \label{fig:one-step-multi-step}
\end{figure}

However, using Transformers to model exchangeable sequences for decision-making presents its own challenges. Since the existing literature has primarily focused on traditional prediction tasks or the replication of supervised algorithms rather than decision-making, it has overlooked the perspective that epistemic (reducible) uncertainty stems from missing data. Accurately distinguishing between epistemic (reducible) and aleatoric (irreducible) uncertainty is crucial for decision-oriented applications. A significant limitation in the current literature is its predominant focus on one-step predictive uncertainty~\cite{MullerHoArGrHu22, NguyenGr22} or one-step predictions~\cite{HegselmannBuLaAgJiSo23}, which fails to differentiate between epistemic and aleatoric uncertainty. 
In contrast, multi-step inference provides a more robust framework for differentiating between epistemic and aleatoric uncertainty~\cite{WenOsQiLuIbDwAsVa22, OsbandWeAsDwLuIbLaHaDoRo22} (see Figure \ref{fig:one-step-multi-step} to come for an illustrative example).

Through both empirical analysis and theoretical examination, we demonstrate  impact of one-step and multi-step inference on uncertainty quantification  (Theorem \ref{thm:uq_one_step_suffers_non_contextual}, Figure \ref{fig:one_step_multi_step_all_three}) and downstream decision-making tasks, particularly in multi-armed bandit and active learning settings. Our findings show that multi-step inference consistently outperforms one-step inference in these tasks (see Section \ref{sec:inference-inductive-bias}). These results align with those of \citet{ZhangCaNaRu24}, who also proposed multi-step inference to enhance active exploration via Thompson sampling.

\begin{figure}[t]
    \centering
    \includegraphics[width=0.9\linewidth]{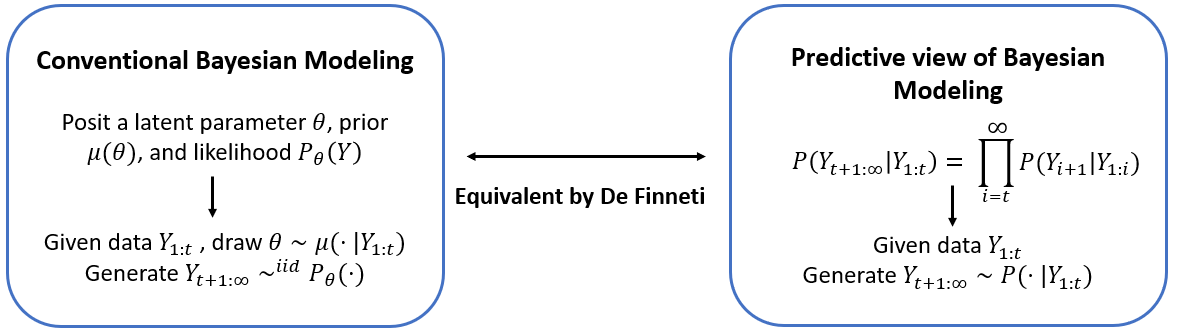}
    \caption{Equivalence between conventional Bayesian modeling and predictive view of Bayesian inference is established by De Finneti (valid only under infinite exchangeability of $Y_{1:\infty} $). It establishes that epistemic uncertainty in latent parameter $\theta \sim \mu(\cdot|y_{1:t})$ is equivalent to predictive uncertainty of future observations $Y_{t+1:\infty} \sim P(\cdot|y_{1:t})$. }
    \label{fig:predictive_view}
\end{figure}
This brings us to the next key question: What kind of architectures should we use for modeling exchangeable sequences, particularly when performing multi-step inference? Architectural choice is a crucial source of inductive bias, and existing approaches have attempted to incorporate exchangeability as an inductive bias through specialized masking strategies~\cite{MullerHoArGrHu22, NguyenGr22, YeNa24}.
However, a critical gap in the literature remains: these prior works conflate exchangeability with the invariance properties enforced by their masking schemes that don't necessarily guarantee valid probabilistic inference.

A key contribution of our work is in articulating this gap. We introduce conditional permutation invariance (Property \ref{property:conditional-permutation-invariant}) as a way to formalize the invariance enforced by these masking strategies. Correcting previous works~\cite{MullerHoArGrHu22, NguyenGr22, YeNa24} that implicitly assume their architectures achieve exchangeability, we show that enforcing conditional permutation invariance alone is insufficient to guarantee full exchangeability. Specifically, existing masking-based approaches do not ensure the conditionally identically distributed (c.i.d.) property (Property \ref{property:c-i-d}), which is essential for the validity of probabilistic inference in exchangeable sequence models. As a result, despite their intended design, such models may still violate exchangeability when used in practice (see Section~\ref{sec:Architecture_inductive_biases}). By clearly distinguishing between exchangeability and the weaker notion of conditional permutation invariance, our work not only clarifies existing misconceptions in the literature but also establishes a more rigorous foundation for developing exchangeable sequence models.

Moreover, a significant drawback of this masking scheme, as discussed in Section \ref{sec:Architecture_inductive_biases}, is that it introduces computational overhead without yielding any tangible improvements in model performance. To evaluate its impact, we empirically assess the effectiveness of enforcing conditional-permutation invariance (Property \ref{property:conditional-permutation-invariant}) within the model architecture in controlled synthetic settings, comparing it to standard causal masking, which does not enforce permutation invariance. Surprisingly, our results show that enforcing Property \ref{property:conditional-permutation-invariant} not only fails to provide any performance benefits but actually performs worse than causal masking (Figure \ref{fig:comparing_two_architectures_main_diagram}).
These findings underscore the need for new research directions to develop more effective inductive biases for Transformers in exchangeable sequence modeling.

The primary objective of this paper is to investigate the conceptual foundations of the predictive approach to Bayesian inference, with a particular emphasis on adapting autoregressive sequence modeling of exchangeable sequences for decision-making tasks. Our key contributions include:

\begin{itemize}
    \item \textbf{Meta-learned sequence models for decision-making:} We demonstrate that sequence models, trained on tabular data, can effectively support decision-making tasks such as active learning and multi-armed bandits, extending their utility beyond conventional supervised learning.
    \item \textbf{Inferential inductive biases:} We show that one-step inference is insufficient for decision-making as it fails to distinguish between epistemic and aleatoric uncertainty. In contrast, multi-step inference provides better uncertainty quantification, leading to improved decision-making.
    \item \textbf{Architectural inductive biases:} We show that, contrary to prior assumptions, existing masking-based architectures do not guarantee full exchangeability. We formally introduce conditional permutation invariance, distinguishing it from true exchangeability, and demonstrate why enforcing it does not necessarily achieve full exchangeability. Moreover, in practice, these architectures underperform compared to standard causal models.
\end{itemize}

We critically examine the role and objectives of the above inductive biases from both theoretical and empirical perspectives, identifying their strengths and limitations. Through this exploration, we resolve ambiguities in the existing literature and contribute to a deeper understanding of exchangeable sequence modeling. We hope that the insights presented in this work will help refine and advance the application of Transformers trained on tabular data for decision-making and active exploration.

\section{Conceptual background}
\label{section:background}

In this section, we first review \textit{De Finetti's predictive view of Bayesian inference}~\cite{  CifarelliRe96, DeFinetti33, DeFinetti37,
DeFinetti17} and  how autoregressive sequence modeling of exchangeable sequences  can power it.  For most of our discussion, we focus on the setting where observations are given by $\Obs_{1:\infty}$. However, this framework can be easily extended to contextual settings where observations are $(X_{1:\infty}, \Obs_{1:\infty})$, with $X$ serving as the context (Section \ref{sec:contextual_setting}). We start by reviewing the conventional Bayesian modeling paradigm, where the modeler posits a latent parameter $\theta$, along with a prior $\mu(\theta)$, and  likelihood $\P(\Obs_{1:\infty}\mid \theta) \equiv P_\theta(\Obs_{1:\infty})$. The joint probability of  observations $Y_{1:\infty}$ is then expressed as 

\[  \P(\Obs_{1:\infty} = \obs_{1:\infty})
  = \int \prod_{t=1}^\infty P_\theta(\Obs_t = \obs_t) \mu(d \theta).\]

Given any observable data $Y_{1:t}$ the posterior over latent parameter is expressed as  $\mu(\cdot|Y_{1:t})$. Note that $\mu(\cdot|Y_{1:t})$ represents the epistemic (reducible) uncertainty, which gets resolved as more data is collected while the likelihood $\P(Y|\theta)$ represents the aleatoric uncertainty and comes due to inherent randomness in the data.


\paragraph{Predictive view of uncertainty} 
Instead of positing explicit priors and likelihoods over a proposed latent parameter space, we
consider a different probabilistic modeling approach where we
directly model the observable $\Obs_{1:\infty}$ without explicitly  relying on any latent parameter. This view heavily relies on the \textbf{infinite exchangeability} of the sequence $Y_{1:\infty}$, defined as:
\begin{align}
\P(\Obs_{1},\cdots, \Obs_n) =\P(\Obs_{\pi(1)},\cdots, \Obs_{\pi(n)})
    \label{eq:exchangeability}
\end{align}
for any $n$ and permutation $\pi$.
De Finetti's theorem states that if an infinite sequence is exchangeable then the sequence can be represented as a mixture of i.i.d. (independent and identically distributed) random variables. 
\begin{theorem}[De Finetti's theorem]
  If a sequence $\Obs_{1:\infty}$ is infintely exchangeable then there exists a latent parameter $\theta$ and a unique measure $\mu(\cdot)$ over $\theta$, such that
  \begin{equation}
    \label{eqn:definetti-equivalence}
    \P(\Obs_{1:\infty} = \obs_{1:\infty}) = \int \prod_{t=1}^\infty \P(\Obs_t =
    \obs_t|\theta) \mu(d \theta).
  \end{equation}
  \label{thm:definneti}
\end{theorem}  
  \noindent  In addition to justifying conventional Bayesian modeling,  De Finetti’s theorem also establishes that for infinitely exchangeable sequence $Y_{1:\infty}$, the epistemic uncertainty in the latent parameter $\theta$ in Theorem \ref{thm:definneti} arises solely from the unobserved  $\Obs_{1:\infty}$~\cite{BertiDrPrRi21, FortiniLaRe00, FortiniPe14, HahnMaWa18}. In other words, epistemic uncertainty in $\theta$ is the same as predictive uncertainty in $Y_{1:\infty}$. \citet{DeFinetti37, HewittSa55} in fact show that the latent
parameter $\theta$ in Equation (\ref{eqn:definetti-equivalence}) is entirely a function of the $Y_{1:\infty}$, that is, $\theta = f(Y_{1:\infty})$. 

To illustrate, consider coin $B$ from Figure \ref{fig:one-step-multi-step} and suppose it is tossed repeatedly. Let  $Y_t^{(B)}$ denote the outcome of  $t$-th  toss of coin $B$ where $Y_t^{(B)}=1$ for heads, and $Y_t^{(B)}=0$ for tails. The sequence  $\{Y_1^{(B)}, \cdots, Y_\infty^{(B)}\}$ is exchangeable, since its probability distribution  depends only on the outcomes of the tosses, not their \emph{order}. In this case, De Finetti's Theorem holds trivially, and the latent parameter  $\theta$ represents the probability of obtaining `heads' when the coin $B$ is flipped. By the Law of Large Numbers, this parameter $\theta$ can be estimated as $\theta = \lim_{t\to \infty}\frac{1}{t}\sum_{i=1}^t Y_i^{(B)}$. The same reasoning applies for coin $A$ as well.

By abusing notation and writing $\P(\cdot|Y_{1:\infty}) \equiv \P(\cdot|\theta)$ where $\theta = f(Y_{1:\infty})$, we can interpret $\P(\cdot|Y_{1:\infty})$ itself as the latent parameter $\theta$. Therefore, given some observation $Y_{1:t}$,  generating $Y_{t+1:\infty} \sim \P(\cdot|Y_{1:t})$ is equivalent to sampling $\theta \sim \mu(\cdot|Y_{1:t})$.   This indicates we can do equivalent Bayesian inference~\cite{FongHoWa23} using $\P(\cdot|Y_{1:t})$.

\paragraph{Differentiating epistemic and aleatoric uncertainty} Since epistemic uncertainty is equivalent to the predictive uncertainty of future observations, it can be reduced with additional observations. In contrast, \emph{aleatoric} uncertainty refers to uncertainty that remains irreducible, even with more observations. For example, in the coin toss scenario, the uncertainty in Coin $A$ is aleatoric, arising from inherent randomness of a fair coin toss. No matter how many times coin $A$ is flipped, there will always be uncertainty about the outcome of the next coin $A$ toss.  Coin $B$, on the other hand, has epistemic uncertainty which can be eliminated by flipping the coin once.

To quantify epistemic uncertainty and distinguish from aleatoric uncertainty, we must  autoregressively generate $Y_{1:\infty} \sim \P(\cdot)$.  As illustrated in Figure \ref{fig:one-step-multi-step}, it is important to consider the full sequence of coin tosses  rather than just a single toss. If we only examine one coin toss, both the coins exhibit the same predictive uncertainty, making it impossible to differentiate between the two coins. Formally, since $ \P(Y_{t+1}|Y_{1:t}) = \int \P(Y_{t+1}|\theta) \mu(\theta|Y_{1:t})$, one-step inference fails to distinguish between epistemic and aleatoric uncertainty.   We discuss this in more detail in Section \ref{sec:inference-inductive-bias}.

 
\paragraph{Learning a probabilistic model from data}
We now shift our focus to directly learning $\P(\Obs_{1:\infty} = \obs_{1:\infty})$. 
We represent the transformer-based autoregressive sequence model,  parametrized by $\phi$, as $\what{P}_\phi$. Rewrite $\what{P}_\phi(\hat{\Obs}_{t+1:\infty}=\obs_{t+1:\infty})$ as an autoregressive product: $\prod_{i=t}^\infty \what{P}_\phi(\hat{\Obs}_{i+1}= \obs_{i+1}\mid \hat{\Obs}_{1:i}=\obs_{1:i}),$
where $\what{P}(\hat{\Obs}_{i+1}\mid \hat{\Obs}_{1:i})$ represents a one-step predictive distribution, often referred to as posterior predictive in conventional Bayesian terminology.
The transformer is trained to minimize the KL-divergence between the true data generating process $\P$ and the model $\what{P}_\phi$
$$\dkl{\P}{\what{P}_\phi} \propto -\E_{y_{i+1}\sim  \P(\cdot|y_{1:i})} \left[\log\left( \what{P}_\phi \left(\hat{Y}_{i+1}=y_{i+1}|\hat{Y}_{1:i}=y_{1:i}\right)\right)\right].$$ 

Given
a dataset of sequences $\{y_{1:T}^j: 1\leq j \leq N\}$ generated from the true data generating process $\P(\cdot)$, we train the autoregressive sequence model (transformer) on this data using the following objective: 
\begin{align}
  \label{eqn:training_transformer}
    -\frac{1}{N} \sum_{j=1}^N \sum_{i=0}^{T-1}
    \log \what{P}_\phi\left( \hat{\Obs}_{i+1}^j = \obs_{i+1}^j \mid  \hat{\Obs}_{1:i}^j = \obs_{1:i}^j  \right).
\end{align}
For  simplicity, denote  $\what{P}_\phi(\hat{Y}_{i+1} =  y|\hat{Y}_{1:i})$ as $\what{P}_\phi^{(i+1)}(y)$.
This training procedure enables the model to learn the single-step predictive distributions that collectively define the full sequence likelihood. Once trained, given any observed data $y_{1:t}$, the transformer sequence model can generate future samples autoregressively: $\hat{Y}_{t+1:\infty}\sim \what{P}_\phi(\cdot |y_{1:t})$.

Next, we examine how these trained sequence models can be applied to decision-making, highlighting the limitations of one-step inference and how multi-step inference effectively overcomes them.


\section{Inferential inductive biases: one-step vs. multi-step inference for decision making}
\label{sec:inference-inductive-bias}


To make informed and effective decisions, it is crucial to distinguish between epistemic and aleatoric uncertainty. 
As mentioned earlier in Section \ref{section:background} and elaborated on later, one-step inference does not adequately differentiate between these two types of uncertainty. However, much of the existing literature on autoregressive sequence modeling has focused primarily on one-step inference or prediction. For example, recent work by \citet{HegselmannBuLaAgJiSo23} empirically demonstrates the scalability of sequence models for training on tabular data, but their study emphasizes one-step prediction. Similarly, while \citet{MullerHoArGrHu22, NguyenGr22, Hollmannetal25} discuss predictive uncertainty, their focus remains exclusively on one-step predictive uncertainty.
 In contrast, multi-step inference (predictive distributions) provide a more effective means of quantifying epistemic uncertainty, which is crucial for sequential decision making applications~\cite{WenOsQiLuIbDwAsVa22, OsbandWeAsDwLuIbLaHaDoRo22}. (This distinction was informally illustrated in Figure \ref{fig:one-step-multi-step}.)
 
Similarly, in \textit{active learning} (Figure \ref{fig:motivating_example}), where one-step inference fails to differentiate between individuals whose disease outcome is inherently uncertain and those whose uncertainty stems from a lack of sufficient ground truth data.
To further illustrate the difference between one-step and multi-step inference, we present a more formal example:


\begin{myexample} 
Consider a setting where the observation $Y \sim P_\theta = N(\theta, \tau^2)$  follows a normal distribution with mean $\theta \sim \mu = N(a,\sigma^2)$.  Given observations $y_{1:t}$, the posterior distribution $\mu(\theta|y_{1:t}) = N(a_t, \sigma_t^2)$ represents the epistemic (reducible) uncertainty, while $Y \sim N(\theta, \tau^2)$ captures the aleatoric (irreducible) uncertainty  caused by inherent randomness in the data.

\noindent\textit{One-step inference}  corresponds to $ \P(Y_{t+1}|y_{1:t}) $, which is equivalent to  $ \int P_\theta(Y_{t+1}) \mu(\theta|y_{1:t}) d\theta  = N(a_t, \sigma_t^2 +\tau^2) $. In this case, the predictive uncertainty for $Y_{t+1}$ (characterized by $\sqrt{\sigma_t^2+\tau^2}$), combines both  epistemic ($\sigma_t^2$)  and aleatoric ($\tau^2$) uncertainties. Importantly, this approach does not allow us to separate the two components.

\noindent\textit{Multi step inference,} on the other hand involves $\P(Y_{t+1:\infty}|Y_{1:t})$, which is (by De Finneti's) equivalent to $ \int \prod_{i=t+1}^\infty P_\theta(Y_i) \mu(\theta|y_{1:t})d\theta$. Furthermore, as $T \to \infty$, we have  $\lim_{T \to \infty}\frac{1}{T}\sum_{i=t+1}^\infty Y_t =\theta$ where $\theta \sim \mu(\cdot|y_{1:t})  = N(a_t, \sigma_t^2) $. Hence, the variance of the long-term average $\frac{1}{T}\sum_{i=t+1}^\infty Y_t$  is approximately $\sigma_t^2$ - isolates the epistemic uncertainty. Therefore, multi-step inference enables the differentiation between epistemic ($\sigma_t^2$)  and aleatoric ($\tau^2$) uncertainties. 
\end{myexample}
Recall that in  De Finneti's theorem, we recover $\theta \sim \mu(\cdot|y_{1:t})$ only when we generate $Y_{t+1:\infty}\sim \P(\cdot|Y_{1:t})$. Therefore, valid Bayesian inference necessitates multi-step inference: generating $Y_{t+1:\infty}\sim \P(\cdot|y_{1:t})$ allows averaging out any idiosyncratic noise. In sequence modeling using transformers, the sequences   $\hat{Y}_{t+1:\infty} \sim \what{P}_\phi(\cdot|y_{1:t})$ are generated autoregressively.

\subsection{Theoretical characterization of impact on decision-making}

We emphasize the significance of the gap between single-step and multi-step inference by examining its impact on \emph{decision-making}. Recall Figure \ref{fig:one-step-multi-step} where we needed to decide whether to flip coin A or B  first to maximize the  total number of heads over $M$ coin tosses. One-step inference fails to distinguish between the two coins, leading to a suboptimal decision of selecting a coin randomly. Similarly, in \textit{active learning} setting (Figure \ref{fig:motivating_example}),  one-step inference may lead us to select individuals whose disease outcomes are inherently uncertain, whereas our primary focus should be on individuals where uncertainty arises due to insufficient ground truth data. This misallocation results in unnecessary expenses. In contrast, multi-step inference addresses this issue, enabling optimal decision-making. We now formally characterize the suboptimality in uncertainty quantification and decision-making that arises from one-step inference.


\paragraph{Characterizing information loss in one-step inference v/s multi-step inference:} We  analyze how our uncertainty quantification---and inference in general---deteriorates when relying solely on one-step inference instead of multi-step inference. Let $y_{t+1:T} \sim \P(\cdot|y_{1:t})$ be some data from the data generating process. Further, let $\what{P}^M_\phi(y_{t+1:T})   \equiv \prod_{i=t+1}^{T}\what{P}_\phi(\hat{Y}_{t+1}=y_{i}|y_{1:t})$ be the one-step (marginal) inference model and let $\what{P}_\phi^J(y_{t+1:T}) \equiv   \prod_{i=t+1}^{T}\what{P}_\phi(\hat{Y}_i=y_i|y_{1:i-1})$ be the multi-step inference model, where we generate $\hat{Y}_i$ autoregressively.  Then the expected difference between the log-likelihood of the  multi-step inference model and single-step inference model is given by the following result.
\begin{theorem}  Assuming $\what{P}_\phi = \P$, then the difference  $\E \left(\log\left[ \what{P}^J_\phi(y_{t+1:T})\right]-\log\left[ \what{P}^M_\phi(y_{t+1:T})  \right] \right)$ is equal to 
    \begin{align}
     \sum_{i=t+1}^T I(y_i; y_{t+1:i-1}|y_{1:t}) 
        \label{expression:one-multi-step-compare_non_contextual}
    \end{align}   
where $I(A;B|C)$ is the mutual information between $A$ and $B$ conditional on $C$ and expectation is w.r.t. $y_{1:T}\sim \P(\cdot)$.   
\label{thm:uq_one_step_suffers_non_contextual}
\end{theorem}

This demonstrates that relying solely on one-step inference results in the loss of mutual information among \( y_{t+1:T} \). Consequently, single-step inference is inherently less effective than multi-step inference.  We further analyze expression (\ref{expression:one-multi-step-compare_non_contextual}) in a specific setting (Example \ref{example:non_contextual_example}) and show that as \( \sigma^2 \) (epistemic uncertainty) increases, the performance gap between one-step and multi-step inference widens, with one-step inference becoming increasingly suboptimal.


\begin{myexample} Assuming
$Y=\theta+\epsilon$ 
where 
$\epsilon \sim N(0,\tau^2)$ and $\theta 
\sim N(\mu, \sigma^2)$. Let  
$y_{t+1:T}\sim \P(\cdot|y_{1:t})$ and $(\sigma')^2 = \left(\frac{1}{\sigma^2} +\frac{t}{\tau^2}\right)^{-1}$, then  
expression (\ref{expression:one-multi-step-compare_non_contextual})
is equal to
$\frac{1}{2} \log \left(\left(1+\frac{\sigma'^2}{\tau^2}\right)^{T-t} \right)- \frac{1}{2} \log\left(1+(T-t)\frac{\sigma'^2}{\tau^2} \right)
$.
\label{example:non_contextual_example}
\end{myexample}

\textbf{Characterizing impact on decision making:}
We now examine how one-step inference affects downstream decision-making tasks.  Specifically, we consider a one-armed bandit problem where the reward of the first arm is given by  $Y^{(1)} \sim N(\theta, \tau^2)$ with $\theta \sim N(\mu,\sigma^2)$. While the second arm has a constant reward  $Y^{(2)} = 0$. It is well known that Thompson sampling suffers $O(\sqrt{T})$ Bayesian regret in multi armed bandits~\cite{RussoVaKaOsWe18}. We now consider the performance of Thompson sampling implemented based on just the one-step inference  or one-step inference (predictive undertainty), with proof in Section \ref{sec:proofs}. 

\begin{theorem} There exists  one-armed bandit scenarios in which Thompson sampling incurs \( O(T) \) Bayesian regret if  it relies solely on one-step predictions from autoregressive sequence models.
\label{thm:bandit_theorem}
\end{theorem}


\textbf{Generalizing to the contextual setting:}
We extend our analysis to the contextual setting in Section \ref{sec:proofs}, where the context \( X \sim_{\text{i.i.d.}} P_X \). Additionally, we characterize the information loss associated with one-step inference in Bayesian linear regression and Gaussian processes in the same section.


\subsection{Empirical investigation}
\label{sec:inference_experiments}

\begin{figure}[t]
\centering
\begin{minipage}[b]{0.32\textwidth}
\centering \includegraphics[width=0.9\textwidth,height=4cm]{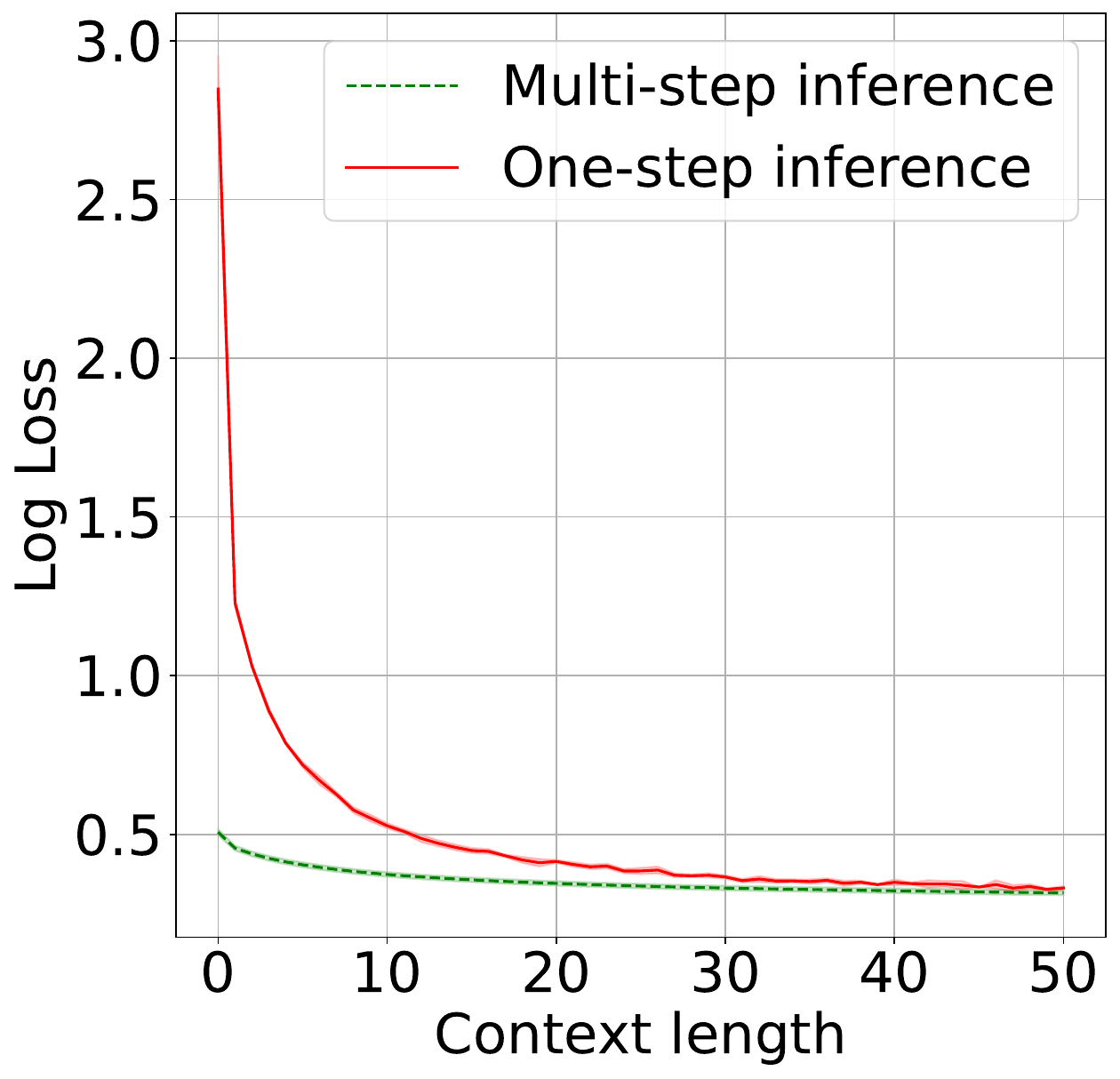}
  \caption*{\textbf{(a)} Uncertainty Quantification}
\end{minipage}
\hfill
\begin{minipage}[b]{0.32\textwidth}
\centering \includegraphics[width=0.9\textwidth,height=4cm] {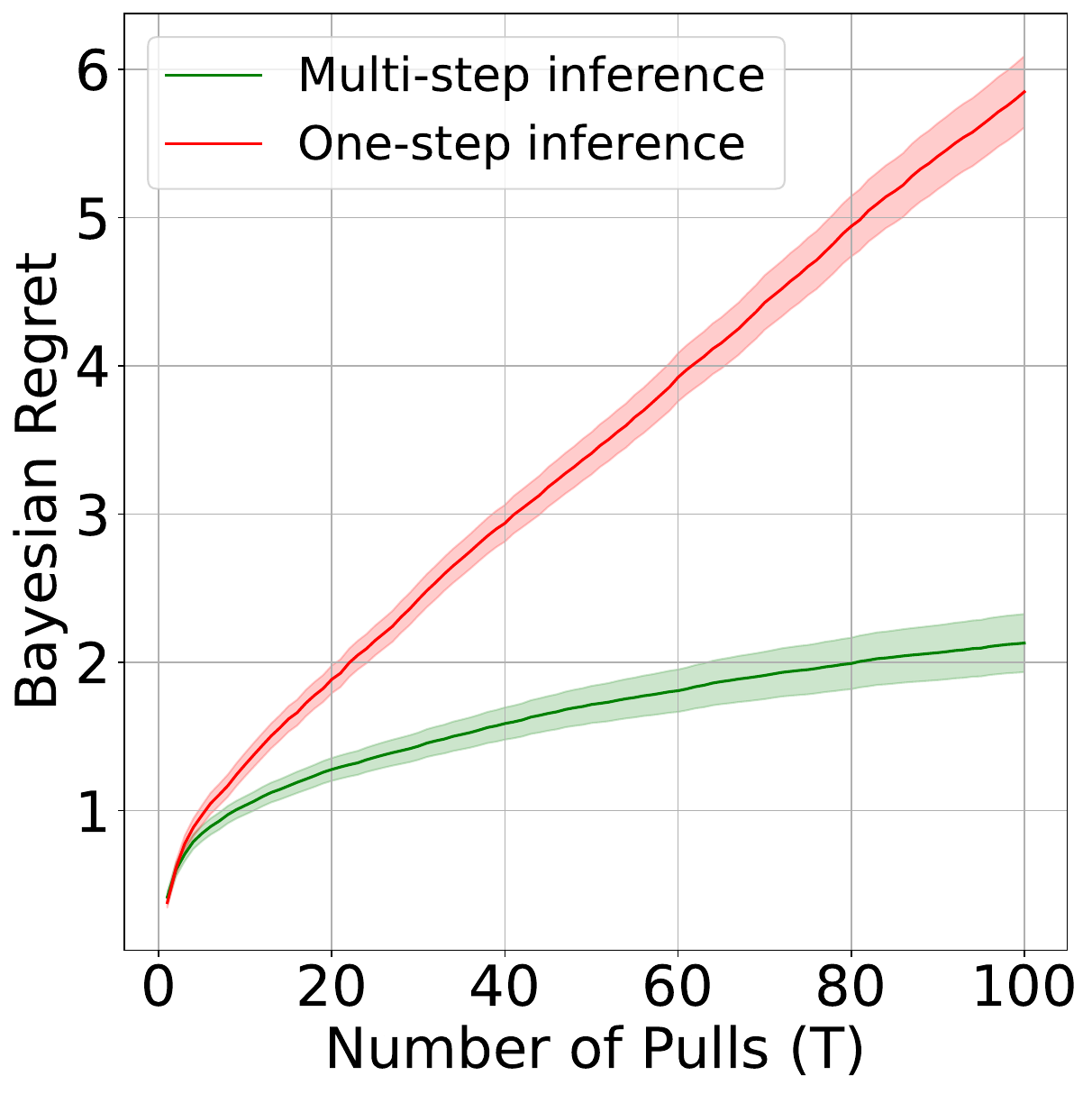}
  \caption*{\textbf{(b)} Multi-armed Bandits}
\end{minipage}
\hfill
\begin{minipage}[b]{0.32\textwidth}
\centering 
\includegraphics[width=0.9\textwidth,height=4cm]
{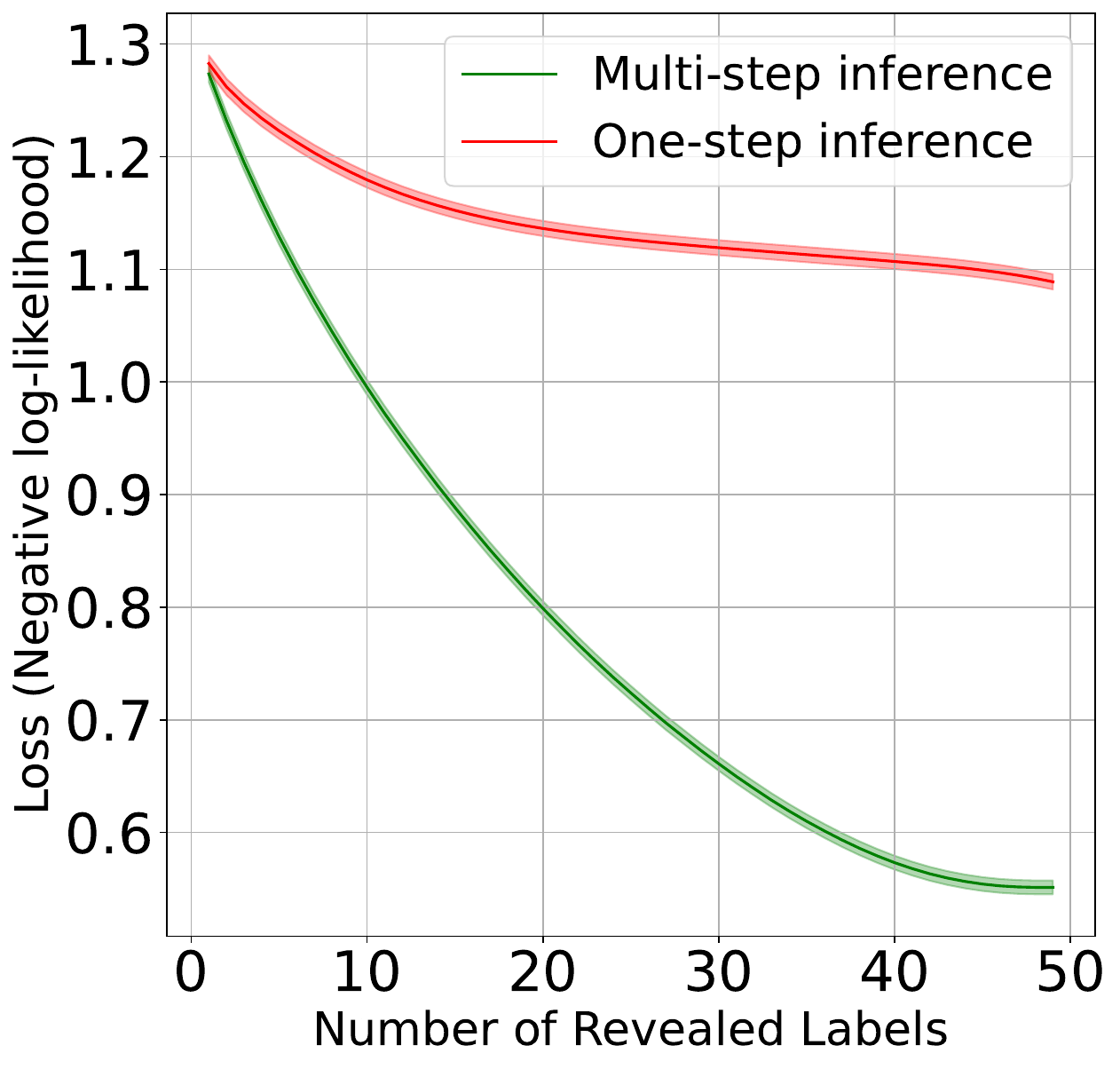}
  \caption*{\textbf{(c)} Active Learning}
\end{minipage}
\caption{\textbf{Comparing one-step inference and multi-step inference (lower is better):} \textbf{(a)} Uncertainty Quantification: Comparing multi-step log-loss for one-step and multi-step inference  (Train horizon: 100, Target Length: $50$) \textbf{(b)} Multi-armed Bandits: Comparing Bayesian regret  under Thompson sampling algorithm using one-step and multi-step inference. \textbf{(c)} Active Learning: Comparing log-loss under uncertainty sampling algorithm using one-step and multi-step inference.}
\label{fig:one_step_multi_step_all_three}
\end{figure}

In this section, we empirically evaluate the impact of single-step and multi-step inference on uncertainty quantification, as well as on downstream optimization tasks such as multi-armed bandits and active learning. We find that multi-step inference significantly outperforms one-step inference, being up to $60\%$ more efficient in bandit settings and requiring up to $10$ times less data in active learning to achieve the same predictive performance.\footnote{Our code repository is available at: \\  \url{https://github.com/namkoong-lab/Inductive-biases-exchangeable-sequence}.}

\subsubsection{Uncertainty quantification}
We evaluate one-step and multi-step inference in a controlled synthetic setting by generating a synthetic dataset using a Gaussian Process (GP). Specifically, we employ a GP with an RBF kernel: $f \sim \mathcal{GP}(m,\mc{K}) $, where $\mc{K}(X,X') = \sigma_f^2 \exp\left(-{||X-X'||_2^2}/{2\loss^2}\right)$. Additionally,   Gaussian   noise $N(0,\sigma^2)$ is added to the outputs.  The input $X$ is drawn i.i.d. from {$P_X$}. To compare the performance of the two inference strategies, we use the multi-step log-loss metric. Further details on the metrics and experimental setup can be found in Section  \ref{section:experiments-details}. 
Figure \ref{fig:one_step_multi_step_all_three}\textcolor{blue}{(a)} illustrates the comparison of multi-step log-loss performance between one-step and multi-step inference. Consistent with our theoretical results (Theorems \ref{thm:uq_one_step_suffers_non_contextual} and \ref{thm:uq_one_step_suffers}), the results demonstrate that one-step inference performs worse than multi-step inference.
\subsubsection{Multi-armed Bandits}

\textbf{Problem:} We consider a two-armed Bayesian bandit  setting with arms $\{C, D\}$ and $T$ rounds during which the arms are pulled.  In each round $t$, based on the information collected so far, $\{(A_i, Y_i^{A_i}): 1\leq i\leq t-1\}$, an arm $A_t$ is selected, and a reward $Y_t^{A_t}$ is observed. For each arm $a \in \{C,D\}$, the rewards are distributed as  $Y^{(a)}_{1:T} \simiid  N\left(\theta^{(a)}, \left(\tau^{(a)}\right)^2\right)$, where the mean reward  $\theta^{(a)}$ follows a prior distribution $\theta^{(a)} \sim N\left(\mu^{(a)}, \left(\sigma^{(a)}\right)^2\right)$. The objective is to determine which arm $A_t \in \{C, D\}$ to pull in each round  $t$ in order to  minimize the Bayesian regret, defined as:
\begin{align}
 \E \left(\max_{a\in\{C,D\}} \{T\theta^{(a)}\} - \sum_{i=1}^T Y_t^{(A_t)}\right),
 \label{eq:bayesian_regret}
\end{align}

where expectation is taken over the randomness in the rewards ($Y$), the actions/policy ($A_t$), and the means ($\theta$).


\textbf{Training and Evaluation:} We train two transformers, one for each arm (C and D). 
To compare the two inference strategies, we first sample $\theta^* \sim \mu$ for each arm.  We then implement Thompson Sampling (using the respective inference strategy) with the trained transformers to acquire rewards over a horizon $T$. The regret is evaluated as $\left[\max_{a\in\{C,D\}} \{T\theta^{*(a)}\} - \sum_{i=1}^T Y_t^{(A_t)}\right]$.   Finally, we average the regret over $1000$ different samples of $\theta^*$, with each run consisting of $T=100$ steps, to compute the Bayesian regret (\ref{eq:bayesian_regret}). Additional details about training and algorithm implementation are provided in Section \ref{section:experiments-details}.

\textbf{Results:} Our results are summarized in Figure  \ref{fig:one_step_multi_step_all_three}\textcolor{blue}{(b)}. As expected, cumulative regret increases with the number of pulls for both inference strategies. The figure also demonstrates that multi-step inference significantly outperforms one-step inference having upto $60\%$ less regret.

\subsubsection{Active Learning}
\textbf{Problem:} In active learning, the goal is to adaptively collect labels  $Y$ for inputs $X$ to maximize the performance of a model $\psi(\cdot)$. We focus on a pool-based setting, where a pool of data points $ \mathcal{X}^{pool}$ is given,  and the objective is to sequentially query labels  $Y$ for $X \in \mathcal{X}^{pool}$.  We consider a regression setting in which inputs $X\simiid P_X$, and outcomes are generated from an unknown function $f^*$, such that $ Y=f^*(X)+\epsilon_X$, where the noise $\epsilon_X \sim N(0, \tau_X^2)$ is heteroscedastic. 
Additionally, the data-generating function $f^*$ drawn from a distribution $\mu$.


\noindent\textbf{Training and Evaluation:} We consider a meta learning setup where we train the sequence model (transformer) on data $\{(X_{1:N}^{(j)},Y_{1:N}^{(j)}):j \in [1,M]\}$ generated from the original data generating process. 
To evaluate the two inference strategies, we first sample $f^* \sim \mu$ and generate a dataset $\mathcal{X}\times\mathcal{Y} \equiv \mathcal{D}$, which contains both pool dataset ($\mathcal{D}^{pool}$)  and test dataset ($\mathcal{D}^{test}$). Using the trained transformer and the respective inference strategy  for uncertainty sampling,  we sequentially select $X \in \mathcal{X}^{pool}$ for which labels are queried.  At each time step $t$, given the collected data $\mathcal{D}^t \subset \mathcal{D}^{pool}$,  we evaluate  the transformer model's performance as  $\left[- \sum_{(X,Y)\in \mathcal{D}^{test}}\what{P}_\phi(Y|X, \mathcal{D}^t)\right]$. 


\noindent\textbf{Results:} The results are summarized in Figure \ref{fig:one_step_multi_step_all_three}\textcolor{blue}{(c)}. For both inference strategies, prediction accuracy  improves, and loss decreases as more data points are acquired. However, multi-step inference significantly outperforms single-step inference, achieving the same performance level with nearly $10$ times fewer samples.

Now that we have established the importance of multi-step inference, the next key question arises: What architectures are best suited for modeling exchangeable sequences, especially when performing multi-step inference?




\section{Architectural inductive biases}
\label{sec:Architecture_inductive_biases}

\begin{figure}[t]
\centering
\centering \includegraphics[width=0.45\linewidth, height=4cm] {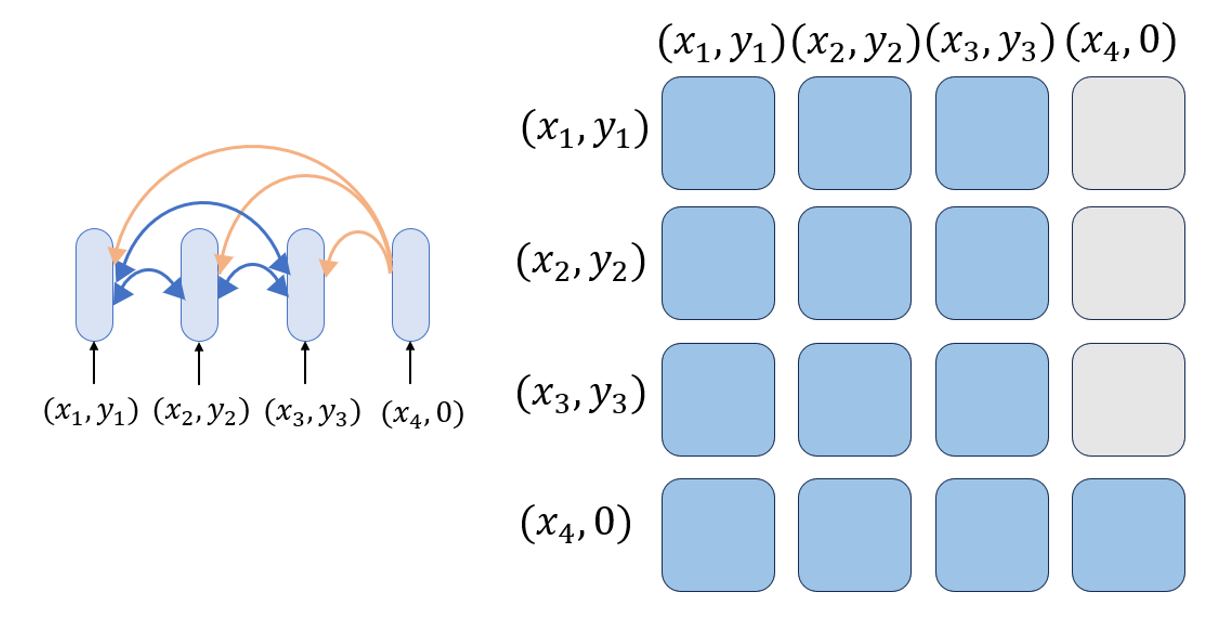}
  \caption{A representative attention mechanism and masking scheme widely used in prior literature to enforce exchangeability. However, it only ensures the \textbf{conditionally permutation-invariant} property.}
\label{fig:permutation-invariant-mask-prior}
\end{figure}

Ensuring that the sequence model $\what{P}_\phi$  is infinitely exchangeable enables robust performance and reliable statistical inference~\cite{YeNa24}. Several prior works have proposed architectural approaches to enforce exchangeability. \citet{MullerHoArGrHu22} introduced a masking scheme designed to enforce exchangeability (Figure \ref{fig:permutation-invariant-mask-prior}). In this scheme, all context points attend to one another, allowing the model to condition on the entire context set without any predefined ordering constraints. This design ensures that the model's predictions remain invariant to the order of context points, aligning with the principle that exchangeable sequences should not depend on the specific order in which observations are presented.
Figure \ref{fig:permutation-invariant-mask-prior} illustrates the masking scheme corresponding to this architecture, where $(x_1, y_1), \cdots, (x_3,y_3)$ denote the context points, and $(x_4,0)$ represents the target point for which $\hat{y}_4$ is to be predicted. 
\citet{NguyenGr22} proposed a similar architecture with modifications to improve training efficiency; however, their design contained an error where the target point did not attend to itself. This issue was later corrected by \citet{YeNa24}.

Despite these efforts, the exchangeability guarantees of these architectures remain uncertain. Previous works implicitly assume that their masking schemes enforce exchangeability but fail to provide a formal characterization of what is actually achieved. To address this gap, we introduce a precise definition that explicitly captures the invariance imposed by these architectures. Our analysis reveals that all transformer-based architectures in the literature~\cite{MullerHoArGrHu22, NguyenGr22, YeNa24} aim to enforce a \textit{“conditional permutation invariance property,”} which we formally define below.

\begin{property} 
$\what{P}_{\phi}$ is conditionally permutation invariant  if
    \begin{align}
 \what{P}_\phi(\hat{Y}_{t+1}|\hat{Y}_{\pi(1)},\cdots,\hat{Y}_{\pi(t)}) = \what{P}_\phi(\hat{Y}_{t+1}|\hat{Y}_1,\cdots,\hat{Y}_t) 
\end{align}
\label{property:conditional-permutation-invariant}
\end{property}
\noindent This property ensures that the predictive uncertainty in $y_{t+1}$ remains the same under any permutation of the context $ y_{1:t}$. We refer the reader to Section \ref{sec:contextual_setting} for the corresponding definitions in contextual settings. Additionally, in Section \ref{sec:conditionally_permutation_invariant_architecture}, we provide a detailed discussion of the transformer architecture with the conditional permutation invariance property, including its efficient training procedure and the computational requirements for inference.

However, while conditional permutation invariance is a necessary characteristic of exchangeability, it is not sufficient. Enforcing this property alone does not guarantee full exchangeability—a critical oversight in prior work. For example, all exchangeable sequence models must also satisfy another crucial property called the conditionally identically distributed (c.i.d.) property, also known as the martingale property:
\begin{property} Recalling $\what{P}^t_{\phi}(y) \triangleq \what{P}_\phi\left(\hat{Y}_{t}=y \mid \hat{Y}_{1:t-1} \right)$, $\what{P}_{\phi}$ is conditionally Identically Distributed (c.i.d.) if
   \begin{align}
       \E \left[ \what{P}^{t+1}_\phi(y) \mid \hat{Y}_{1:t-1}\right]  =  \what{P}^t_{\phi}(y),
   \end{align}
\label{property:c-i-d}
\end{property}
\noindent This property ensures that the expected predictive distribution at time $t+1$, given past observations ($\hat{Y}_{1:t-1}$), is consistent with the predictive distribution at time $t$.   The importance of c.i.d. property in exchangeable sequence models is what powers their ability to quantify epistemic uncertainty, as emphasized by previous work in Bayesian statistics~\cite{BertiPrRi04, BertiDrLePrRi22, FongHoWa23, FalckWaHo24}.

However, an autoregressive sequence model that satisfies Property \ref{property:conditional-permutation-invariant} does not necessarily satisfy Property \ref{property:c-i-d}. To illustrate this, consider the following example of an autoregressive sequence model:

\begin{myexample}
    Suppose $\what{P}^1_\phi(0):=\what{P}_\phi(0) = \frac{1}{3}$. Further, conditioned on first observation, assume that $\what{P}^1_\phi(1):=\what{P}_\phi(1)=\frac{2}{3} $
 $\what{P}^2_\phi(0):=\what{P}_\phi(0|\hat{Y}_1) = \frac{1}{3}$; $\what{P}^2_\phi(1):=\what{P}_\phi(1|\hat{Y}_1)=  \frac{2}{3}$. Observe that $Y_1$ and $Y_2$ are i.i.d.
Finally, conditioned on first two observations,  assume that $\what{P}^3_\phi(0):=\what{P}_\phi(0|\hat{Y}_1,\hat{Y}_2) = \frac{\hat{Y}_1+\hat{Y}_2}{2}$ and $\what{P}^3_\phi(1):=\what{P}_\phi(1|\hat{Y}_1,\hat{Y}_2) = 1-\frac{\hat{Y}_1+\hat{Y}_2}{2}$.   Here, it is evident that $\what{P}_\phi(y|Y_{\pi(1)},Y_{\pi(2)}) = \what{P}_\phi(y|Y_1,Y_2)$ satisfying Property \ref{property:conditional-permutation-invariant}. However, we observe that  $\E \left( \what{P}^{3}_\phi(0) \mid \hat{Y}_{1}\right) = \E \left( \frac{\hat{Y}_1 + \hat{Y}_2}{2}\mid \hat{Y}_1\right) =  \frac{\hat{Y}_1}{2}+\frac{1}{3}$  is not equal to $\what{P}^2_{\phi}(0) = \frac{1}{3}$.  Thus, $  \E \left( \what{P}^{t+1}_\phi(y) \mid \hat{Y}_{1:t-1}\right) \neq  \what{P}^t_{\phi}(y)$, indicating that the sequence model is not exchangeable.

\end{myexample}


Although we have established that C-permutation invariant architectures do not achieve full exchangeability, it is still important to assess whether incorporating Property \ref{property:conditional-permutation-invariant} into transformer architectures offers any advantages. To investigate this, we compare it to the standard causal transformer architecture (Section \ref{sec:standard_causal_architecture}), which does not exhibit this property. Specifically, we analyze two architectures: (1) the conditionally permutation-invariant architecture and (2) the standard causal masking architecture. We describe their respective masking schemes, efficient training procedures, and computational requirements for inference.


\subsection{Conditionally permutation invariant architecture} 
\label{sec:conditionally_permutation_invariant_architecture}
As we have established, achieving conditional permutation invariance requires ensuring that all context points can attend to one another. We adopt the same masking scheme shown in Figure \ref{fig:permutation-invariant-mask-prior}, which, as previously discussed, has also been utilized in prior works such as~\cite{MullerHoArGrHu22, Hollmannetal25}.



\paragraph{An efficient training procedure:} Training directly in a naive manner can be inefficient, as it would require processing $T$ separate sequences, each of length  $i \in \{1, 2, \cdots, T\}$, for an sequence data of length $T$. A more efficient approach is to fix a context length $i$ and  train the transformer on multiple target points simultaneously. This process is then repeated for all possible context lengths $i$. The corresponding masking scheme is shown in Figure $\ref{fig:efficient-training-permutation-invariant-mask}$. Importantly, this method is equivalent to training for $P(y|x_{1:4}, y_{1:4}, x)$ across multiple values of 
 $x$ in parallel, solely to optimize training efficiency. However, during inference, a multi-step (autoregressive) prediction approach should be adopted, as described in Section \ref{sec:inference-inductive-bias}.

\paragraph{Inference compute:}   Predicting a sequence of length $T$ requires approximately $O(T^3)$ computational effort. A major drawback of this masking scheme is that, even with $KV$ caching, the compute cannot be reduced to $O(T^2)$. This limitation arises because,  at each inference step, all outputs from the attention heads must be recomputed as every point attends to the newly added context point (see Figure \ref{fig:inference-permutation-invariant} for details). This behavior contrasts with the typical causal masking approach, where such recomputation is avoided.

\begin{figure*}[t]
\centering \includegraphics[width=0.45\linewidth, height=4cm] {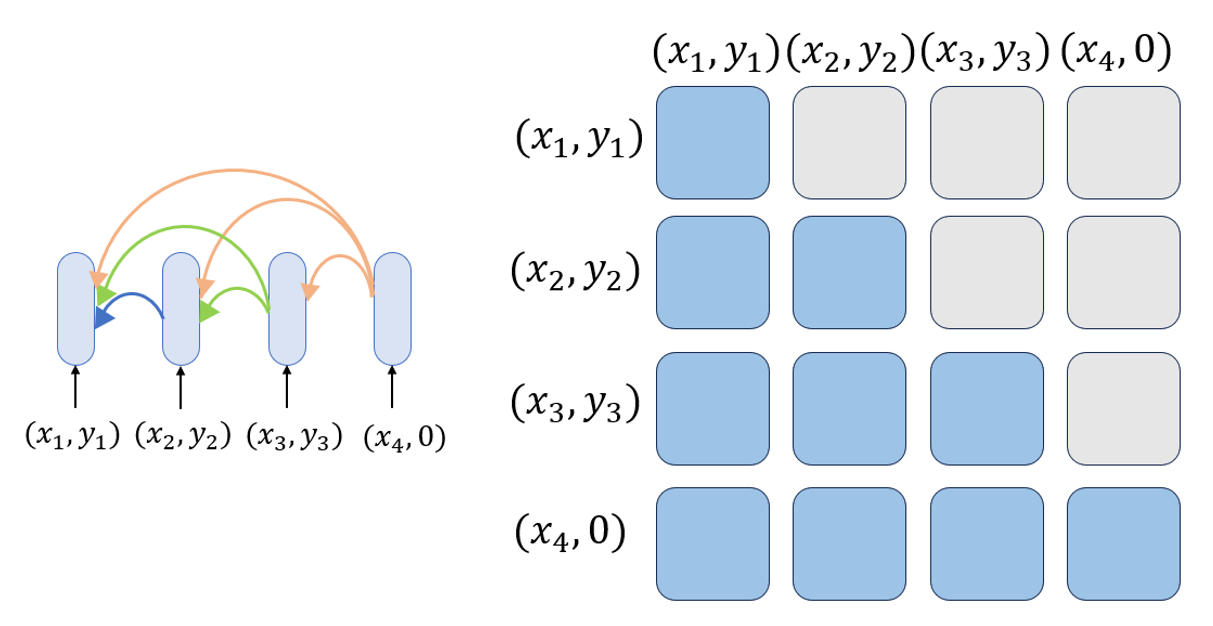}
  \caption{\textbf{Standard causal} transformer architecture: attention mechanism and masking scheme}
\label{fig:causal-mask}
\end{figure*}

\subsection{Standard causal architecture} 
\label{sec:standard_causal_architecture}

The masking scheme for this architecture is illustrated in Figure \ref{fig:causal-mask}. In this scheme, each context point attends only to the previous context points and not to any future points.  Specifically, the $i^{th}$ context point attends only to the context points within  $[1:i-1]$.

\paragraph{An efficient training procedure:}    As before, an efficient training procedure involves fixing a context length $i$ and training the transformer on multiple target points simultaneously. This process is then repeated for all possible context lengths. The corresponding masking scheme is shown in Figure $\ref{fig:efficient-traininig-causal-mask}$.

\paragraph{Inference compute:}   Predicting a  sequence of length $T$ requires approx. $O(T^3)$ computational effort. However, with $KV$ caching, this can be reduced to $O(T^2)$  (see Figure \ref{fig:inference-causal-mask}). 

\begin{figure}[h!]
\centering
\begin{minipage}[b]{0.32\textwidth}
\centering \includegraphics[width=0.9\textwidth, height=4cm]{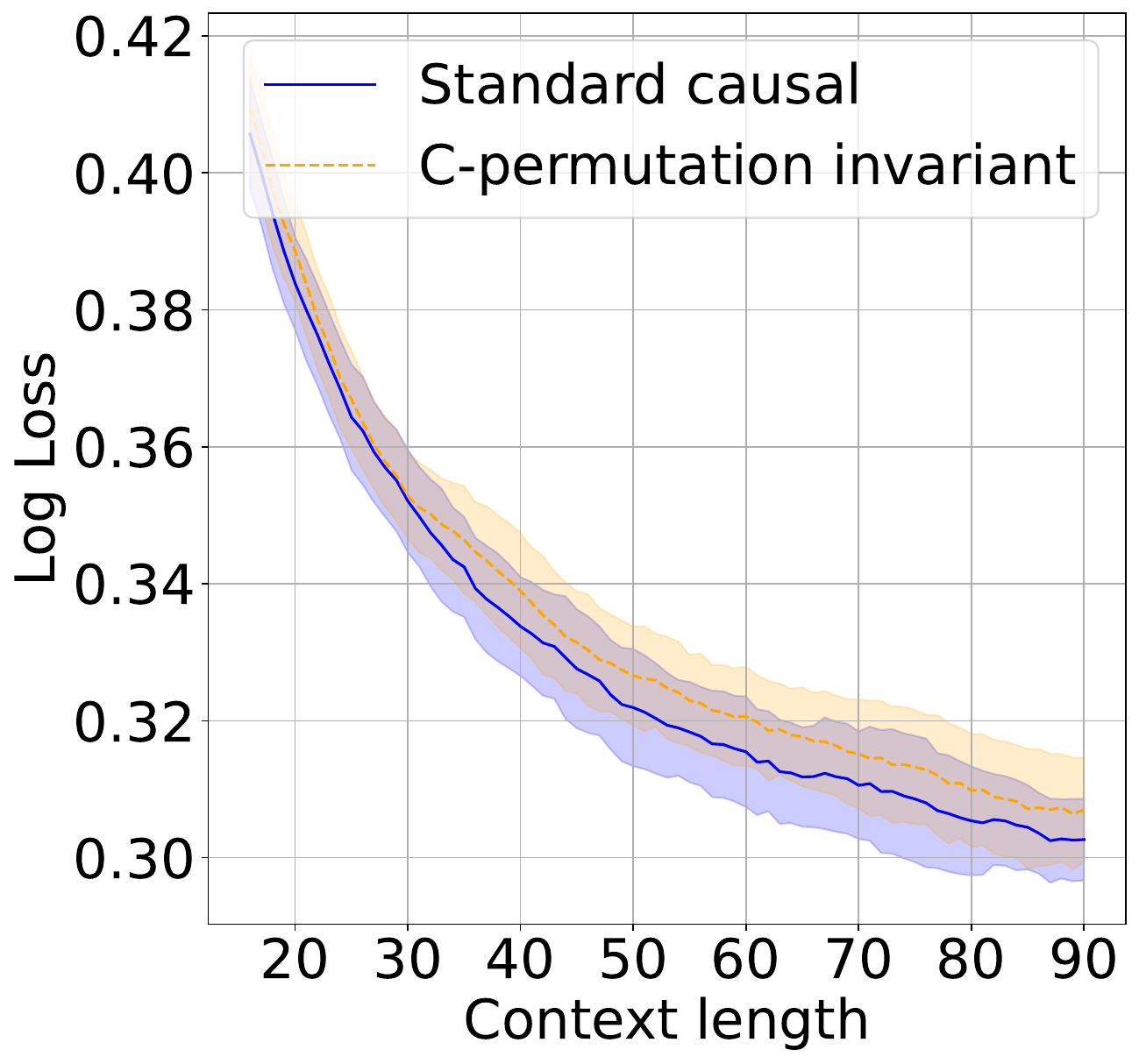}
  \caption*{\textbf{(a)} In-training horizon performance of two architectures}
\end{minipage}
\hfill
\begin{minipage}[b]{0.32\textwidth}
\centering \includegraphics[width=0.9\textwidth,height=4cm] {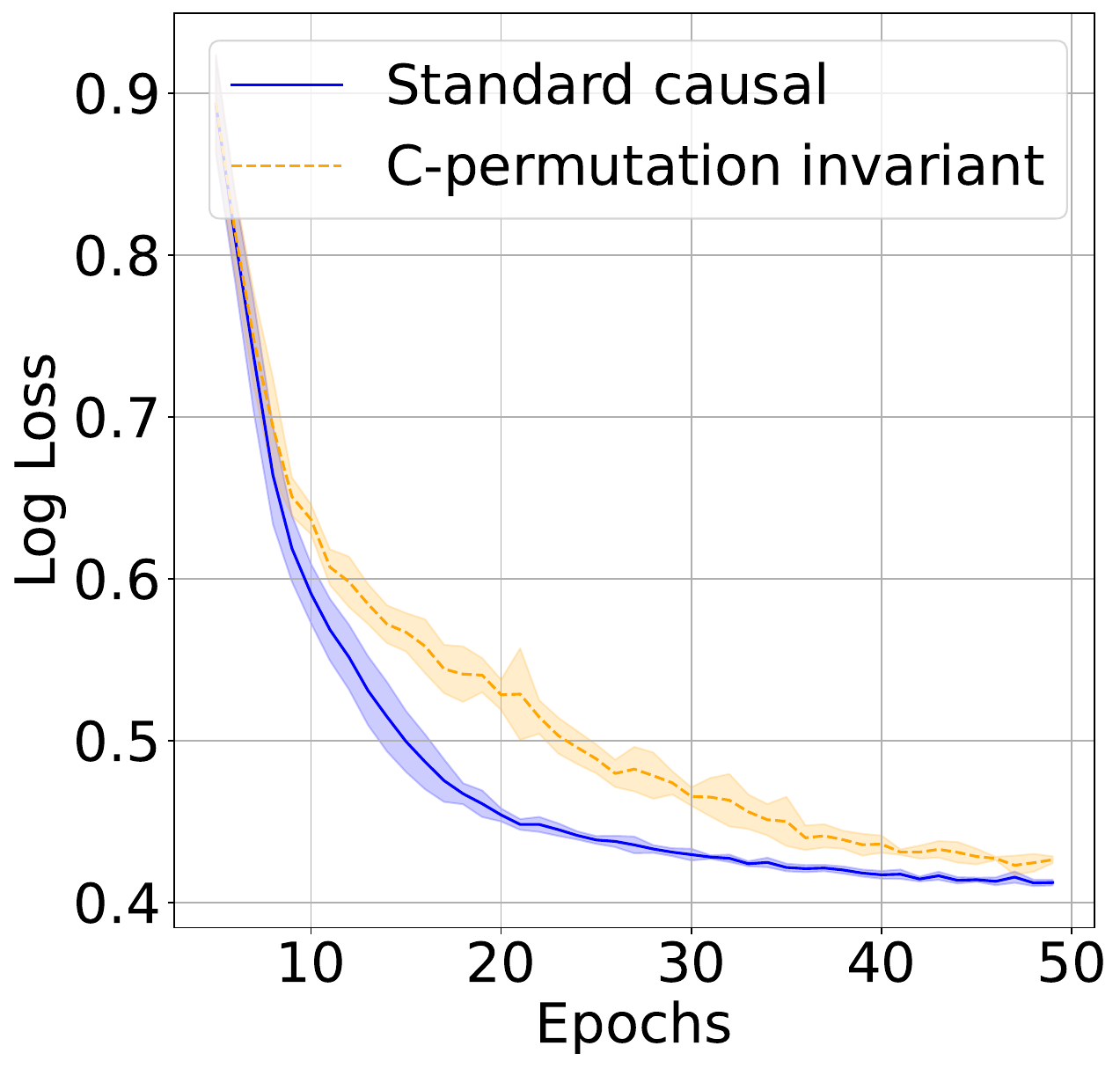}
  \caption*{\textbf{(b)} Training/data efficiency of two architectures}
\end{minipage}
\hfill
\begin{minipage}[b]{0.32\textwidth}
\centering \includegraphics[width=0.9\textwidth, height=4cm]
{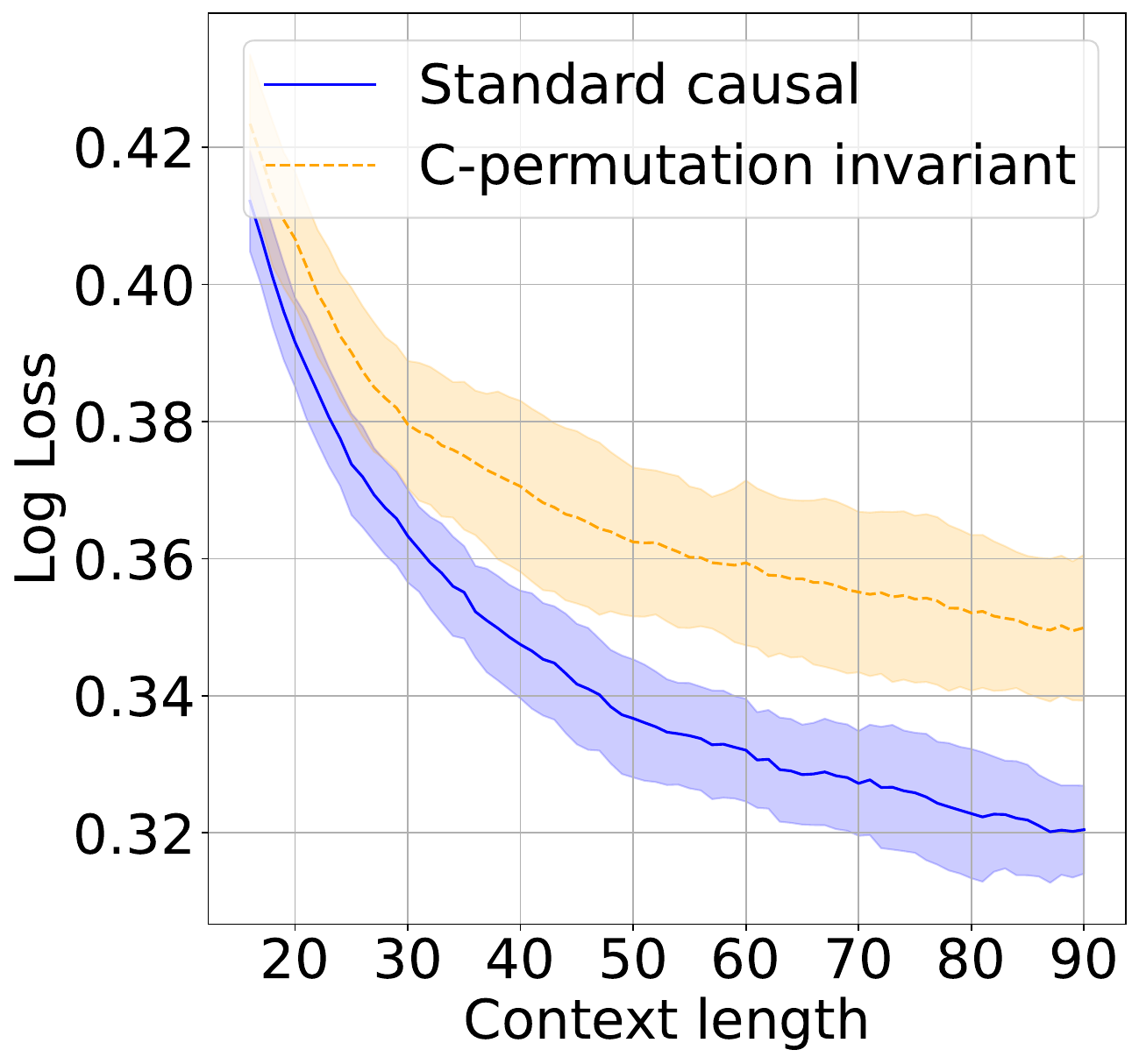}
\caption*{\textbf{(c)} Out-of-training horizon performance of two architectures}
\end{minipage}
\caption{\textbf{Comparing C-permutation invariant architecture and Standard causal architecture (lower is better):} \textbf{(a)} In-training horizon performance [Training horizon: 100, Metric: Multi-step log-loss, Target length: 10] \textbf{(b)} Training/Data efficiency [Training Horizon:100, Metric: Multi-step log-loss, Target length: 100] \textbf{(c)} Out-of-training horizon performance [Training horizon: 15, Metric: Multi-step log-loss, Target length: 10]}
\label{fig:comparing_two_architectures_main_diagram}
\end{figure}

\vspace{-3mm}

\subsection{Empirical investigation}
\label{sec:architectural_experiments}
In this section, we compare the contextual permutation invariance architecture and standard causal architectures. As is common in the literature~\cite{MullerHoArGrHu22, NguyenGr22,OsbandWeAsDwLuIbLaHaDoRo22}, we use the standard negative log-likelihood metric to compare these architectures and defer the analysis of downstream performance to Section \ref{sec:additional_experiments}.  Our findings indicate that the in-training horizon performance of the C-permutation-invariant architecture is comparable to that of the standard causal architecture. However, the standard causal architecture outperforms the C-permutation-invariant model on the out-of-training horizon by approximately $10\%$ and demonstrates greater training and data efficiency by up to $20\%$. Furthermore, as noted earlier in Section \ref{sec:Architecture_inductive_biases}, the standard causal architecture requires significantly less inference computation due to benefits from KV caching.


\vspace{-5mm}

\paragraph{Data generating process:} We evaluate these architectures in a controlled synthetic setting by generating a synthetic dataset using a Gaussian Process (GP). Specifically, we employ a GP with an RBF kernel: $f \sim \mathcal{GP}(m,\mc{K}) $, where $\mc{K}(X,X') = \sigma_f^2 \exp\left(-{||X-X'||_2^2}/{2\loss^2}\right)$. Additionally,   Gaussian   noise $N(0,\sigma^2)$ is added to the outputs.  The input $X$ is drawn i.i.d. from {$P_X$}. 


\vspace{-3mm}

\paragraph{Evaluation Metric:} To compare these two architectures, we use two metrics - one-step log-loss and multi-step log-loss. These metrics are described in detail in Section \ref{section:experiments-details}.

Both masking schemes are trained on data generated from Gaussian Processes using the loss function defined in (\ref{eqn:training_transformer}). Additional details about the architectures and training process are provided in Section \ref{section:experiments-details}. For brevity, we present only the results based on multi-step log-loss in the main body, while results for one-step log-loss are deferred to Section \ref{sec:additional_experiments}.


\vspace{-5mm}
\paragraph{Results:}  
We  evaluate these architectures across three dimensions - performance on sequence lengths within the training horizon, training/data efficiency, and performance on sequence lengths beyond the training horizon.

\begin{enumerate}
    \item \textit{In-training horizon performance:}  Figure \ref{fig:comparing_two_architectures_main_diagram}\textcolor{blue}{(a)} shows that the  performance of both  masking schemes is comparable, with no evident advantage of enforcing conditional-permutation invariance. Furthermore, additional ablation studies (see Section \ref{sec:additional_experiments}) indicate that standard-causal masking may outperform conditional-permutation invariance masking. 
    \item \textit{Data/Training efficiency:} To evaluate the data/training efficiency of these architectures, we compare their performance across various training epochs.    Figure \ref{fig:comparing_two_architectures_main_diagram}\textcolor{blue}{(b)} indicates that standard causal masking exhibits superior training efficiency. Similar findings were consistently observed in the ablation studies  (Section \ref{sec:additional_experiments}).
    \item \textit{Out-of-training horizon performance:} To assess this performance, we train both architectures up to a horizon $m$ and evaluate their performance on horizons beyond $m$. Figure~\ref{fig:comparing_two_architectures_main_diagram}\textcolor{blue}{(c)}, along with additional experiments (see Section \ref{sec:additional_experiments}), suggests that the standard causal masking may achieve better performance on out-of-training horizons.
\end{enumerate}

\section{Conclusion and future work}
\label{sec:conclusion}

We empirically and theoretically demonstrate that one-step inference, which has been the primary focus of the literature thus far, is insufficient for distinguishing between epistemic and aleatoric uncertainty, ultimately leading to suboptimal decision-making. In contrast, multi-step inference effectively overcomes this limitation, as evidenced by empirical results in multi-armed bandit and active learning settings.
On the architectural side, much of the existing work has focused on enforcing a masking scheme that, as we identify, only ensures conditional permutation invariance in transformers rather than full exchangeability. Empirically, we find that this approach performs worse than standard causal masking and significantly increases the computational cost of multi-step inference, as it cannot leverage KV caching.
These findings highlight a crucial research direction: developing improved architectural inductive biases for transformers to better model exchangeable sequences, making them more effective for decision-making and active exploration.

\newpage




\bibliographystyle{abbrvnat}

\setlength{\bibsep}{.7em}


\begin{thebibliography}{32}
\providecommand{\natexlab}[1]{#1}
\providecommand{\url}[1]{\texttt{#1}}
\expandafter\ifx\csname urlstyle\endcsname\relax
  \providecommand{\doi}[1]{doi: #1}\else
  \providecommand{\doi}{doi: \begingroup \urlstyle{rm}\Url}\fi

\bibitem[Aggarwal et~al.(2014)Aggarwal, Kong, Gu, Han, and Philip]{AggarwalKoGuHaPh14}
C.~C. Aggarwal, X.~Kong, Q.~Gu, J.~Han, and S.~Y. Philip.
\newblock Active learning: A survey.
\newblock In \emph{Data classification}, pages 599--634. Chapman and Hall/CRC, 2014.

\bibitem[Berti et~al.(2004)Berti, Pratelli, and Rigo]{BertiPrRi04}
P.~Berti, L.~Pratelli, and P.~Rigo.
\newblock Limit theorems for a class of identically distributed random variables.
\newblock \emph{Annals of Probability}, 32\penalty0 (3):\penalty0 2029 -- 2052, 2004.

\bibitem[Berti et~al.(2021)Berti, Dreassi, Pratelli, and Rigo]{BertiDrPrRi21}
P.~Berti, E.~Dreassi, L.~Pratelli, and P.~Rigo.
\newblock A class of models for bayesian predictive inference.
\newblock \emph{Bernoulli}, 27\penalty0 (1):\penalty0 702 -- 726, 2021.

\bibitem[Berti et~al.(2022)Berti, Dreassi, Leisen, Pratelli, and Rigo]{BertiDrLePrRi22}
P.~Berti, E.~Dreassi, F.~Leisen, L.~Pratelli, and P.~Rigo.
\newblock Bayesian predictive inference without a prior.
\newblock \emph{Statistica Sinica}, 34\penalty0 (1), 2022.

\bibitem[Brown et~al.(2020)Brown, Mann, Ryder, Subbiah, Kaplan, Dhariwal, Neelakantan, Shyam, Sastry, and Askell]{BrownEtAl20}
T.~B. Brown, B.~Mann, N.~Ryder, M.~Subbiah, J.~Kaplan, P.~Dhariwal, A.~Neelakantan, P.~Shyam, G.~Sastry, and A.~Askell.
\newblock Language models are few-shot learners.
\newblock In \emph{Advances in Neural Information Processing Systems 33}, 2020.

\bibitem[Cifarelli and Regazzini(1996)]{CifarelliRe96}
D.~M. Cifarelli and E.~Regazzini.
\newblock {D}e {F}inetti's contribution to probability and statistics.
\newblock \emph{Statistical Science}, 11\penalty0 (4):\penalty0 253--282, 1996.

\bibitem[{De Finetti}(1933)]{DeFinetti33}
B.~{De Finetti}.
\newblock Classi di numeri aleatori equivalenti. la legge dei grandi numeri nel caso dei numeri aleatori equivalenti. sulla legge di distribuzione dei valori in una successione di numeri aleatori equivalenti.
\newblock \emph{R. Accad. Naz. Lincei, Rf S 6a}, 18:\penalty0 107--110, 1933.

\bibitem[{De Finetti}(1937)]{DeFinetti37}
B.~{De Finetti}.
\newblock La pr{\'e}vision: ses lois logiques, ses sources subjectives.
\newblock In \emph{Annales de l'institut Henri Poincar{\'e}}, volume~7, pages 1--68, 1937.

\bibitem[{De Finetti}(2017)]{DeFinetti17}
B.~{De Finetti}.
\newblock \emph{Theory of probability: A critical introductory treatment}, volume~6.
\newblock John Wiley \& Sons, 2017.

\bibitem[Dosovitskiy et~al.(2021)Dosovitskiy, Beyer, Kolesnikov, Weissenborn, Zhai, Unterthiner, Dehghani, Minderer, Heigold, Gelly, Uszkoreit, and Houlsby]{Dosovitskiyetal21}
A.~Dosovitskiy, L.~Beyer, A.~Kolesnikov, D.~Weissenborn, X.~Zhai, T.~Unterthiner, M.~Dehghani, M.~Minderer, G.~Heigold, S.~Gelly, J.~Uszkoreit, and N.~Houlsby.
\newblock An image is worth 16x16 words: Transformers for image recognition at scale.
\newblock In \emph{International Conference on Learning Representations}, 2021.
\newblock URL \url{https://openreview.net/forum?id=YicbFdNTTy}.

\bibitem[Falck et~al.(2024)Falck, Wang, and Holmes]{FalckWaHo24}
F.~Falck, Z.~Wang, and C.~Holmes.
\newblock Is in-context learning in large language models bayesian? a martingale perspective, 2024.
\newblock URL \url{https://arxiv.org/abs/2406.00793}.

\bibitem[Fong et~al.(2023)Fong, Holmes, and Walker]{FongHoWa23}
E.~Fong, C.~Holmes, and S.~G. Walker.
\newblock Martingale posterior distributions.
\newblock \emph{Journal of the Royal Statistical Society, Series B}, 2023.

\bibitem[Fortini and Petrone(2014)]{FortiniPe14}
S.~Fortini and S.~Petrone.
\newblock Predictive distribution (de f inetti's view).
\newblock \emph{Wiley StatsRef: Statistics Reference Online}, pages 1--9, 2014.

\bibitem[Fortini et~al.(2000)Fortini, Ladelli, and Regazzini]{FortiniLaRe00}
S.~Fortini, L.~Ladelli, and E.~Regazzini.
\newblock Exchangeability, predictive distributions and parametric models.
\newblock \emph{Sankhy{\=a}: The Indian Journal of Statistics, Series A}, pages 86--109, 2000.

\bibitem[Gardner et~al.(2024)Gardner, Perdomo, and Schmidt]{GardnerPeSc24}
J.~P. Gardner, J.~C. Perdomo, and L.~Schmidt.
\newblock Large scale transfer learning for tabular data via language modeling.
\newblock In \emph{The Thirty-eighth Annual Conference on Neural Information Processing Systems}, 2024.

\bibitem[Hahn et~al.(2018)Hahn, Martin, and Walker]{HahnMaWa18}
P.~R. Hahn, R.~Martin, and S.~G. Walker.
\newblock On recursive bayesian predictive distributions.
\newblock \emph{Journal of the American Statistical Association}, 113\penalty0 (523):\penalty0 1085--1093, 2018.

\bibitem[Han et~al.(2024)Han, Yoon, Arik, and Pfister]{HanYoArPf24}
S.~Han, J.~Yoon, S.~O. Arik, and T.~Pfister.
\newblock Large language models can automatically engineer features for few-shot tabular learning.
\newblock In \emph{Proceedings of the 41st International Conference on Machine Learning}, 2024.

\bibitem[Hegselmann et~al.(2023)Hegselmann, Buendia, Lang, Agrawal, Jiang, and Sontag]{HegselmannBuLaAgJiSo23}
S.~Hegselmann, A.~Buendia, H.~Lang, M.~Agrawal, X.~Jiang, and D.~Sontag.
\newblock Tabllm: Few-shot classification of tabular data with large language models.
\newblock In F.~Ruiz, J.~Dy, and J.-W. van~de Meent, editors, \emph{Proceedings of the 26 International Conference on Artificial Intelligence and Statistics}, volume 206 of \emph{Proceedings of Machine Learning Research}, pages 5549--5581. PMLR, 25--27 Apr 2023.
\newblock URL \url{https://proceedings.mlr.press/v206/hegselmann23a.html}.

\bibitem[Hewitt and Savage(1955)]{HewittSa55}
E.~Hewitt and L.~J. Savage.
\newblock Symmetric measures on cartesian products.
\newblock \emph{Transactions of the American Mathematical Society}, 80\penalty0 (2):\penalty0 470--501, 1955.

\bibitem[Hollmann et~al.(2025)Hollmann, M{\"u}ller, Purucker, Krishnakumar, K{\"o}rfer, Hoo, Schirrmeister, and Hutter]{Hollmannetal25}
N.~Hollmann, S.~M{\"u}ller, L.~Purucker, A.~Krishnakumar, M.~K{\"o}rfer, S.~B. Hoo, R.~T. Schirrmeister, and F.~Hutter.
\newblock Accurate predictions on small data with a tabular foundation model.
\newblock \emph{Nature}, 637\penalty0 (8045):\penalty0 319--326, 2025.

\bibitem[Kaufmann et~al.(2012)Kaufmann, Cappe, and Garivier]{KaufmannCaGa12}
E.~Kaufmann, O.~Cappe, and A.~Garivier.
\newblock On bayesian upper confidence bounds for bandit problems.
\newblock In N.~D. Lawrence and M.~Girolami, editors, \emph{Proceedings of the Fifteenth International Conference on Artificial Intelligence and Statistics}, volume~22 of \emph{Proceedings of Machine Learning Research}, pages 592--600, La Palma, Canary Islands, 21--23 Apr 2012. PMLR.

\bibitem[M{\"u}ller et~al.(2022)M{\"u}ller, Hollmann, Arango, Grabocka, and Hutter]{MullerHoArGrHu22}
S.~M{\"u}ller, N.~Hollmann, S.~P. Arango, J.~Grabocka, and F.~Hutter.
\newblock Transformers can do bayesian inference.
\newblock In \emph{Proceedings of the Tenth International Conference on Learning Representations}, 2022.

\bibitem[Nguyen and Grover(2022)]{NguyenGr22}
T.~Nguyen and A.~Grover.
\newblock Transformer neural processes: Uncertainty-aware meta learning via sequence modeling.
\newblock In \emph{Proceedings of the 39th International Conference on Machine Learning}, 2022.

\bibitem[Osband et~al.(2022)Osband, Wen, Asghari, Dwaracherla, Lu, Ibrahimi, Lawson, Hao, O'Donoghue, and Roy]{OsbandWeAsDwLuIbLaHaDoRo22}
I.~Osband, Z.~Wen, S.~M. Asghari, V.~Dwaracherla, X.~Lu, M.~Ibrahimi, D.~Lawson, B.~Hao, B.~O'Donoghue, and B.~V. Roy.
\newblock {The Neural Testbed: Evaluating Joint Predictions}.
\newblock In \emph{Advances in Neural Information Processing Systems}, 2022.

\bibitem[Russo et~al.(2018)Russo, Van~Roy, Kazerouni, Osband, and Wen]{RussoVaKaOsWe18}
D.~J. Russo, B.~Van~Roy, A.~Kazerouni, I.~Osband, and Z.~Wen.
\newblock A tutorial on thompson sampling.
\newblock \emph{Foundations and Trends{\textregistered} in Machine Learning}, 11\penalty0 (1):\penalty0 1--96, 2018.

\bibitem[Settles(2009)]{Settles09}
B.~Settles.
\newblock Active learning literature survey.
\newblock 2009.

\bibitem[Weber(1992)]{weber92}
R.~Weber.
\newblock On the gittins index for multiarmed bandits.
\newblock \emph{The Annals of Applied Probability}, pages 1024--1033, 1992.

\bibitem[Wen et~al.(2022)Wen, Osband, Qin, Lu, Ibrahimi, Dwaracherla, Asghari, and Roy]{WenOsQiLuIbDwAsVa22}
Z.~Wen, I.~Osband, C.~Qin, X.~Lu, M.~Ibrahimi, V.~Dwaracherla, M.~Asghari, and B.~V. Roy.
\newblock From predictions to decisions: The importance of joint predictive distributions.
\newblock \emph{arXiv:2107.09224 [cs.LG]}, 2022.

\bibitem[Yan et~al.(2024)Yan, Zheng, Xu, Zhu, Chen, Sun, Wu, and Chen]{YanZhXuZhChSuWuCh24}
J.~Yan, B.~Zheng, H.~Xu, Y.~Zhu, D.~Chen, J.~Sun, J.~Wu, and J.~Chen.
\newblock Making pre-trained language models great on tabular prediction.
\newblock In \emph{The Twelfth International Conference on Learning Representations}, 2024.

\bibitem[Ye and Namkoong(2024)]{YeNa24}
N.~Ye and H.~Namkoong.
\newblock Exchangeable sequence models quantify uncertainty over latent concepts, 2024.
\newblock URL \url{https://arxiv.org/abs/2408.03307}.

\bibitem[Zhang et~al.(2024)Zhang, Namkoong, Russo, et~al.]{ZhangCaNaRu24}
K.~W. Zhang, H.~Namkoong, D.~Russo, et~al.
\newblock Posterior sampling via autoregressive generation.
\newblock \emph{arXiv:2405.19466 [cs.LG]}, 2024.

\bibitem[Zhao et~al.(2023)Zhao, Birke, and Chen]{zhaoBiCh23}
Z.~Zhao, R.~Birke, and L.~Chen.
\newblock Tabula: Harnessing language models for tabular data synthesis.
\newblock \emph{arXiv preprint arXiv:2310.12746}, 2023.

\end{thebibliography}

\newpage
\appendix
\section{Additional Demonstrations}
\label{sec:Additional_demonstrations}

\begin{figure*}[h!]
\centering
\begin{minipage}[b]{0.48\textwidth}
\centering \includegraphics[width=0.7\textwidth, height=5.5cm]{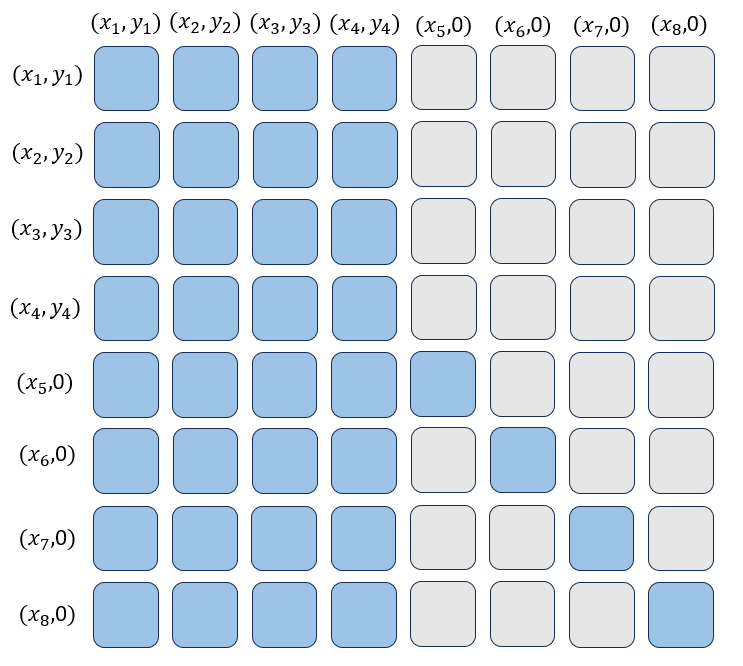}
  \caption{Efficient training procedure for conditionally permutation invariant transformer}
\label{fig:efficient-training-permutation-invariant-mask}
\end{minipage}
\hfill
\begin{minipage}[b]{0.48\textwidth}
\centering \includegraphics[width=0.7\textwidth, height=5.5cm]{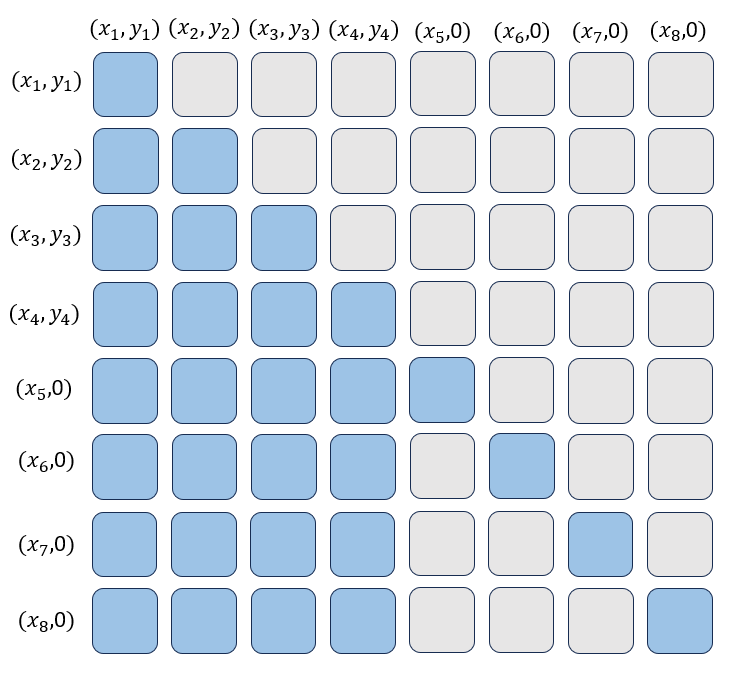}
  \caption{Efficient training procedure for standard causal transformer architecture}
\label{fig:efficient-traininig-causal-mask}
\end{minipage}
\end{figure*}

\begin{figure}[h!]
\centering \includegraphics[width=0.9\textwidth, height=5cm]{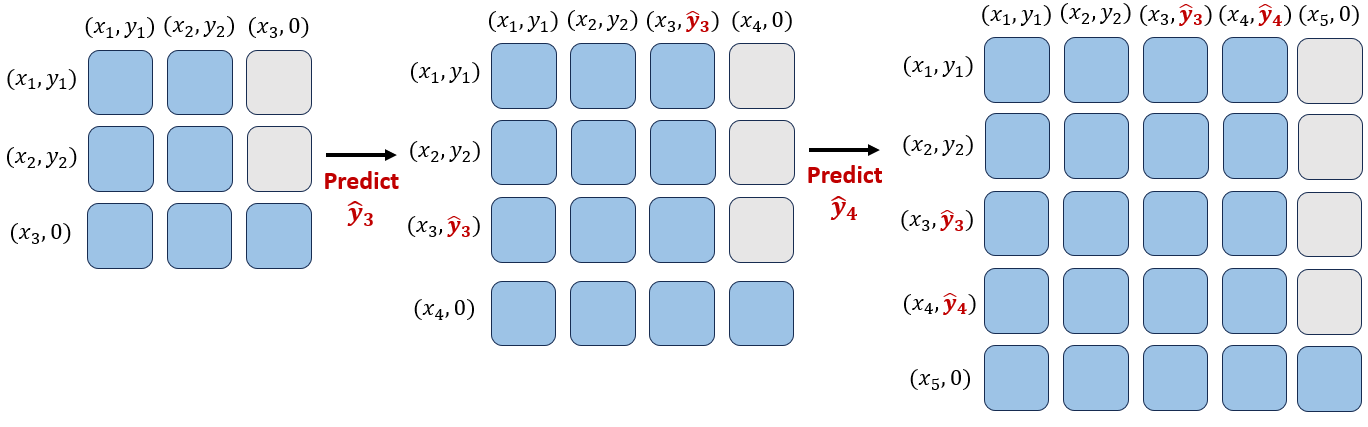}
\caption{Conditional permutation invariant architecture - masking scheme at inference stage}
\label{fig:inference-permutation-invariant}
\end{figure}

\begin{figure}[h!]
\centering \includegraphics[width=0.9\textwidth, height=5cm]{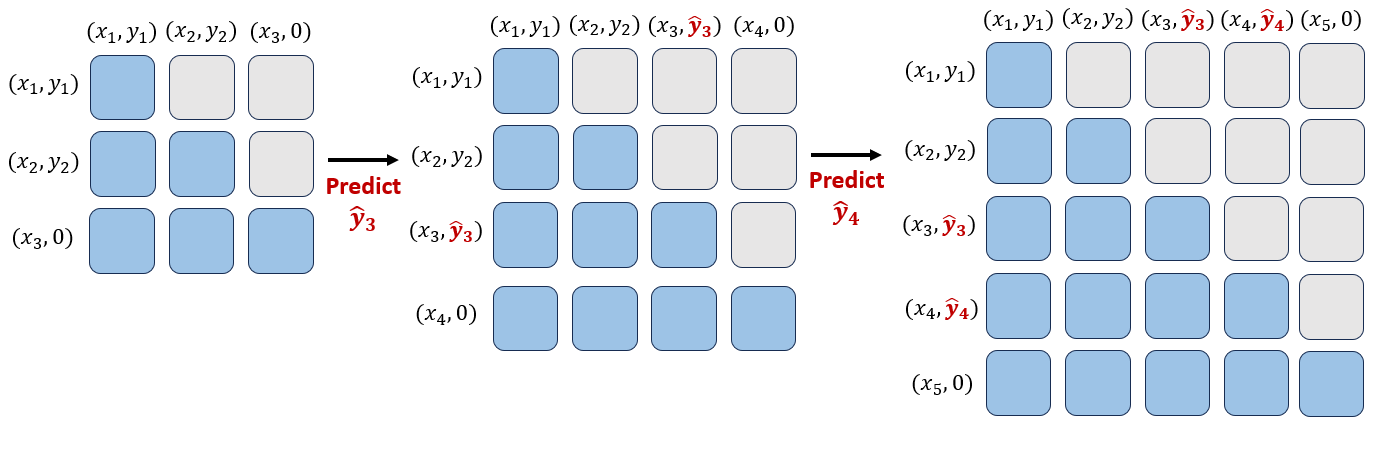}
\caption{Standard causal architecture - masking scheme at inference stage}
\label{fig:inference-causal-mask}
\end{figure}

\newpage

\section{Experiments Details}
\label{section:experiments-details}

In this section, we present the experimental details from Sections \ref{sec:inference_experiments} and \ref{sec:architectural_experiments}.

\subsection{Details for experiments in Section \ref{sec:architectural_experiments}}

\paragraph{Data generating process:} As previously mentioned, we generate data synthetically using Gaussian processes. Specifically, we employ a Gaussian Process (GP) with a Radial Basis Function (RBF) kernel: $f \sim \mathcal{GP}(m,\mc{K}) $, where $m(X)$ represents the mean function, and $\mc{K}(X,X') = \sigma_f^2 \exp\left(-\frac{||X-X'||_2^2}{2\loss^2}\right)$ represents the covariance function. Additionally,   Gaussian   noise $N(0,\sigma^2)$ is added to the outputs.  The input $X$ is drawn i.i.d. from {$P_X$}. Unless stated otherwise, the parameters are set as follows: $m(X)=0$,   $X\sim U[-2.0,2.0]$, $\sigma_f =1.0$, $\loss=1.0$, $\sigma=0.1$. 

\paragraph{Evaluation Metric:} To compare these two architectures, we employ two metrics: one-step log-loss and multi-step log-loss. A detailed description of these metrics is provided below.
\begin{enumerate}
    \item \textit{One-step log-loss:} Given $(x_{1:m},y_{1:m},x_{m+1})$, we generate ${y}_{m+1}$ from the true data generating process, i.e., ${y}_{m+1} \sim \P (\cdot|x_{m+1},x_{1:m},y_{1:m})$. The one-step log-loss for the sequence model is then calculated as:
$-\log \left[ \what{P}_\phi({y}_{m+1}|x_{i},x_{1:m},y_{1:m}) \right].$
This value is further averaged over multiple instances of $(x_{1:m},y_{1:m},x_{m+1})$.
\item  \textit{Multi-step log-loss:}
Given $(x_{1:m},y_{1:m},x_{m+1:T})$, we generate ${y}_{m+1:T}$ from the true data-generating process, i.e., 
${y}_{m+1:T} \sim \P(\cdot|x_{m+1:T},x_{1:m},y_{1:m})$.  The multi-step log-loss for the sequence model is computed as:
$-\log \left[ \what{P}_\phi ({y}_{m+1:T}|x_{m+1:T},x_{1:m},y_{1:m})\right]$, which can be expressed as: $-\sum_{i=m}^{T-1} \log \left[  \what{P}_\phi ({y}_{i+1}|x_{m+1:i+1},x_{1:m},y_{1:m},{y}_{m:i})\right].$
This value is then averaged over multiple instances of $(x_{1:m},y_{1:m},x_{m+1:T})$.
\end{enumerate}


We refer to $t$  as the \textit{context-length} and $(T-t)$ as the \textit{target length}.

\paragraph{Transfromer architecture and training details:} To compare the conditionally permutation-invariant architecture with the standard causal architecture, we use a decoder-only transformer with the following parameters. Both architectures share the same parameters, differing only in their masking schemes. The model parameters are as follows:
\begin{itemize}
    \item Model dimension: 64
    \item Feedforward dimension: 256
    \item Number of attention heads: 4
    \item Number of transformer layers: 4 
    \item Dropout: 0.1
    \item Activation function: GELU
\end{itemize}
For embedding $(x,y)$, 
we use a neural network with two layers of sizes 
 $[256, 64]$. Additionally, a final linear layer is used to predict the mean 
 $\mu$ and standard deviation $\sigma$ of the output distribution, modeled as $Y \sim N(\mu,\sigma^2)$. For training the transformers, we use the Adam optimizer with default parameters, and the learning rate is adjusted using a cosine scheduler. The training parameters are as follows: Warmup ratio is $0.03$, minimium learning rate is $3.0e^{-5}$, learning rate is   $0.0003$, weight decay to $0.01$ and batch size is $64$.

 \paragraph{Evaluation:} We evaluate the trained transformers, corresponding to each architecture, using marginal log-likelihood and joint log-likelihood. For each experiment, we use $8192$ test samples and conduct evaluations across five different random seeds. This process includes retraining the models on different training datasets and evaluating them on distinct test datasets.


\subsection{Experimental details for Section \ref{sec:inference_experiments} - Bayesian Bandits}

\paragraph{Data generating process:} Recall that, for each arm  $a \in \{C,D\}$, the rewards are distributed as  $Y^{(a)}_{1:T} \simiid  N\left(\theta^{(a)}, \left(\tau^{(a)}\right)^2\right)$, where the mean reward  $\theta^{(a)}$ follows a prior distribution $\theta^{(a)} \sim N\left(\mu^{(a)}, \left(\sigma^{(a)}\right)^2\right)$. Where for Arm $C$, we set $\mu^{(C)} = 0$, $\sigma^{(C)}=0.5$ and $\tau^{(C)}=0.5$. While for Arm $D$, we set $\mu^{(D)} = 0$, $\sigma^{(D)}=0.9$ and $\tau^{(D)}=0.1$.

\textbf{Exploration Algorithm:} Numerous algorithms have been proposed in the literature to address this trade-off, such as Thompson Sampling~\cite{RussoVaKaOsWe18}, Bayes-UCB~\cite{KaufmannCaGa12} and Gittin's index~\cite{weber92}. However, to compare the two inference strategies, we fix the algorithm to Thompson Sampling and implement it using the respective strategy.

\paragraph{Transformer Architecture and Training:} As specified earlier for each arm $a\in\{C,D\}$, we train a separate transformer (decoder-only). The model parameters and training details are specified below:

\begin{itemize}
    \item Model dimension: 64
    \item Feedforward dimension: 256
    \item Number of attention heads: 4
    \item Number of transformer layers: 4 
    \item Dropout: 0.1
    \item Activation function: GELU
\end{itemize}

For embedding $(x,y)$, 
we use a neural network with two layers of sizes 
 $[256, 64]$. Note that, as in this case there is no context, we set $x$ as $0$. Additionally, a final linear layer is used to predict the mean 
 $\mu$ and standard deviation $\sigma$ of the output distribution, modeled as $Y \sim N(\mu,\sigma^2)$. For training the transformers, we use the Adam optimizer with default parameters, and the learning rate is adjusted using a cosine scheduler. The training parameters are as follows: Warmup ratio is $0.03$, minimium learning rate is $3.0e^{-5}$, learning rate is   $0.0003$, weight decay to $0.01$ and batch size is $64$.

\paragraph{Evaluation Details:} The results presented in Figure \ref{fig:one_step_multi_step_all_three}\textcolor{blue}{(b)} are averaged over $1000$ different experiments, with each experiment run for $100$ steps.

\paragraph{One-step / Multi-step inference (Thompson sampling):} In Algorithm \ref{alg:ts-gaussian-one-multi-step-inference},   we outline the implementation of a Thompson sampling algorithm for one-step and multi-step inference using a trained transformer. In Algorithm  \ref{alg:ts-gaussian-one-multi-step-inference} setting $J=1$ corresponds to  one-step inference, while choosing $J>1$  enables multi-step inference. In our experiments, we set $J=100$ for multi-step inference. For reference, we also provide the standard Thompson sampling algorithm for Gaussian-Gaussian multi-armed bandits (see Algorithm \ref{alg:ts-gaussian}). In our experiments, we set $J=100$ for multi-step inference.

        


\begin{algorithm}[h!]
\caption{One-step and Multi-step inference (Thompson sampling)  using sequence models (transformers) in multi armed bandits setting}
\label{alg:ts-gaussian-one-multi-step-inference}
\begin{algorithmic}[1]
    \REQUIRE  Trained transformer $(\what{P}_\phi^{(a)})$ for each of the Arm $a \in [1,K]$, horizon \(T\), Number of autoregressive generations in each iteration ($J$) : In one-step inference $J=1$, while for multi-step inference $J>1$.
    \STATE \textbf{Initialization:}
    \FOR{\(a = 1 \to K\)}
       
        \STATE $\mathcal{Y}^{(a)} \leftarrow \{\}$ \hfill // Observed rewards for each arm  \(a\)
        
        \STATE \(n^{(a)} \leftarrow 0\) \hfill // number of pulls for arm \(a\)
        
    \ENDFOR

    \FOR{\(t = 1 \to T\)}
        \STATE \textbf{Autoregressively} generate J rewards for each arm, conditioned on observed rewards
        \FOR{$j=1\to J$}
        \STATE \[
           \hat{Y}^{(a)}_{t, n^{(a)}+j} \sim \what{P}_\phi(\cdot|\mathcal{Y}^{(a)}, \hat{Y}^{(a)}_{t, n^{(a)}+1}, \cdots, \hat{Y}^{(a)}_{t, n^{(a)}+j-1})
        \]  \hfill // In $\hat{Y}^{(a)}_{t, n^{(a)}+1}$, $t$ indicates time-step and $ n^{(a)}+j$ indicates conditioning on $|\mathcal{Y}^{(a)}| = n^{(a)}$ observations and $j-1$ generations.
        \ENDFOR
        \STATE \textbf{Select} arm \(A_t = \displaystyle \arg\max_{1 \le a \le K} \frac{1}{J}\sum_{j=1}^J\hat{Y}^{(a)}_{t, n^{(a)}+j}\).
        \STATE \textbf{Pull} arm \(A_t\) and \textbf{observe} reward \(Y_t^{(A_t)}\).
        \STATE \textbf{Update} the collection of observations for arm \(A_t\):
        \[
           n^{(A_t)} \leftarrow n^{(A_t)} + 1,
           \quad
           \mathcal{Y}^{(A_t)} \leftarrow \mathcal{Y}^{(A_t)} \cup \{Y_t^{(A_t)}\}.
        \]
    \ENDFOR
\end{algorithmic}
\end{algorithm}

\begin{algorithm}[h!]
\caption{Thompson Sampling for Multi Armed Bandits (Gaussian-Gaussian setting)}
\label{alg:ts-gaussian}
\begin{algorithmic}[1]
    \REQUIRE Number of arms \(K\), prior mean \(\mu^{(a)}\), prior variance \(\left(\sigma^{(a)}\right)^2\), known reward variance \(\left(\tau^{(a)}\right)^2\), horizon \(T\).
    \STATE \textbf{Initialization:}
   
    \FOR{\(a = 1 \to K\)}
        
        \STATE \(n^{(a)} \leftarrow 0\) \hfill // number of pulls for arm \(a\)
        
        \STATE \(S^{(a)} \leftarrow 0\) \hfill // sum of rewards for arm \(a\)
        
        \STATE \(\hat{\mu}^{(a)} \leftarrow \mu^{(a)}\) \hfill // posterior mean for arm \(a\)
        
        \STATE \(\left(\hat{\sigma}^{(a)}\right)^2 \leftarrow \left(\sigma^{(a)}\right)^2\) \hfill // posterior variance for arm \(a\)
    \ENDFOR

    \FOR{\(t = 1 \to T\)}
        \STATE \textbf{Sample} a mean from each arm’s posterior:
        \[
           \tilde{\mu}_t^{(a)} \sim \mathcal{N}\left(\hat{\mu}^{(a)},\, \left(\hat{\sigma}^{(a)}\right)^2\right)
           \quad \text{for } a = 1, \dots, K.
        \]
        \STATE \textbf{Select} arm \(A_t = \displaystyle \arg\max_{1 \le a \le K} \tilde{\mu}_t^{(a)}\).
        \STATE \textbf{Pull} arm \(A_t\) and \textbf{observe} reward \(Y_t^{(A_t)}\).
        \STATE \textbf{Update} the sufficient statistics for arm \(A_t\):
        \[
           n^{(A_t)} \leftarrow n^{(A_t)} + 1,
           \quad
           S^{(A_t)} \leftarrow S^{(A_t)} + Y_t^{(A_t)}.
        \]
        \STATE \textbf{Update} posterior for arm \(A_t\).

        \[
           \text{Posterior mean: } 
           \hat{\mu}^{(A_t)} \leftarrow 
           \frac{\frac{1}{\left(\sigma^{(A_t)}\right)^2}\,\mu^{(A_t)} \;+\; \frac{S^{(A_t)}}{\sigma^2}}{\frac{1}{\left(\sigma^{(A_t)}\right)^2} + \frac{n^{(A_t)}}{\tau^2}}
       \]\[
           \text{Posterior variance: } 
           \left(\hat{\sigma}^{(A_t)}\right)^2 \leftarrow 
           \left(\frac{1}{\left(\sigma^{(A_t)}\right)^2} + \frac{n^{(A_t)}}{\tau^2}\right)^{-1}.
        \]

    \ENDFOR
\end{algorithmic}
\end{algorithm}


\subsection{Experimental details for Section \ref{sec:inference_experiments} - Active Learning}

\paragraph{Data generating process:} Recall that our data generating process is as follows - features $X\simiid P_X$, and outcomes are generated from an unknown function $f^*$, such that $ Y=f^*(X)+\epsilon_X$, with noise $\epsilon_X \sim N(0, \tau_X^2)$ being heteroscedastic and the data-generating function $f^*$ drawn from a distribution $\mu$.  
Our $P_X$ consists of 100 \textit{non-overlapping clusters}.  Further, within each cluster $f^*(X)$ is highly correlated, while across clusters, the correlation is low. Further on some clusters the noise $\epsilon_X$ is high, while on others it is low. This setting introduces a setting where it is necessary to differentiate between aleatoric and epistemic noise for efficiently querying the labels for model improvement.

\noindent\textbf{Exploration Algorithm:} There are various active learning query strategies, such as  uncertainty sampling techniques (e.g., margin-sampling or entropy) and Bayesian Active Learning by Disagreement (BALD)~\cite{Settles09, AggarwalKoGuHaPh14}. For this study, we focus on Uncertainty Sampling adapted to the regression setting.

\paragraph{Transformer Architecture and Training:} We train a decoder-only transformer on sequential-data $\{(X_{1:N}^{(j)},Y_{1:N}^{(j)}):j \in [1,M]\}$ generated from the original data generating process. The model parameters and training details are specified below.

\begin{itemize}
    \item Model dimension: 64
    \item Feedforward dimension: 256
    \item Number of attention heads: 4
    \item Number of transformer layers: 4 
    \item Dropout: 0.1
    \item Activation function: GELU
\end{itemize}
For embedding $(x,y)$, 
we use a neural network with two layers of sizes 
 $[256, 64]$. Additionally, a final linear layer is used to predict the mean 
 $\mu$ and standard deviation $\sigma$ of the output distribution, modeled as $Y \sim N(\mu,\sigma^2)$. For training the transformers, we use the Adam optimizer with default parameters, and the learning rate is adjusted using a cosine scheduler. The training parameters are as follows: Warmup ratio is $0.03$, minimium learning rate is $3.0e^{-5}$, learning rate is   $0.0003$, weight decay to $0.01$ and batch size is $64$.

\paragraph{Evaluation Details:} Results shown in Figure \ref{fig:one_step_multi_step_all_three}\textcolor{blue}{(c)} are averaged over $50000$ experiments with each experiment run for $50$ steps.

\paragraph{One-step / Multi-step inference (Uncertainty sampling):}
In Algorithm \ref{alg:uncertainty-sampling-one-multi-step-inference} we describe the implementation of   one-step and multi-step inference based uncertainty sampling using trained transformers.  In our multi-step inference experiments, we set 
=
$J=20$ and $I=20$.

\begin{algorithm}[h!]
\caption{One-step and Multi-step inference (Uncertainty sampling)  using sequence models (transformers) in active learning setting}
\label{alg:uncertainty-sampling-one-multi-step-inference}
\begin{algorithmic}[1]
    \REQUIRE  Trained transformer $(\what{P}_\phi)$, Horizon \(T\), Initial data available $\mathcal{D}^{0}$,  Pool to choose from $\mathcal{X}^{pool}$. Number of autoregressive generations in each iteration ($J$) : In one-step inference $J=1$, while for multi-step inference $J>1$; Number of generation paths over which variance is taken $I$.

    \FOR{\(t = 1 \to T\)}
        \FOR{$X\in \mathcal{X}^{pool}$}
        \STATE Generate $I$ trajectories for each sample $X$ and collect the mean output on each trajectory in list $l_{t,X}$.
        \STATE Initialize $l_{t,X} = \{\}$
        \FOR{$i=1 \to I$}
        \STATE \textbf{Autoregressively} generate J sample outputs for $X$ for each arm, conditioned on the available data.
        
        \FOR{$j=1\to J$} 
        \STATE
        \[
           \hat{Y}^{(i,X)}_{t, j} \sim \what{P}_\phi(\cdot|\mathcal{D}^{t-1},  (X,\hat{Y}^{(i,X)}_{t, 1}), \cdots, (X,\hat{Y}^{(i.X)}_{t, j-1}),X)
        \]  
        \ENDFOR
        \STATE $l_{(t,X)} \leftarrow l_{(t,X)} \cup \{\frac{1}{J}\sum_{j=1}^J \hat{Y}^{(i,X)}_{t, j}\}$
        \ENDFOR
        \STATE Estimate variance of the mean output across $I$ trajectories $\hat{V}_{(t,X)} = Variance_{\bar{Y}\in l_{(t,X)}}(\bar{Y})$
        \ENDFOR
        \STATE \textbf{Select}  \(X^{*} = \displaystyle \arg\max_{X \in \mathcal{X}^{pool}} \hat{V}_{t,X}\).
        \STATE \textbf{Query} $X^{*}$ and get the true label/output $Y^{*}$.
        \STATE \textbf{Update} $\mathcal{D}^t \leftarrow \mathcal{D}^{t-1} \cup {(X^*,Y^*)}$
        
    \ENDFOR
\end{algorithmic}
\end{algorithm}


\section{Contextual setting definitions}
\label{sec:contextual_setting}

Consider a contextual setting, where the context $X\simiid P_X$. 

\noindent\textbf{Exchangeability definition:} An infinite sequence $(X_{1:\infty},Y_{1:\infty})$ is exchangeable if .for any $n$ and permutation $\pi$
\begin{align*}
\P((X_1,\Obs_{1}),\cdots, (X_n,\Obs_n)) =\P((X_{\pi(1)},\Obs_{\pi(1)}),\cdots, (X_{\pi(n)},\Obs_{\pi(n)})). 
\end{align*}

\noindent\textbf{Tranformer training:}
\begin{align*}
  \min_{\phi}
  \left\{
    -\frac{1}{N} \sum_{j=1}^N \sum_{i=0}^{T-1}
    \log \what{P}_\phi\left( \hat{\Obs}_{i+1}^j = \obs_{i+1}^j \mid  \hat{\Obs}_{1:i}^j = \obs_{1:i}^j, x_{1:i}^j, x_{i+1}^j  \right)
  \right\}.
\end{align*}

\noindent\textbf{Conditional permutation invariance property:}

\begin{align*}
\what{P}_\phi(\Obs_{t+1}|(X_1,\Obs_{1}),\cdots, (X_t,\Obs_t), X_{t+1}) =\what{P}_\phi(\Obs_{t+1}|(X_{\pi(1)},\Obs_{\pi(1)}),\cdots, (X_{\pi(t)},\Obs_{\pi(t)}), X_{t+1}). 
\end{align*}

\noindent\textbf{Conditionally identically distributed property:} In the presence of covariates,  where 
 $\Cov\sim P_\Cov$ independently, the c.i.d. property extends to the sequence model as follows: 
\begin{align*}
  \E\left(\what{P}_\phi^{t+1}\left( \obs \mid  \cov\right) \mid \hat{\Obs}_{1:t-1}, \Cov_{1:t-1} \right) = \what{P}_\phi^{t}(\obs \mid \cov).
\end{align*}

where $\what{P}_\phi\left(\hat{\Obs}_{t} = \obs \mid  \Cov_{t} = \cov,\hat{\Obs}_{1:t-1}, \Cov_{1:t-1}\right) \eqdef \what{P}_\phi^{t}(\obs\mid\cov)  $ and  $  \what{P}_\phi\left(\hat{\Obs}_{t+1} = \obs \mid  \Cov_{t+1} = \cov,\hat{\Obs}_{1:t}, \Cov_{1:t}\right) \eqdef \what{P}_\phi^{t+1}(\obs\mid\cov)$.

\section{Proofs}
\label{sec:proofs}

\subsection{Proof of Theorem \ref{thm:uq_one_step_suffers_non_contextual}}
\label{sec:proof_of_theorem_2}

Recall that $y_{t+1:T} \sim \P(\cdot|y_{1:t})$ is the data generating process. Further, $\what{P}^M_\phi(y_{t+1:T})   \equiv \prod_{i=t+1}^{T}\what{P}_\phi(\hat{Y}_{t+1}=y_{i}|y_{1:t})$ is the one-step (marginal) inference model and  $\what{P}_\phi^J(y_{t+1:T}) \equiv   \prod_{i=t+1}^{T}\what{P}_\phi(\hat{Y}_i=y_i|y_{1:i-1})$ is the multi-step (joint) inference model, where $\hat{Y}_i$ is generated autoregressively.  

\noindent\textbf{Theorem \ref{thm:uq_one_step_suffers_non_contextual}:} Assuming $\what{P}_\phi = \P$, then the difference  $\E \left(\log\left[ \what{P}^J_\phi(y_{t+1:T})\right]\right) -   \E \left(\log\left[ \what{P}^M_\phi(y_{t+1:T})  \right] \right)$ is equal to 
\[
     \sum_{i=t+1}^TI(y_{t+1:i-1}; y_{i}|y_{1:t})
\]

where $I(A;B|C)$ is the mutual information between $A$ and $B$ conditional on $C$ and expectation is w.r.t. $y_{1:T}\sim \P(\cdot)$.

\begin{proof} As data is generated from $\P(\cdot)$. Therefore, under the assumption that $\what{P}_\phi = \P$, we have that 

\begin{align*}
    \E \left(\log\left[ \what{P}^J_\phi(y_{t+1:T} )\right]\right) -    \E \left( \log\left[ \what{P}^M_\phi(y_{t+1:T})  \right] \right) &=^{(a)}  \E \left(\log\left[ \prod_{i=t+1}^T\what{P}_\phi(\hat{Y}_{i}=y_{i}|y_{1:i-1} )\right]\right) \\& \quad\quad\quad\quad\quad\quad -    \E \left(\log\left[ \prod_{i=t+1}^T \what{P}_\phi(\hat{Y}_{t+1} = y_{i}|y_{1:t})  \right] \right)
    \\ & =^{(b)}  \E \left(\log\left[ \prod_{i=t+1}^T\P({Y}_{i}=y_{i}|y_{1:i-1} )\right]\right)  \\& \quad\quad\quad\quad\quad\quad -    \E \left( \log\left[ \prod_{i=t+1}^T \P({Y}_{t+1} = y_{i}|y_{1:t})  \right] \right)
    \\ & = \E \left(\log\left[ \frac{\prod_{i=t+1}^T\P({Y}_{i}=y_{i}|y_{1:i-1} )}{\prod_{i=t+1}^T \P({Y}_{t+1} = y_{i}|y_{1:t})}\right]\right)
    \\ & = \E \left(\sum_{i=t+1}^T\log\left[ \frac{\P({Y}_{i}=y_{i}|y_{1:i-1} )}{ \P({Y}_{t+1} = y_{i}|y_{1:t})}\right]\right)
    \\ & =  \sum_{i=t+1}^T \E\left(\log\left[ \frac{\P({Y}_{i}=y_{i}|y_{1:i-1} )}{ \P({Y}_{t+1} = y_{i}|y_{1:t})}\right]\right)
    \\ & =^{(c)} \sum_{i=t+1}^T I(y_i, y_{t+1:i-1}|y_{1:t}) 
\end{align*}

Here (a) follows from the definition of $\what{P}^J_\phi$, $\what{P}^M_\phi$. (b) follows from the assumption $\what{P}_\phi = \P$. (c) follows from the following  identity
\[
I\bigl(X;Y \mid Z\bigr)
\;=\;
\mathbb{E}\!\Bigl[
\ln \frac{P(X,Y\mid Z)}{P(X\mid Z)\,P(Y\mid Z)}
\Bigr]
\;=\;
\mathbb{E}\!\Bigl[
\ln \frac{P(X\mid Y,Z)}{P(X\mid Z)}
\Bigr]
\;=\;
\mathbb{E}\!\Bigl[
\ln \frac{P(Y\mid X,Z)}{P(Y\mid Z)}
\Bigr].
\]

Also note that it is equal to 
$\E_{y_{1:t}\sim \P}\left(\dkl{\P(y_{t+1:T}|y_{1:t})}{\prod_{i=t+1}^T \P(y_{i}|y_{1:t})} \right)$

\end{proof}
\subsection{Proof of Example \ref{example:non_contextual_example}}

\begin{proof}

Recall that,
$
Y_i = \theta + \varepsilon_i,
\quad
\varepsilon_i \sim \mathcal{N}(0, \tau^2),
\quad
\theta \sim \mathcal{N}(\mu, \sigma^2).
$. We first estimate KL divergence
\[
D_{\mathrm{KL}}\Bigl(\,\P(y_{t+1:T}\mid y_{1:t})\;\Big\|\;\prod_{i=t+1}^T \P(y_i\mid y_{1:t})\Bigr).
\]

   Because \(Y_i \mid \theta \sim \mathcal{N}(\theta,\tau^2)\), therefore the posterior \(\P(\theta \mid y_{1:t})\) is normal with mean 
     $
       \mu_t 
       \;=\; 
       \frac{ 
         \frac{\mu}{\sigma^2}
         \;+\;\frac{1}{\tau^2}\sum_{i=1}^t y_i
       }{
         \frac{1}{\sigma^2} + \frac{t}{\tau^2}
       }$
       and variance is $
       \sigma_t^2 
       \;=\; 
       \Bigl(\tfrac{1}{\sigma^2} + \tfrac{t}{\tau^2}\Bigr)^{\!-1}.
     $
      Further conditional on \(\theta\), the future \(y_{t+1},\dots,y_T\) are i.i.d. \(\mathcal{N}(\theta,\tau^2)\).   Therefore we get,  
   $
   y_{t+1:T}\mid y_{1:t} \;\sim\; 
   \mathcal{N}\!\Bigl(\mu_t\,\mathbf{1},\;\tau^2\,I + \sigma_t^2\,\mathbf{1}\mathbf{1}^T\Bigr),
   $
   where \(\mathbf{1}\) is the \((T-t)\)-dimensional vector of all ones, and \(I\) the \((T-t)\times (T-t)\) identity.

For the one-step inference 
   $
   \prod_{i=t+1}^T \P(y_i\mid y_{1:t})
   $, each \(y_i\) has distribution
   $
   y_i\mid y_{1:t} \;\sim\;
   \mathcal{N}\bigl(\mu_t,\;\tau^2 + \sigma_t^2\bigr)
   $. Denote this covariance by \(\Sigma_Q = (\tau^2 + \sigma_t^2)\,I\).

   Now, we want the KL divergence between two \((T - t)\) dimensional Gaussians with \(P = \mathcal{N}\!\bigl(\mu_t\mathbf{1},\;\Sigma_P\bigr)\) with
  \(\Sigma_P \;=\; \tau^2 I \;+\; \sigma_t^2\,\mathbf{1}\mathbf{1}^T\).
and  \(Q = \mathcal{N}\!\bigl(\mu_t\mathbf{1},\;\Sigma_Q\bigr)\) with
  \(\Sigma_Q \;=\; (\tau^2 + \sigma_t^2)\,I.\)
For two \(K\)-dimensional Gaussians \(\mathcal{N}(\mu_0,\Sigma_0)\) and \(\mathcal{N}(\mu_1,\Sigma_1)\), the KL divergence is

\[
D_{\mathrm{KL}}\bigl(\mathcal{N}(\mu_0,\Sigma_0)\,\|\,\mathcal{N}(\mu_1,\Sigma_1)\bigr)
=
\tfrac12
\Bigl[
  \mathrm{tr}\bigl(\Sigma_1^{-1}\Sigma_0\bigr)
  \;+\;(\mu_1 - \mu_0)^T \Sigma_1^{-1}\,(\mu_1 - \mu_0)
  \;-\;K
  \;+\;\ln\!\frac{|\Sigma_1|}{|\Sigma_0|}
\Bigr].
\]

Therefore we get - 

\[
\begin{aligned}
D_{\mathrm{KL}}(P\|Q)
&=\;
\tfrac12
\Bigl[
  K\ln\!\Bigl(1 + \frac{\sigma_t^2}{\tau^2}\Bigr)
  \;-\;\ln\!\Bigl(1 + K\,\frac{\sigma_t^2}{\tau^2}\Bigr)
\Bigr].
\end{aligned}
\]

As the expression is independent of $y_{1:t}$, taking expectation over $y_{1:t}$ our final expression remains the same. 







\end{proof}

\subsection{Generalization of Theorem \ref{thm:uq_one_step_suffers_non_contextual} to the contextual setting}

\noindent\textbf{Loss of information in one-step v/s multi-step:} Let $y_{t+1:T} \sim \P(\cdot|y_{1:t},x_{1:T})$ be some data from true data generating process. Further, let $\what{P}^M_\phi(y_{t+1:T}, x_{t+1:T})   \equiv \prod_{i=t+1}^T\what{P}_\phi(\hat{Y}_{t+1}=y_{i}|y_{1:t},x_{1:t}, x_{i})$ and $\what{P}_\phi^J(y_{t+1:T}, x_{t+1:T}) \equiv   \prod_{i=t+1}^T\what{P}_\phi(\hat{Y}_i =y_i|y_{1:i-1},x_{1:i-1},x_i)$. 

\begin{theorem}  Assume {\small $\what{P}_\phi = \P$}. The  difference  {\small $\E \left(\log\left[ \what{P}^J_\phi(y_{t+1:T} , x_{t+1:T})\right] - \what{P}^M_\phi(y_{t+1:T}, x_{t+1:T})  \right)$} is 
 \vspace{-3mm}
    \begin{align}
     \sum_{i=t+1}^TI(y_{i};y_{t+1:i-1} |y_{1:t},x_{1:i})
        \label{expression:one-multi-step-compare}
    \end{align}  
     \vspace{-3mm}
    where expectation is $y_{1:T}\sim \P(\cdot|x_{1:T})$ and $x_{1:T}\simiid P_X$.
  \label{thm:uq_one_step_suffers} 
\end{theorem}

\subsubsection{Proof of Theorem \ref{thm:uq_one_step_suffers}}

 Recall that  $y_{t+1:T} \sim \P(\cdot|y_{1:t},x_{1:T})$ is generated  from the true data generating process. Further, $\what{P}^M_\phi(y_{t+1:T}, x_{t+1:T})   \equiv \prod_{i=t+1}^T\what{P}_\phi(\hat{Y}_{t+1}=y_{i}|y_{1:t},x_{1:t}, x_{i})$ and $\what{P}_\phi^J(y_{t+1:T}, x_{t+1:T}) \equiv   \prod_{i=t+1}^T\what{P}_\phi(\hat{Y}_i =y_i|y_{1:i-1},x_{1:i-1},x_i)$. 

\noindent\textbf{Theorem \ref{thm:uq_one_step_suffers}:} Assuming $\what{P}_\phi = \P$, then the  difference  $\E \left(\log\left[ \what{P}^J_\phi(y_{t+1:T} , x_{t+1:T})\right]\right) -  \E \left(\what{P}^M_\phi(y_{t+1:T}, x_{t+1:T})  \right)$ is equal to 
    \begin{align*}
     \sum_{i=t+1}^TI(y_{i};y_{t+1:i-1} |y_{1:t},x_{1:i})
    \end{align*}   
    where expectation is $y_{1:T}\sim \P(\cdot|x_{1:T})$ and $x_{1:T}\simiid P_X$.

\begin{proof}
We can follow exactly same procedure as in the proof of Theorem \ref{thm:uq_one_step_suffers_non_contextual} for this proof. 
    
\end{proof}

\subsection{Characterizing (\ref{expression:one-multi-step-compare}) for Bayesian linear regression and Gaussian processes }

\begin{myexample} 
\textbf{Bayesian Linear Regression -} Assuming $Y=\theta^T X+\epsilon$ where $\epsilon \sim N(0,\tau^2)$ and $\theta \sim N(\mu, \Sigma)$. Inputs $X$ are drawn i.i.d. from $P_X$.  Let $\mathcal{D}_t := (x_{1:t}, y_{1:t})$ and the posterior $\theta| \mathcal{D}_t \sim N(\mu',\Sigma')$, then  expression \ref{expression:one-multi-step-compare}
is equal to
\[ \frac{1}{2}  \E_{\mathbf{X},\Sigma'}\left[ \log \frac{|\text{diag}(\mathbf{X} \Sigma' \mathbf{X}^T)|}{|\mathbf{X} \Sigma' \mathbf{X}^T|} \right]
\]
where $\mathbf{X}$ is the matrix of ${x_{t+1},x_{t+2}\cdots, x_T}$.  
\label{example:blr}
\end{myexample}

\begin{myexample} 
\textbf{Gaussian Processes -} Assuming $Y=f(X)+\epsilon$ where  $f \sim \mathcal{GP}(m,\mc{K}) $, where $m(X)$ is mean and $\mc{K}(X,X')$ is the covariance. Additionally,   Gaussian   noise $N(0,\sigma^2)$ is added to the outputs.  The input $X$ is drawn i.i.d. from {$P_X$}. 
Let  $\mathcal{D}_t := (x_{1:t}, y_{1:t})$. Suppose, under multi-step inference  $
  P(\mathbf{y}_{t+1:T} \mid \mathbf{y}_{1:t}, \mathbf{X}_{1:T}) = \mathcal{N}(\mathbf{y}_{t+1:T} \mid \boldsymbol{\mu}_P, \mathbf{K}_P)
  $.  Further, under single-step inference 
  $
  Q(\mathbf{y}_{t+1:T}) = \prod_{i=t+1}^T \mathcal{N}(y_i \mid \mu_i, \sigma_i^2) = \mathcal{N}(\mathbf{y}_{t+1:T} \mid \boldsymbol{\mu}_Q, \mathbf{\Sigma}_Q)
  $. Then,  expression \ref{expression:one-multi-step-compare}
is equal to
\[
 \frac{1}{2} \E_{\mathbf{\Sigma}_P,\mathbf{\Sigma}_Q, \mathbf{\mu}_Q, \mathbf{\mu}_P} \left[ \log \left( \frac{\mathbf{\Sigma}_Q}{|\mathbf{K}_P|} \right) - (T - t) + \text{tr}\left( \mathbf{K}_P^{-1} \mathbf{\Sigma}_Q \right) + (\boldsymbol{\mu}_Q - \boldsymbol{\mu}_P)^\top \mathbf{K}_P^{-1} (\boldsymbol{\mu}_Q - \boldsymbol{\mu}_P) \right]
\]
where  \( d = T - t \))

and the posterior $\theta| \mathcal{D}_t \sim N(\mu',\Sigma')$, then  expression \ref{expression:one-multi-step-compare}
is equal to
\[  \frac{1}{2} \E_{\mathbf{X},\Sigma'} \left[ \log \frac{|\text{diag}(\mathbf{X} \Sigma' \mathbf{X}^T)|}{|\mathbf{X} \Sigma' \mathbf{X}^T|} \right]
\]
where $\mathbf{X}$ is the matrix of ${x_{t+1},x_{t+2}\cdots, x_T}$.
\label{example:gp}
\end{myexample}
\subsubsection{Proof of Example \ref{example:blr} and \ref{example:gp}}

\begin{proof}

 Recall that, under multi-step inference  $
  P(\mathbf{y}_{t+1:T} \mid \mathbf{y}_{1:t}, \mathbf{X}_{1:T}) = \mathcal{N}(\mathbf{y}_{t+1:T} \mid \boldsymbol{\mu}_P, \mathbf{K}_P)
  $.  Further, under single-step inference 
  $
  Q(\mathbf{y}_{t+1:T}) = \prod_{i=t+1}^T \mathcal{N}(y_i \mid \mu_i, \sigma_i^2) = \mathcal{N}(\mathbf{y}_{t+1:T} \mid \boldsymbol{\mu}_Q, \mathbf{\Sigma}_Q)
  $. Now,  KL Divergence between two multivariate Gaussian distributions:
\( \mathcal{N}(\mathbf{y} \mid \boldsymbol{\mu}_P, \mathbf{K}_P) \)
 and \( Q = \mathcal{N}(\mathbf{y} \mid \boldsymbol{\mu}_Q, \mathbf{\Sigma}_Q) \) is give by

\[
\dkl{P}{Q} = \frac{1}{2} \left[ \log \left( \frac{|\mathbf{\Sigma}_Q|}{|\mathbf{K}_P|} \right) - d + \text{tr}\left( \mathbf{K}_P^{-1} \mathbf{\Sigma}_Q \right) + (\boldsymbol{\mu}_Q - \boldsymbol{\mu}_P)^\top \mathbf{K}_P^{-1} (\boldsymbol{\mu}_Q - \boldsymbol{\mu}_P) \right]
\]

where: \( d \) is the dimensionality of \( \mathbf{y} \) (i.e., \( d = T - t \)); \( |\cdot| \) denotes the determinant and \( \text{tr}(\cdot) \) denotes the trace of a matrix. Therefore, we get 
\[
\dkl{P}{Q} = \frac{1}{2} \left[ \log \left( \frac{\prod_{i=t+1}^T \sigma_i^2}{|\mathbf{K}_P|} \right) - (T - t) + \text{tr}\left( \mathbf{K}_P^{-1} \mathbf{\Sigma}_Q \right) + (\boldsymbol{\mu}_Q - \boldsymbol{\mu}_P)^\top \mathbf{K}_P^{-1} (\boldsymbol{\mu}_Q - \boldsymbol{\mu}_P) \right]
\]

In \textbf{Bayesian linear regression} the means remain the same, i.e., \( \boldsymbol{\mu}_Q = \boldsymbol{\mu}_P \), further the expression reduces to the following:

\[
\dkl{ \mathcal{N}(\mathbf{X} \mu', \mathbf{X} \Sigma' \mathbf{X}^T + \tau^2 \mathbf{I})}{ \mathcal{N}(\mathbf{X} \mu', \text{diag}(\mathbf{X} \Sigma' \mathbf{X}^T) + \tau^2 \mathbf{I}) }
\]

that is,

\[
\dkl{P}{Q} = \frac{1}{2} \left[ \log \frac{|\mathbf{X} \Sigma' \mathbf{X}^T|}{|\text{diag}(\mathbf{X} \Sigma' \mathbf{X}^T)|} \right]
\]

Final expressions follows from proof of Theorem \ref{thm:uq_one_step_suffers_non_contextual} in Section \ref{sec:proof_of_theorem_2}.

\end{proof}

\subsection{Proof of Theorem \ref{thm:bandit_theorem}}
Recall that our setting was such that the reward of first arm is generated as  $Y^{(1)} \sim N(\theta, \tau^2)$ with $\theta \sim N(\mu,\sigma^2)$. While the second arm has a constant reward  $Y^{(2)} = 0$.

\noindent\textbf{Theorem \ref{thm:bandit_theorem}:} There exists  one-armed bandit scenarios in which Thompson sampling incurs \( O(T) \) Bayesian regret if  it relies solely on one-step predictions from autoregressive sequence models.

\begin{proof}
  Consider a bandit setting, where $\sigma =0$ and $\mu<0$. Implementing Thompson sampling using the one-step inference, does not differentiate between aleatoric and epistemic uncertainty. This means whenever we will sample $Y>0$ we will choose the wrong arm to pull as $\mu \text{(mean reward from arm 1)}<0 \text{(mean reward from Arm 2)}$.  As $Y\sim N(\mu, \tau^2)$ therefore, $P(Y>0) =  \Phi\left(\frac{\mu}{\tau}\right)$. Hence, on an average we will suffer  $\left(T\Phi\left(\frac{\mu}{\tau}\right)\right)$ regret over horizon $T$. 
  
  Similarly, when $\sigma =0$ and $\mu>0$, whenever we will sample $Y<0$ we will choose the wrong arm to pull. Therefore in this case, it suffers $T\Phi \left(\frac{- \mu}{\tau}\right) $ Bayesian regret over horizon $T$.  On the other hand, the multi-step inference (assuming it is done till $\infty$) will have $0$ regret in these two  case.

\end{proof}








\section{Additional Experiments}
\label{sec:additional_experiments}

In this section, we present ablation studies to evaluate the performance of different architectures, specifically comparing conditionally permutation-invariant and standard causal masking schemes. In Section \ref{sec:marginal_log_lik_dim_1}, we provide the corresponding results from Section \ref{sec:architectural_experiments} using marginal log-likelihood as the evaluation metric. Section \ref{sec:ablation_dimension} explores the effect of varying the input dimension $X$, while Section \ref{sec:ablate_noise} examines the impact of different noise levels in the observed output 
$Y$. 

\subsection{One-step log-loss figures corresponding to Section \ref{sec:architectural_experiments}}
\label{sec:marginal_log_lik_dim_1}

\begin{figure*}[t]
\centering
\begin{minipage}[b]{0.48\textwidth}
\centering \includegraphics[width=0.9\textwidth, height=5cm]{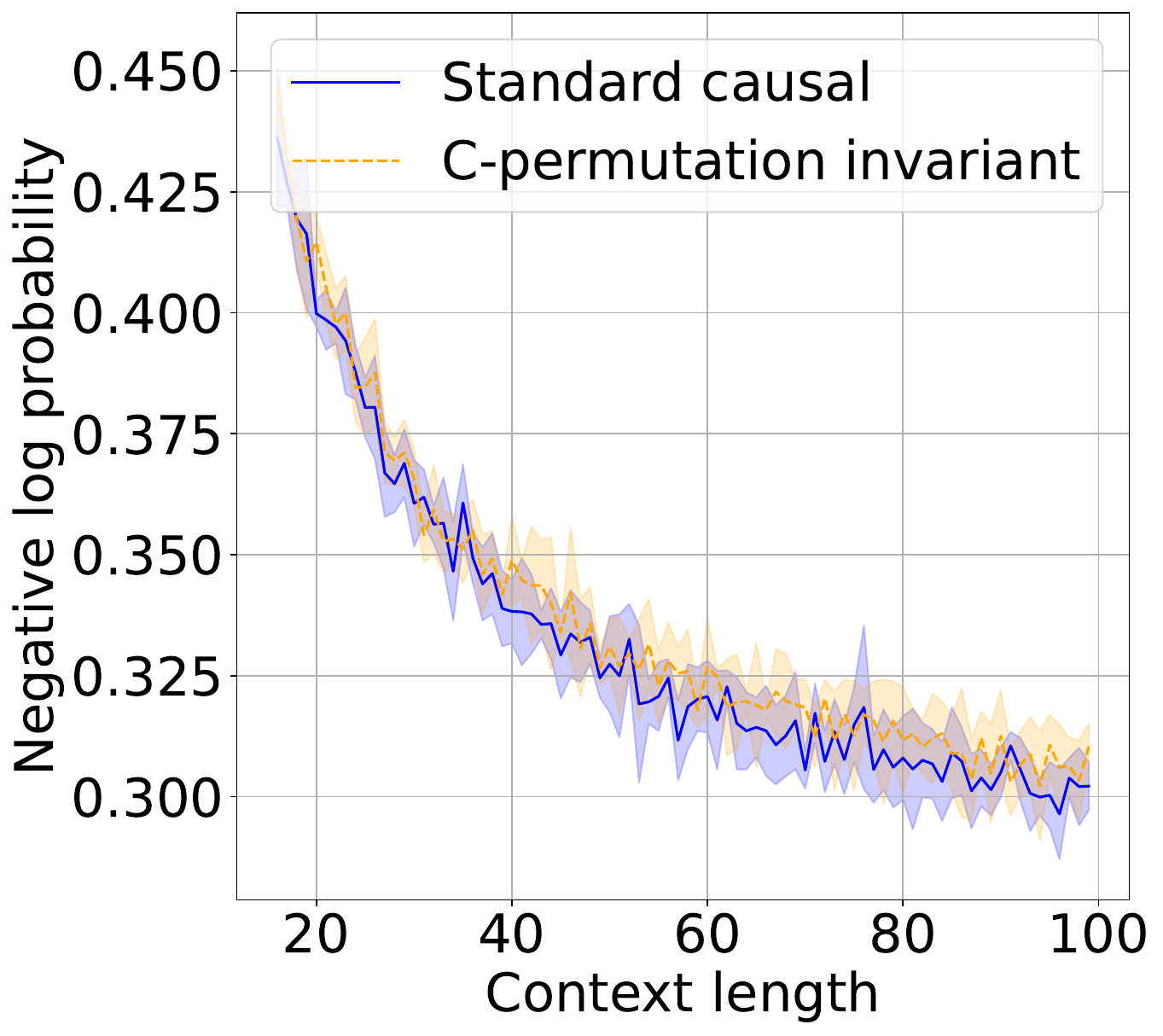}
  \caption*{\textbf{(a)} In-training horizon performance }
\end{minipage}
\hfill
\begin{minipage}[b]{0.48\textwidth}
\centering \includegraphics[width=0.9\textwidth, height=5cm]
{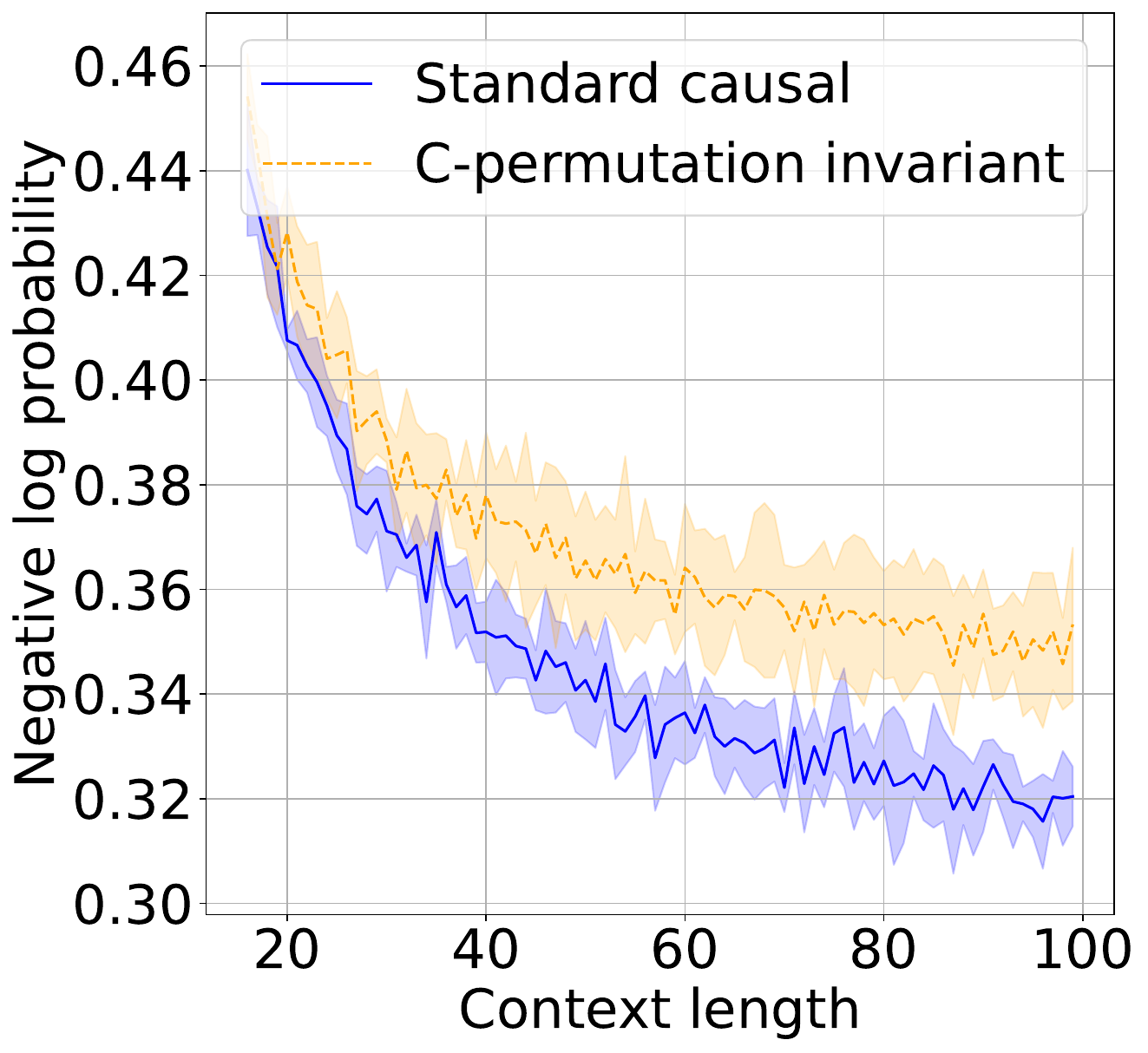}
\caption*{\textbf{(b)} Out-of-training horizon performance}
\end{minipage}
\caption{\textbf{One-step log-loss:} Comparing the two architectures on one-step log-loss metric [Dimension: 1].}
\label{fig:one_step_log_loss}
\end{figure*}

In Figures \ref{fig:one_step_log_loss}\textcolor{blue}{(a)} and \ref{fig:one_step_log_loss}\textcolor{blue}{(b)}, we present the in-training horizon and out-of-training horizon performance for the corresponding settings shown in Figures \ref{fig:comparing_two_architectures_main_diagram}\textcolor{blue}{(a)} and \ref{fig:comparing_two_architectures_main_diagram}\textcolor{blue}{(c)}. The results exhibit similar trends, where both masking schemes perform equally well in-distribution. However, beyond the training horizon, the causal masking approach appears to outperform the conditionally permutation-invariant masking.

\subsection{Ablations on dimensions}
\label{sec:ablation_dimension}
In this section, we present the results of our ablation study on the dimension $d$ of the context, where $X\sim U[-2,2]^d$.

\begin{figure}[t]
\centering
\begin{minipage}[b]{0.32\textwidth}
\centering \includegraphics[width=0.9\textwidth, height=4cm]{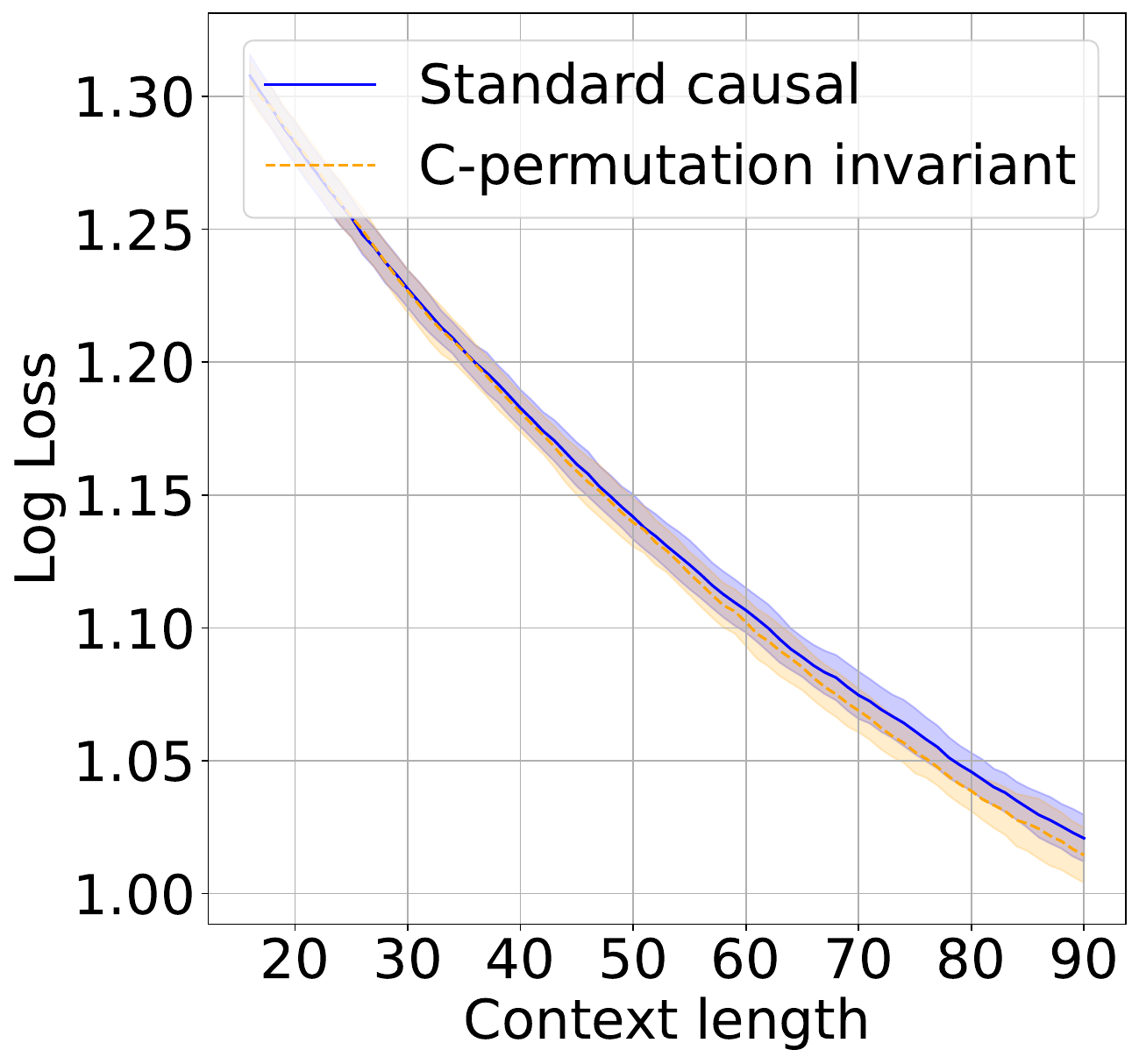}
  \caption*{\textbf{(a)}  Dimension: 4}
\end{minipage}
\hfill
\begin{minipage}[b]{0.32\textwidth}
\centering \includegraphics[width=0.9\textwidth, height=4cm]{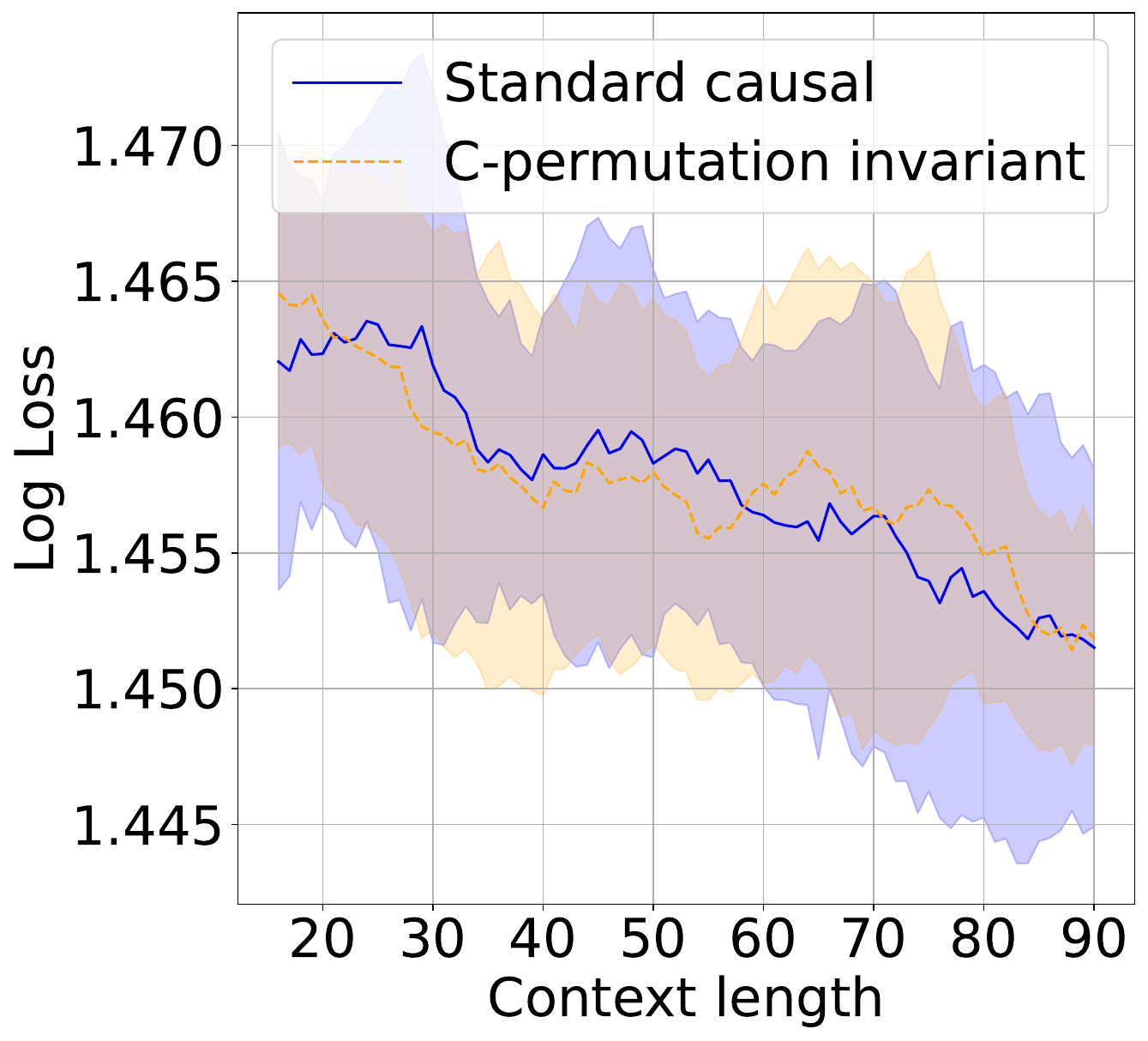}
  \caption*{\textbf{(b)}  Dimension: 8}
\end{minipage}
\hfill
\begin{minipage}[b]{0.32\textwidth}
\centering \includegraphics[width=0.9\textwidth, height=4cm]
{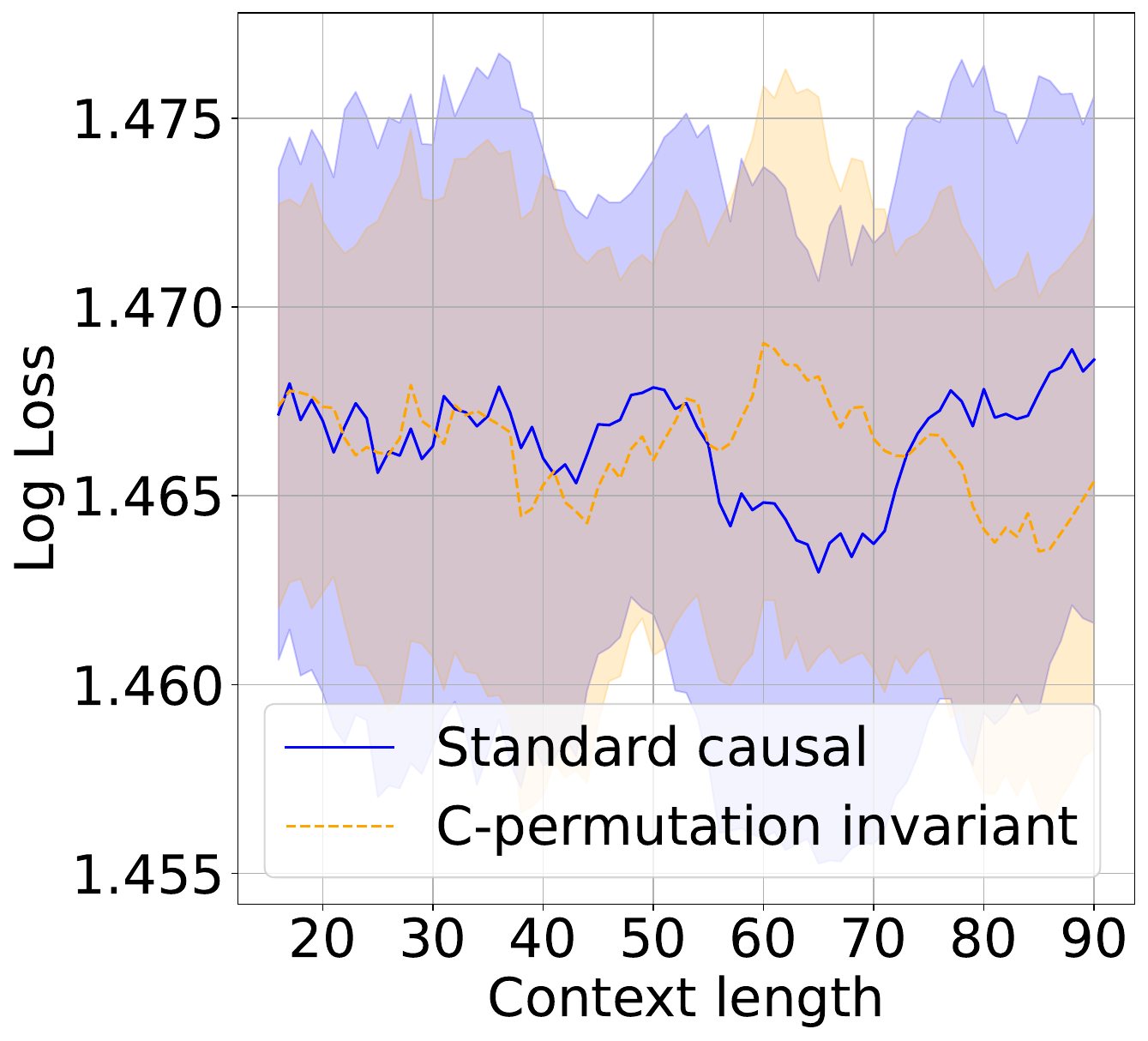}
  \caption*{\textbf{(c)} Dimension: 16}
\end{minipage}
\caption{\textbf{Ablation on dimension (In-training horizon performance):} Comparing two architectures [Training horizon: 100, Metric: Multi-step log-loss, Target length: 10].}
\label{fig:dimension_ablation_in_training}
\end{figure}

\noindent\textit{In-training horizon performance:} Figure \ref{fig:dimension_ablation_in_training} shows the in-training horizon performance for both architectures across different dimensions. Our findings align with previous observations. Notably, for $d=16$, the provided context appears insufficient to improve prediction log-loss.

\begin{figure}[t]
\centering
\begin{minipage}[b]{0.32\textwidth}
\centering \includegraphics[width=0.9\textwidth,height=4cm] {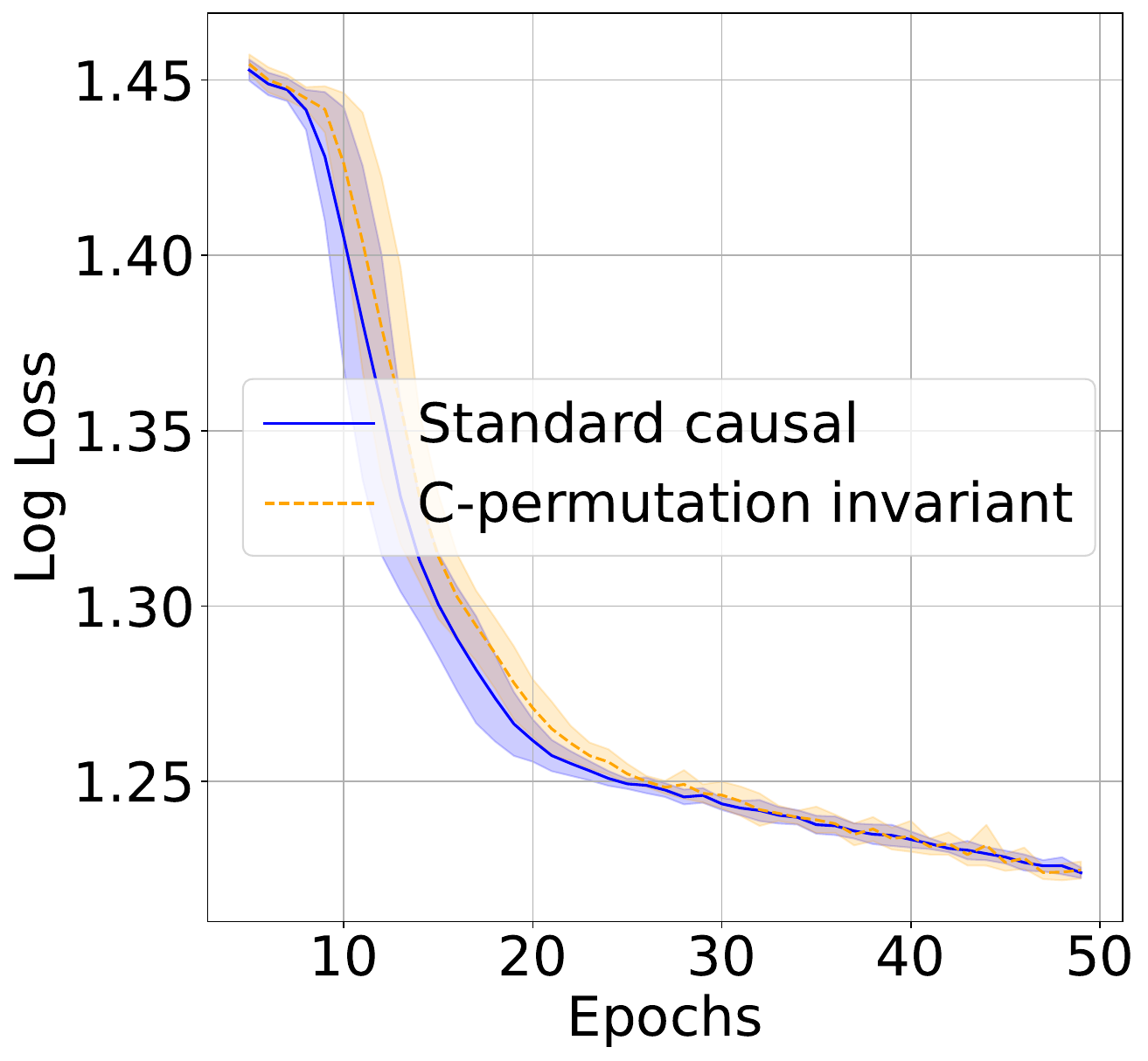}
  \caption*{\textbf{(a)} Dimension: 4}
\end{minipage}
\hfill
\begin{minipage}[b]{0.32\textwidth}
\centering \includegraphics[width=0.9\textwidth,height=4cm] {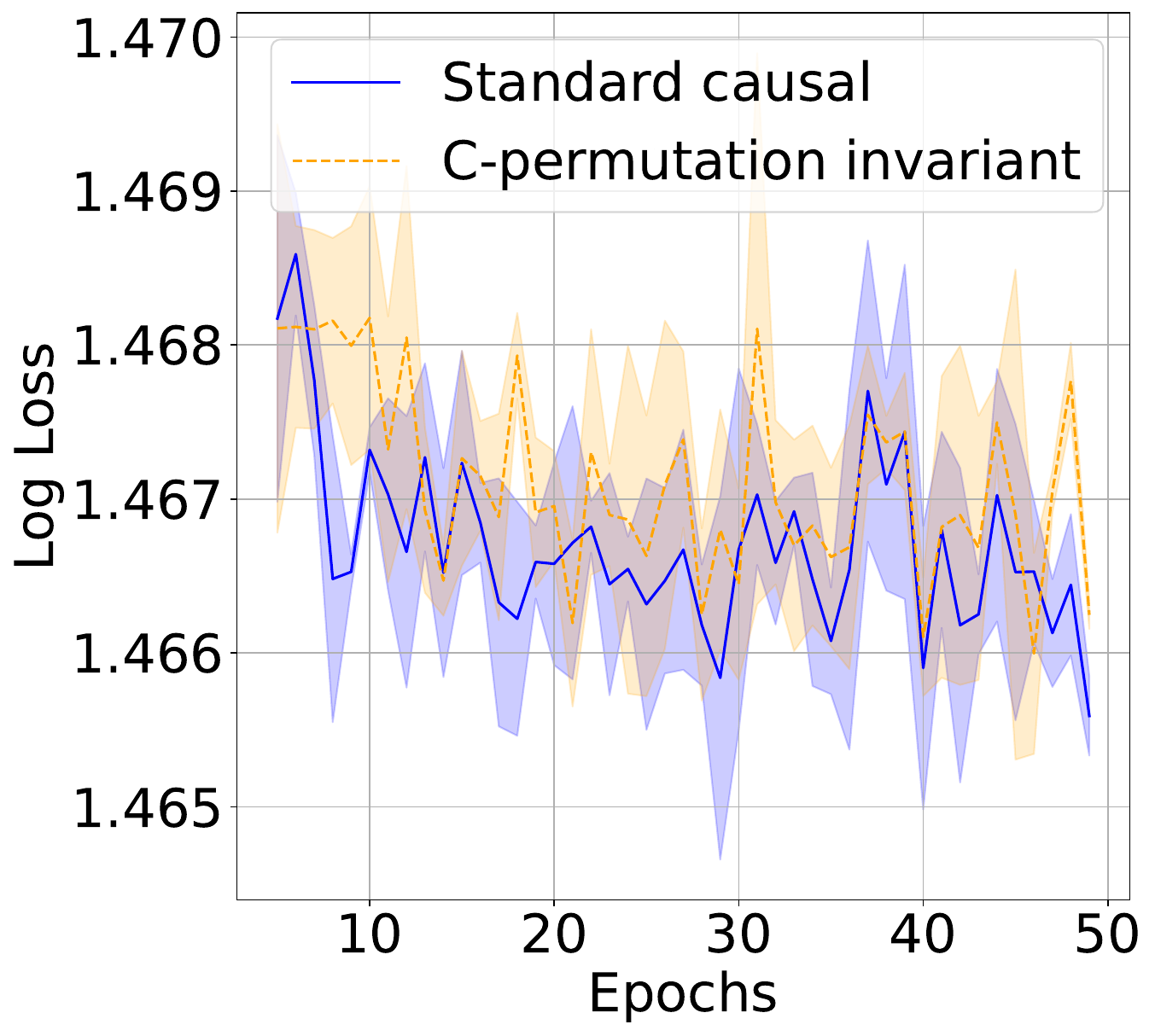}
  \caption*{\textbf{(b)}  Dimension: 8}
\end{minipage}
\hfill
\begin{minipage}[b]{0.32\textwidth}
\centering \includegraphics[width=0.9\textwidth,height=4cm] {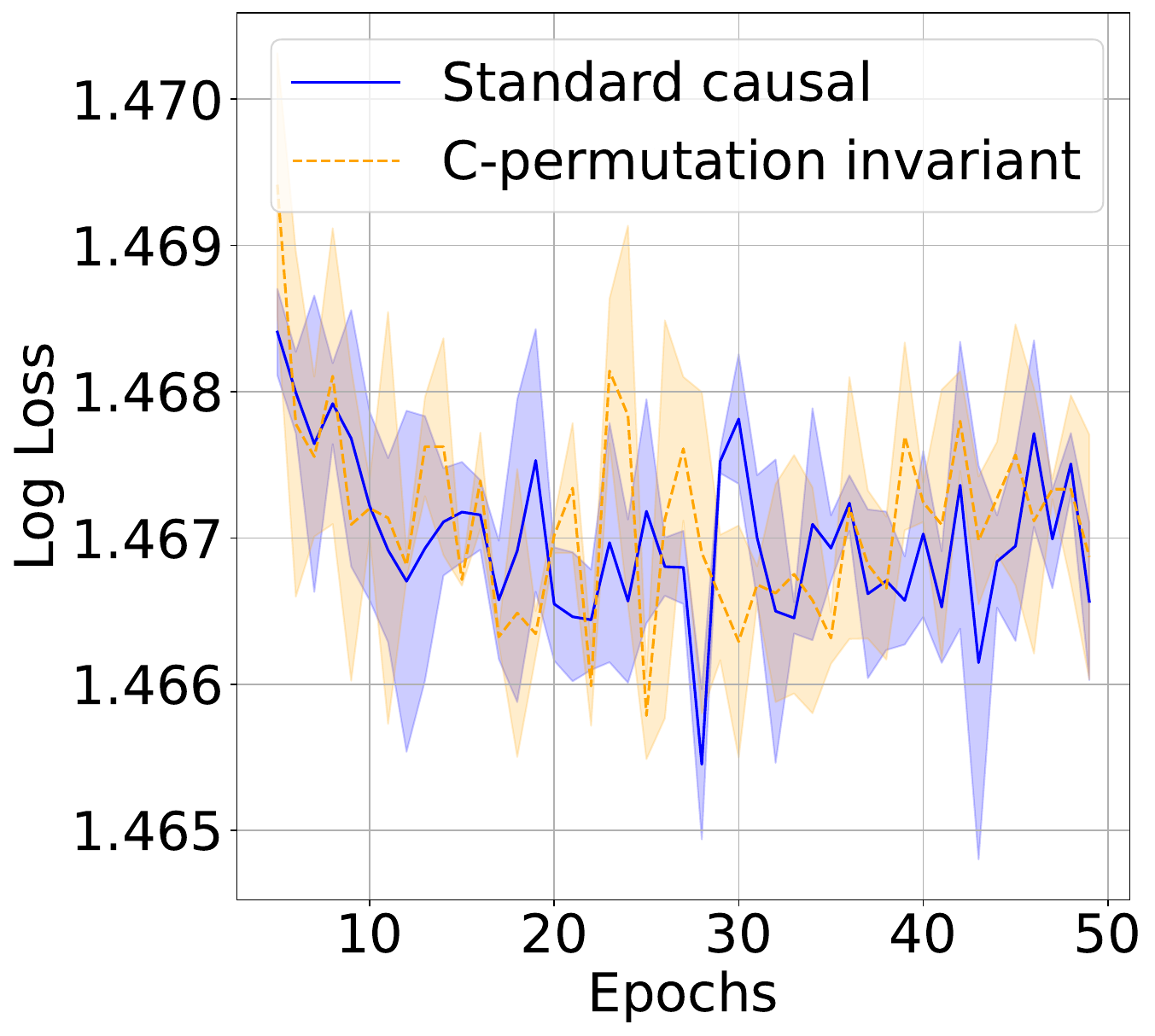}
  \caption*{\textbf{(c)} Dimension: 16}
\end{minipage}
\caption{\textbf{Ablation on dimension (Training/Data efficiency):} Comparing two architectures [Training horizon: 100, Metric: Multi-step log-loss, Target length: 10].}
\label{fig:dimension_ablation_data_efficiency}
\end{figure}

\noindent\textit{Training/Data efficiency:}
Figure \ref{fig:dimension_ablation_data_efficiency} illustrates the training and data efficiency across different dimensions.

\begin{figure}[t]
\centering
\begin{minipage}[b]{0.48\textwidth}
\centering \includegraphics[width=0.9\textwidth, height=5cm]
{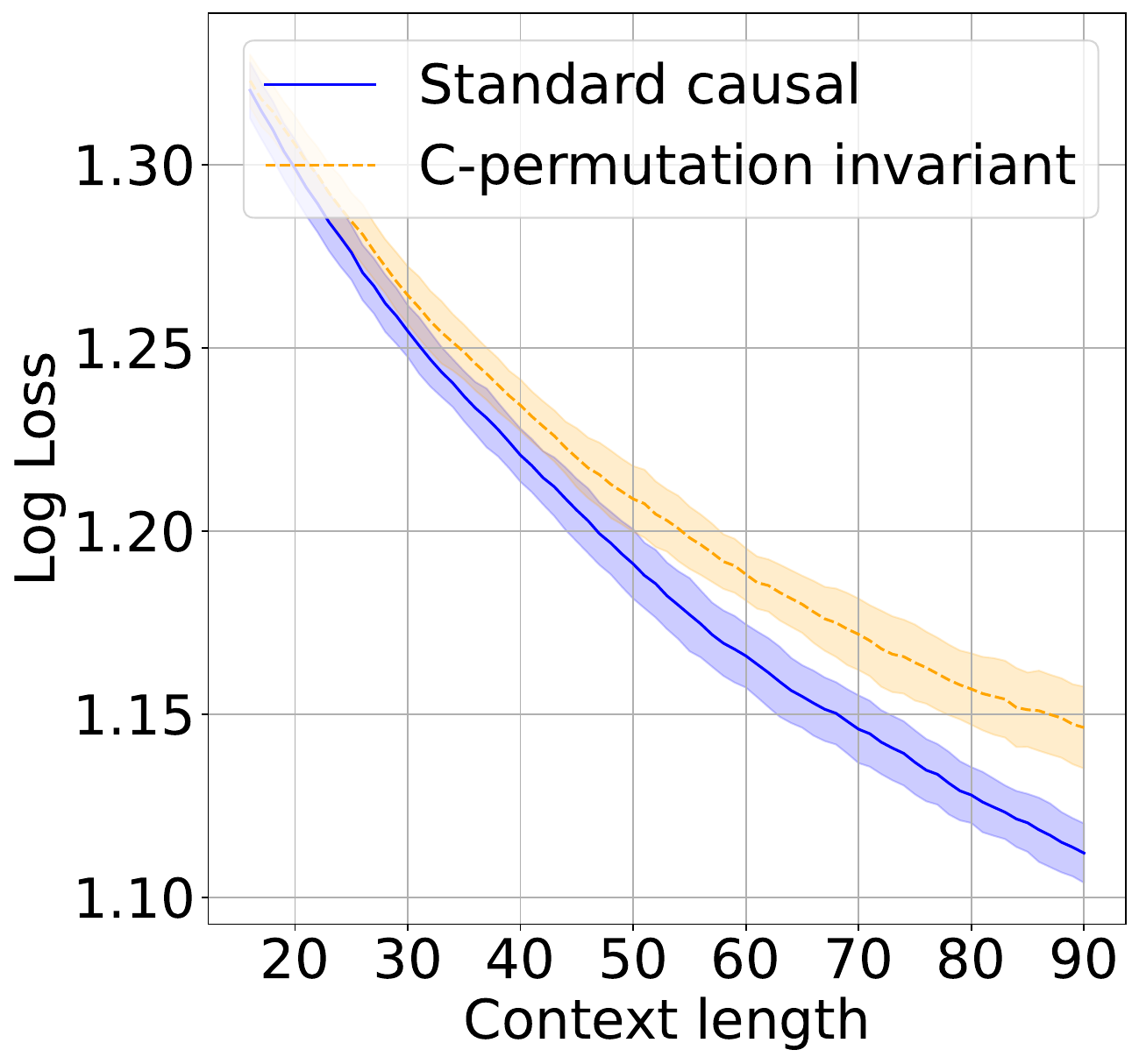}
\caption*{\textbf{(a)}  Dimension: 4, Multi-step log-loss}
\label{fig:ood_performance_dim_4_joint}
\end{minipage}
\hfill
\begin{minipage}[b]{0.48\textwidth}
\centering \includegraphics[width=0.9\textwidth, height=5cm]
{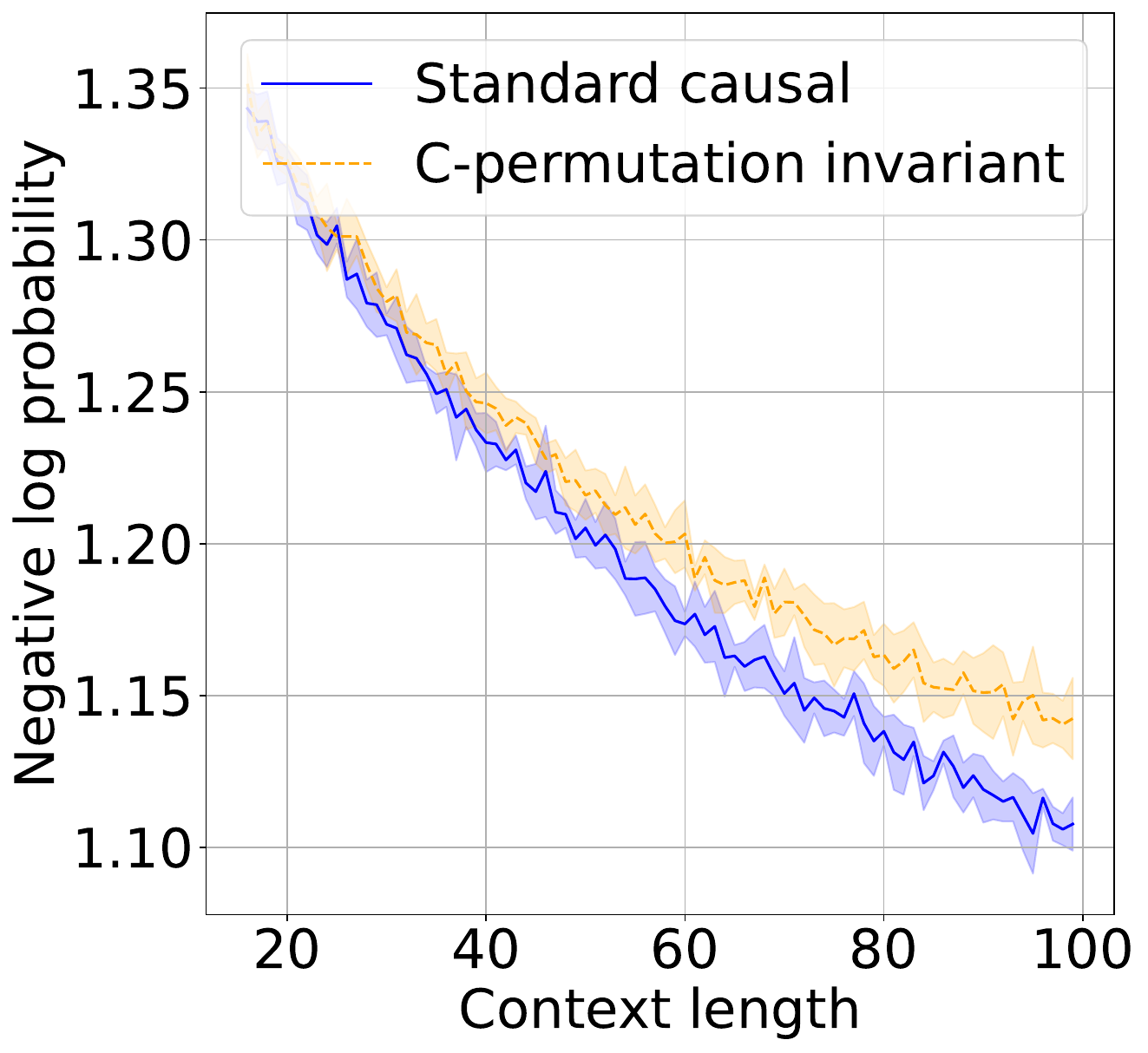}
\caption*{\textbf{(b)}  Dimension: 4,  One-step log-loss}
\end{minipage}
\caption{\textbf{Ablation on dimension (Out-of-training horizon performance):} Comparing two architectures [Training horizon: 15, Target length: 10].}
\label{fig:ood_performance_dim_4}
\end{figure}

\noindent\textit{Out-of-training horizon performance:} We only analyze out-of-training efficieny for the dimension 4 in addition to dimension 1. As shown in Figure \ref{fig:ood_performance_dim_4}, the results remain consistent with our earlier findings.

\begin{figure}[t]
\centering
\begin{minipage}[b]{0.32\textwidth}
\centering \includegraphics[width=0.9\textwidth,height=4cm]
{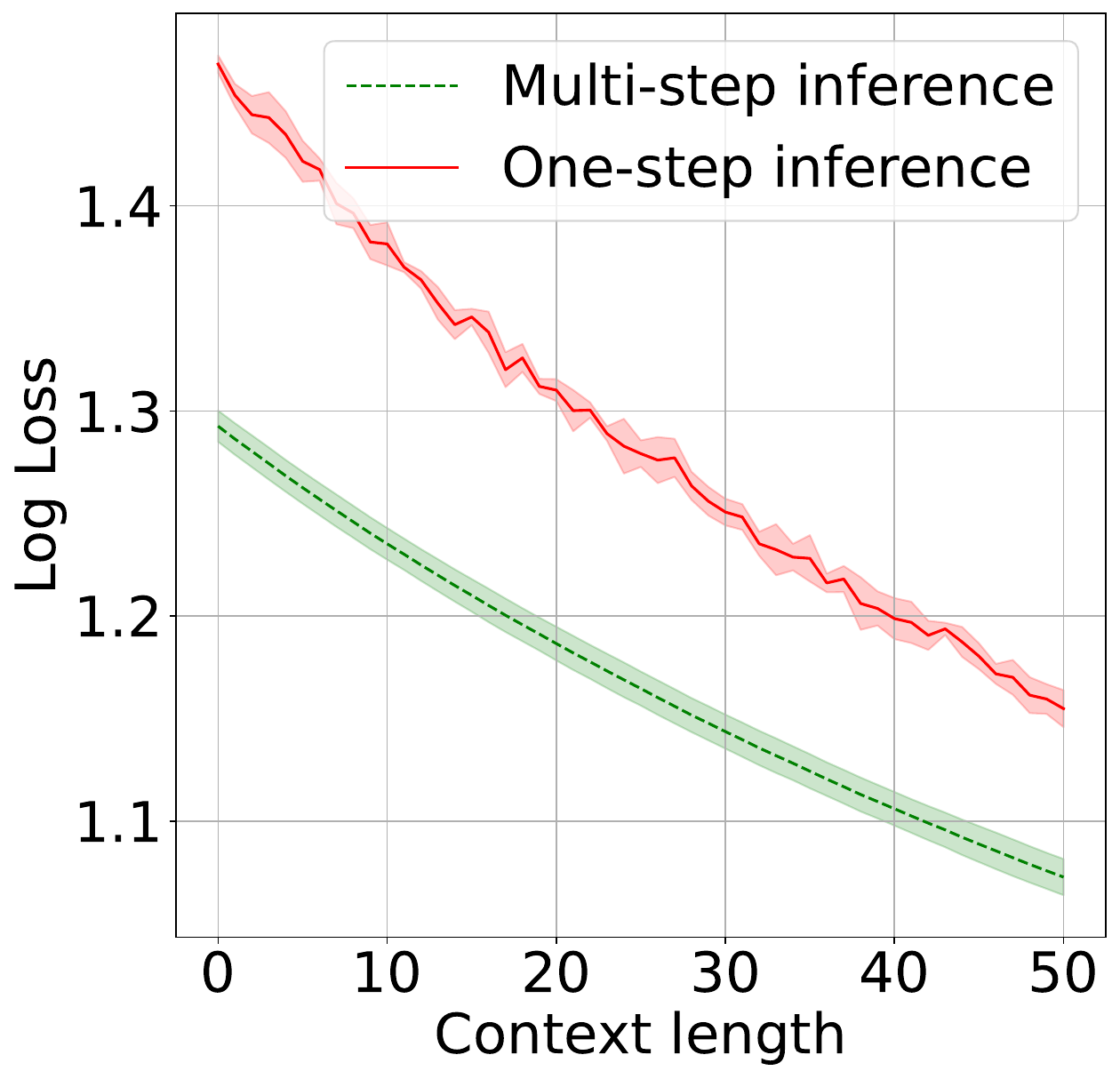}
  \caption*{\textbf{(a)} Dimension: 4}
\end{minipage}
\hfill
\begin{minipage}[b]{0.32\textwidth}
\centering \includegraphics[width=0.9\textwidth,height=4cm] {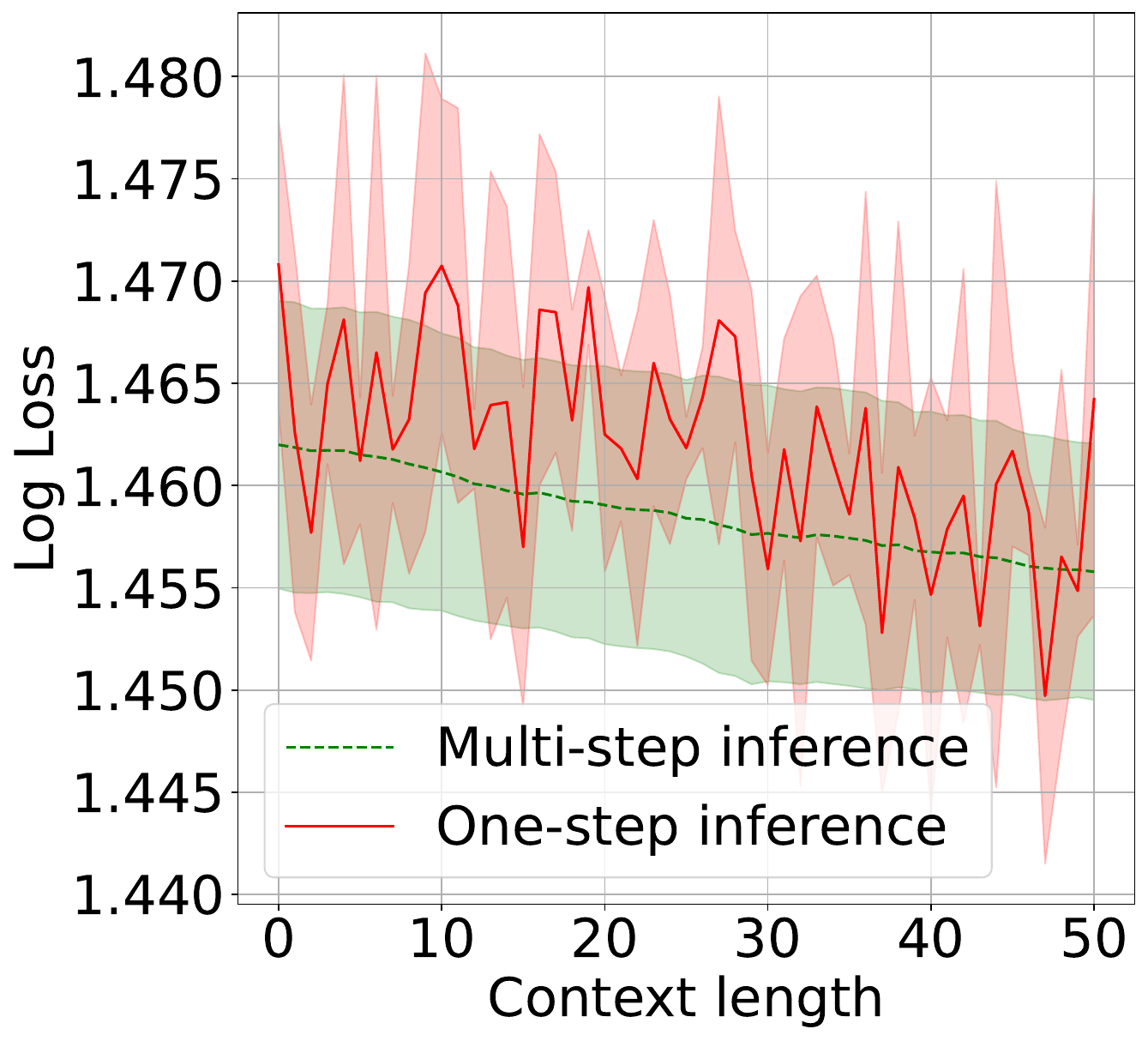}
  \caption*{\textbf{(b)} Dimension:8}
\end{minipage}
\hfill
\begin{minipage}[b]{0.32\textwidth}
\centering \includegraphics[width=0.9\textwidth,height=4cm] {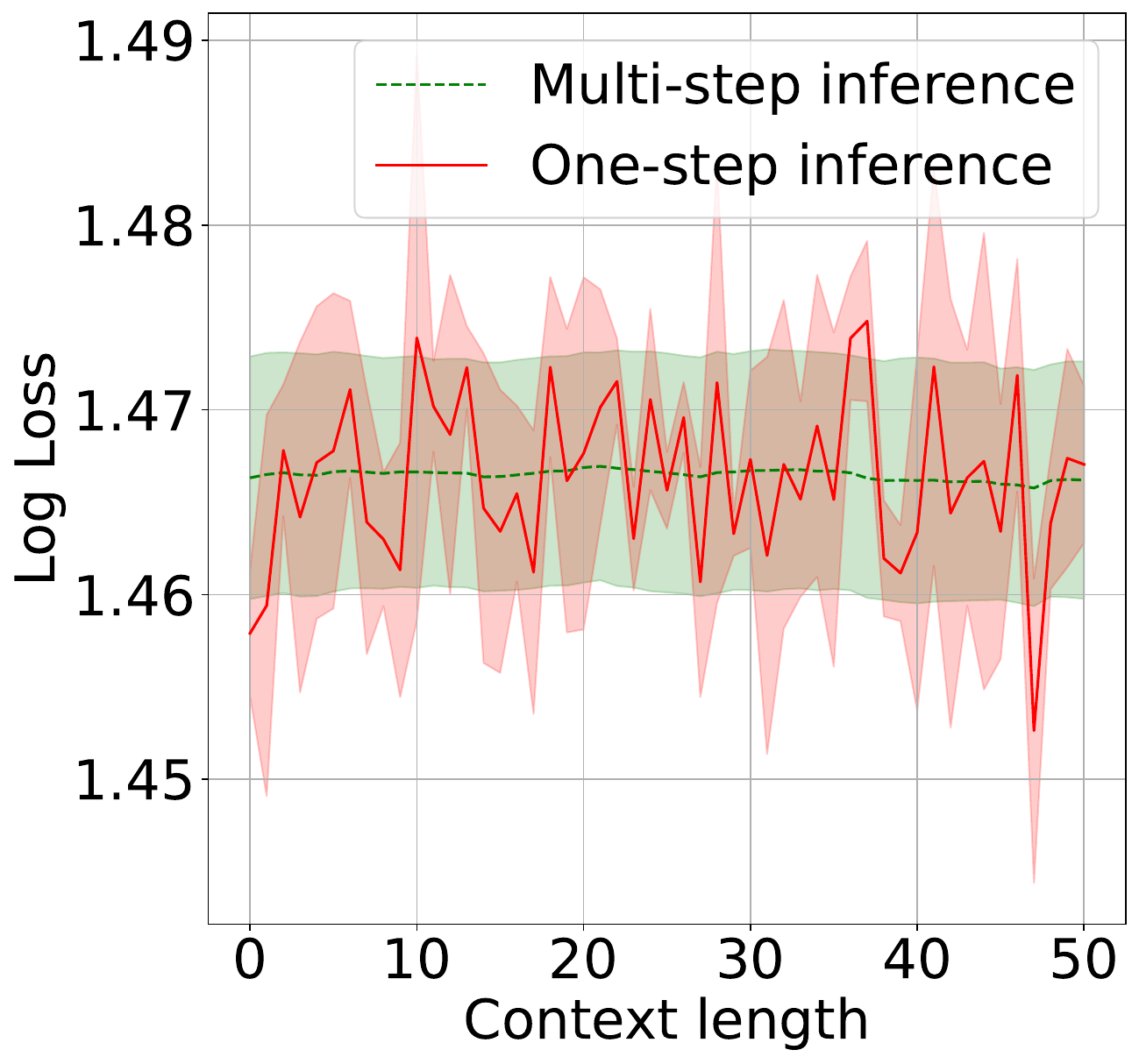}
  \caption*{\textbf{(c)} Dimension: 16}
\end{minipage}
\caption{\textbf{Ablation on dimension (Uncertainty Quantification):} Comparing one-step inference and multi-step inference [Train horizon: 100, Metric: Multi-step log-loss, Target Length: $50$].}
\label{fig:ablation_dim_one_multi}
\end{figure}

\noindent\textit{Multi-step v/s One-step inference:} Figure \ref{fig:ablation_dim_one_multi} present a comparison of multi-step and one-step inference across different dimensions.

\subsection{Ablations on noise}
\label{sec:ablate_noise}

\begin{figure*}[t]
\centering
\begin{minipage}[b]{0.32\textwidth}
\centering \includegraphics[width=0.9\textwidth, height=4cm]{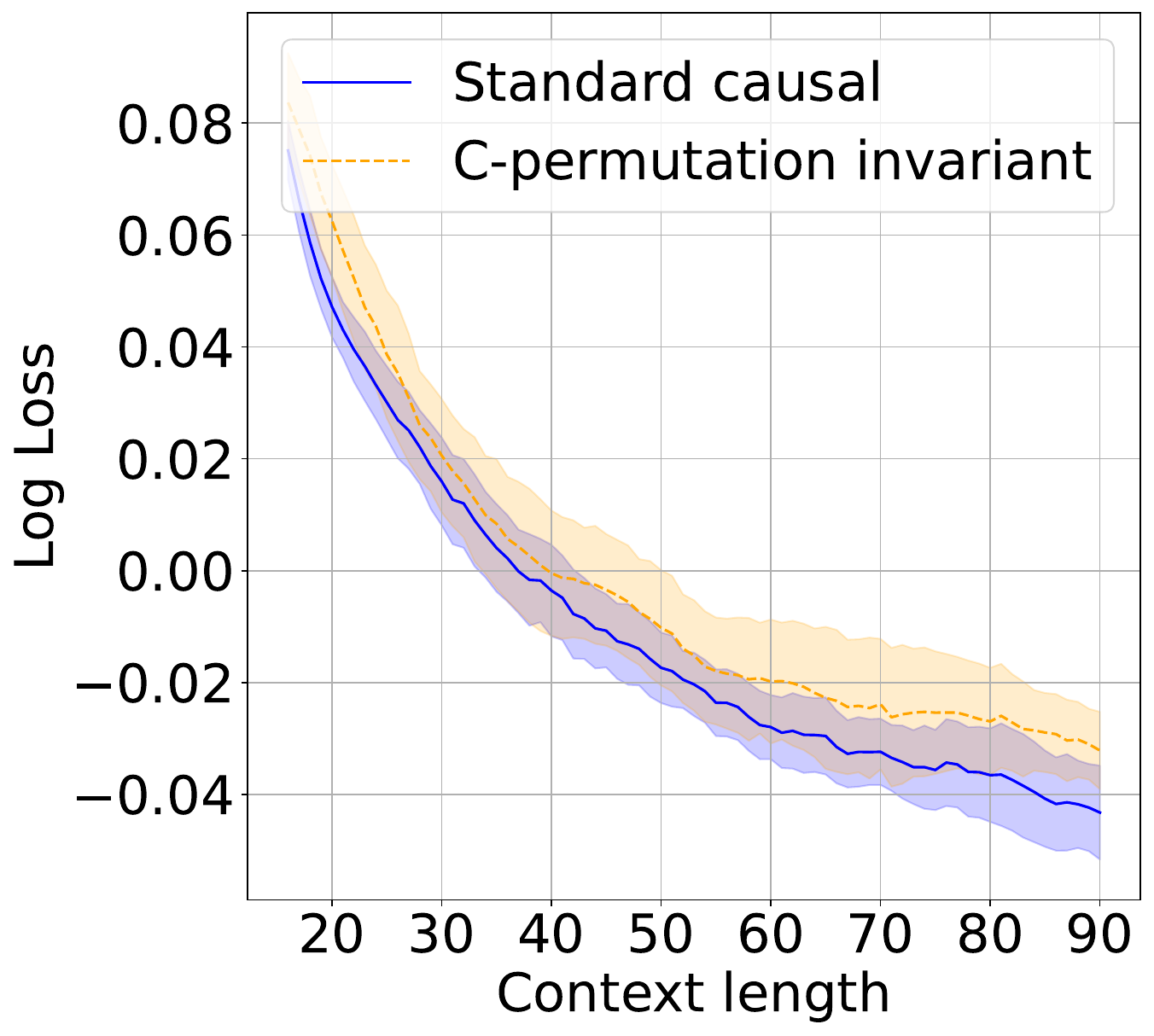}
  \caption*{\textbf{(a)} Noise: 0.05}
\label{fig:jll-ablation-dim-1-epoch-400-noise-0.05}
\end{minipage}
\hfill
\begin{minipage}[b]{0.32\textwidth}
\centering \includegraphics[width=0.9\textwidth, height=4cm]{figures/figures_ablation_error_bars/joint_x_axis_context_autoreg_exchg_Horizon_100_epoch_400_dim_1_target_len_10_ood_16_noise_0.1.pdf}
 \caption*{\textbf{(b)} Noise: 0.1}
\label{fig:jll-ablation-dim-1-epoch-400-noise-0.1-here}
\end{minipage}
\hfill
\begin{minipage}[b]{0.32\textwidth}
\centering \includegraphics[width=0.9\textwidth, height=4cm]
{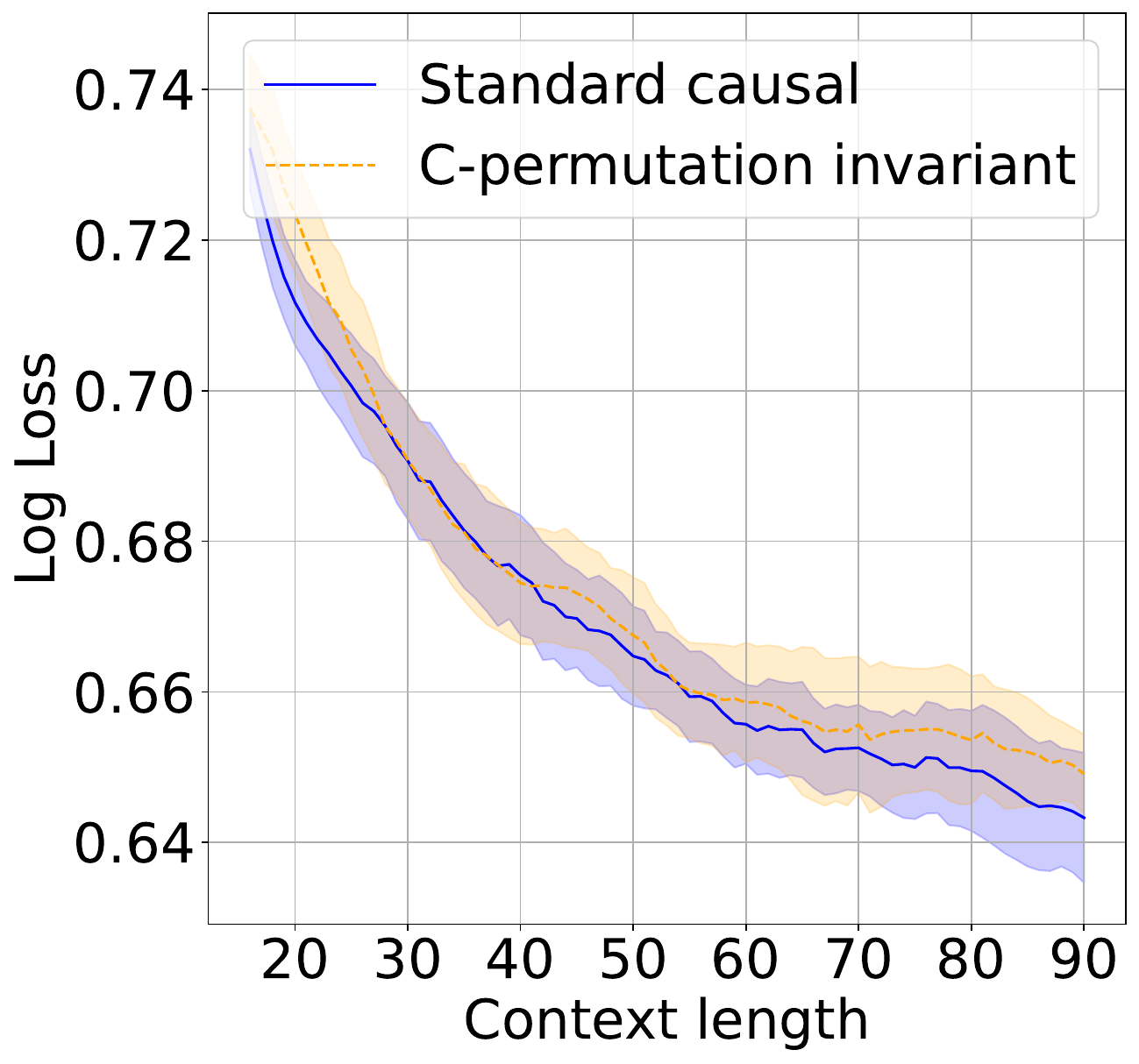}
  \caption*{\textbf{(c)} Noise: 0.2}
\label{fig:jll-ablation-dim-1-epoch-400-noise-0.1-this}
\end{minipage}
\hfill
\vspace{10mm}

\begin{minipage}[b]{0.32\textwidth}
\centering \includegraphics[width=0.9\textwidth, height=4cm]{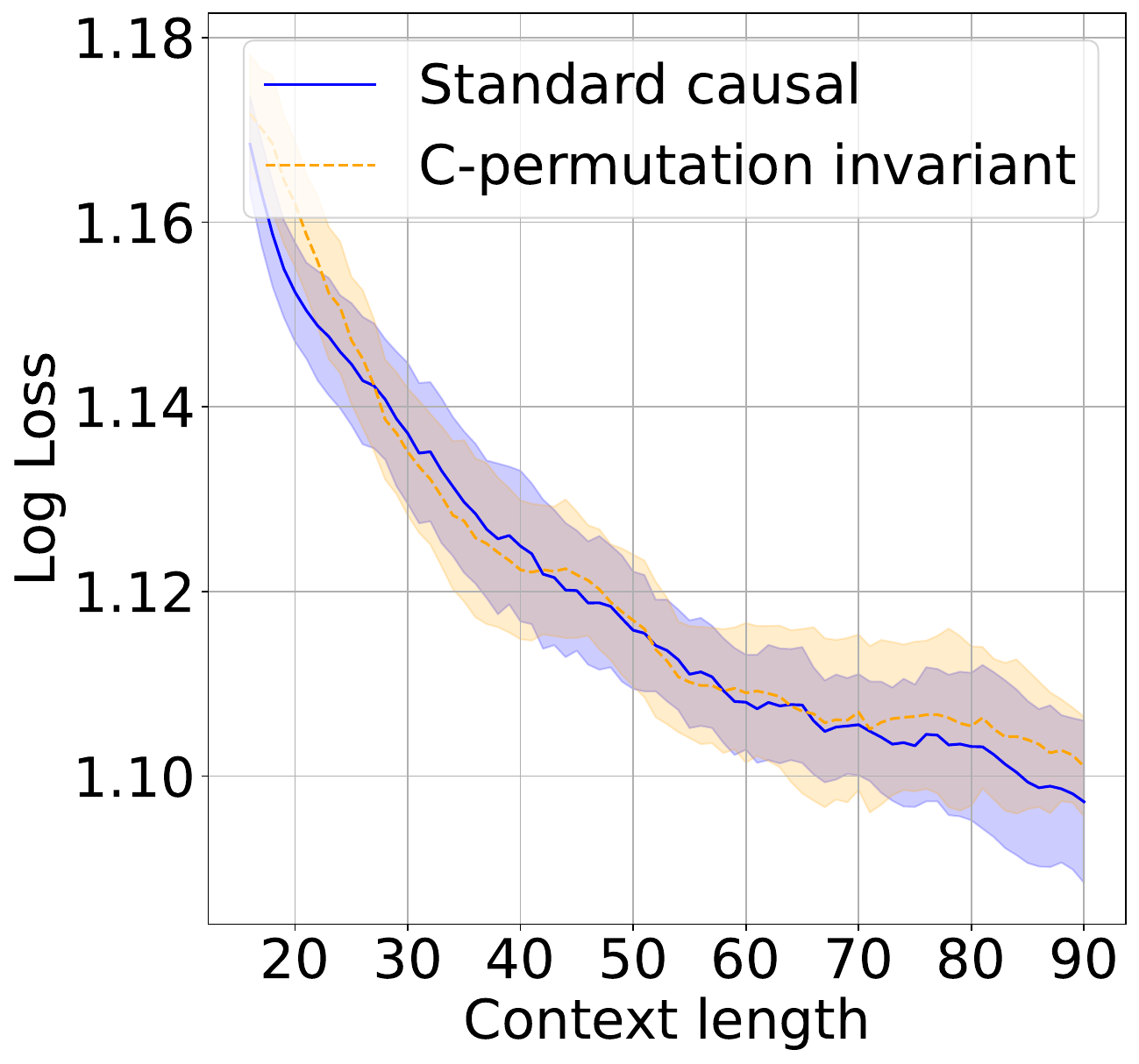}
  \caption*{\textbf{(d)} Noise: 0.5}
\label{fig:jll-ablation-dim-4-epoch-400-noise-0.5}
\end{minipage}
\hfill
\begin{minipage}[b]{0.32\textwidth}
\centering \includegraphics[width=0.9\textwidth, height=4cm]{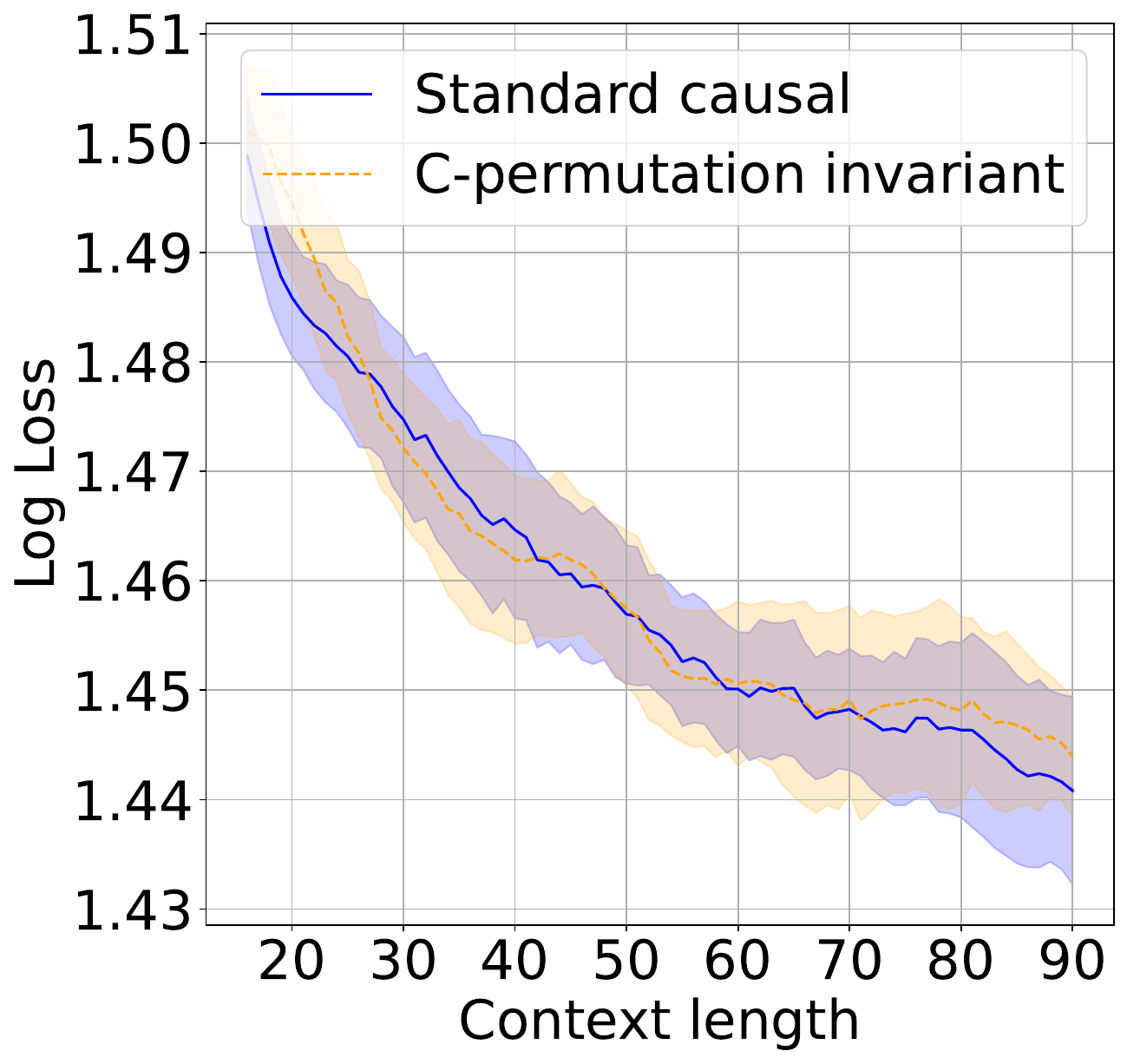}
  \caption*{\textbf{(e)} Noise: 1.0}
\label{fig:jll-ablation-dim-1-epoch-400-noise-1.0-again}
\end{minipage}
\hfill
\begin{minipage}[b]{0.32\textwidth}
\centering \includegraphics[width=0.9\textwidth, height=4cm]
{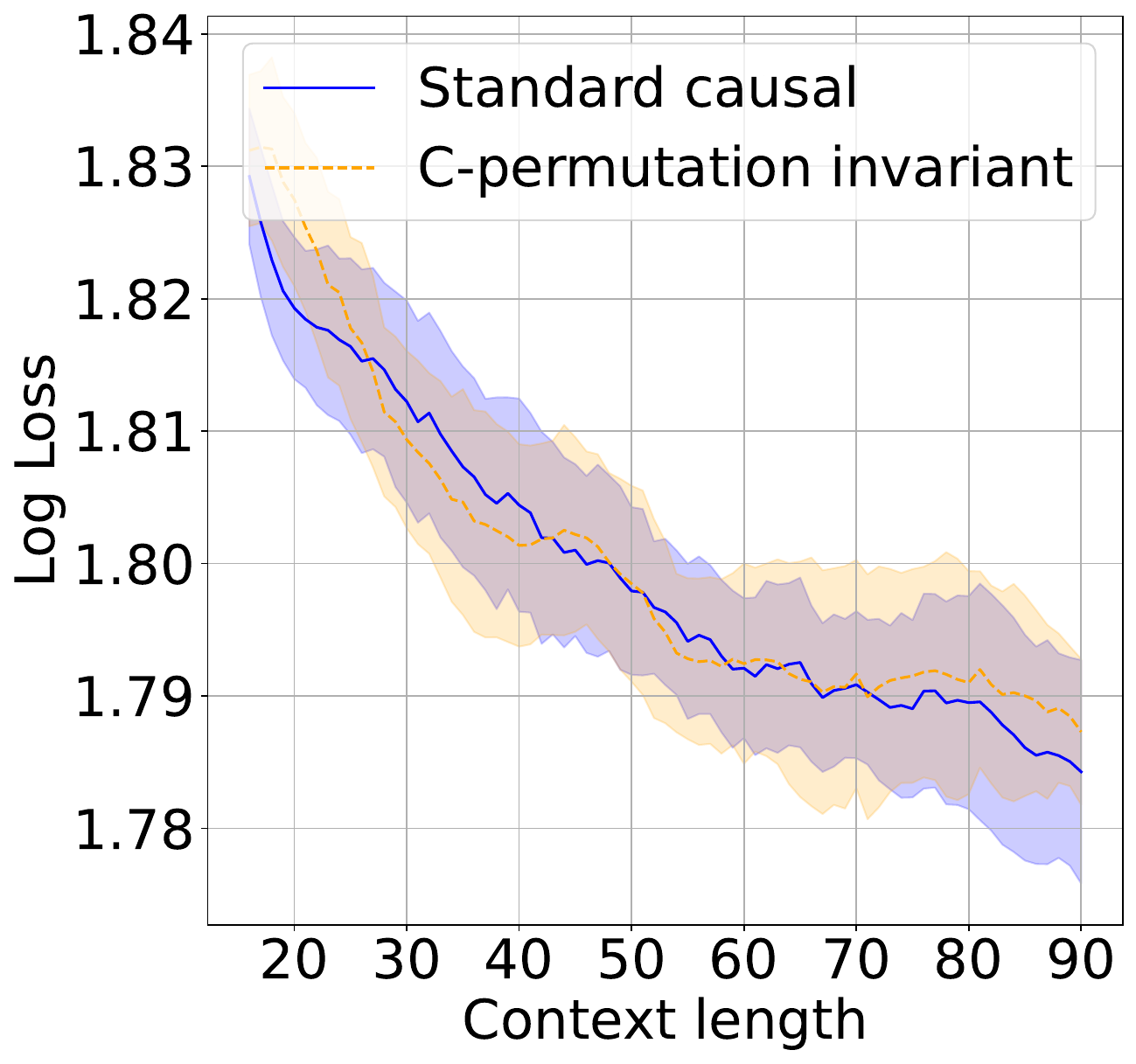}
  \caption*{\textbf{(f)} Noise: 2.0}
\label{fig:jll-ablation-dim-1-epoch-400-noise-2.0-again}
\end{minipage}
 \caption{\textbf{Ablation on noise (In-training horizon performance):} Comparing two architectures [Training horizon: 100, Metric: Multi-step log-loss, Target length: 10].}
 \label{fig:ablation_noise_in_training}
\end{figure*}

\noindent\textit{In-training horizon performance:}
Figures \ref{fig:ablation_noise_in_training} illustrates the in-training horizon performance for both architectures under varying levels of observation noise. Our findings remain consistent with previous observations.

\begin{figure*}[t]
\centering
\begin{minipage}[b]{0.32\textwidth}
\centering \includegraphics[width=0.9\textwidth,height=4cm] {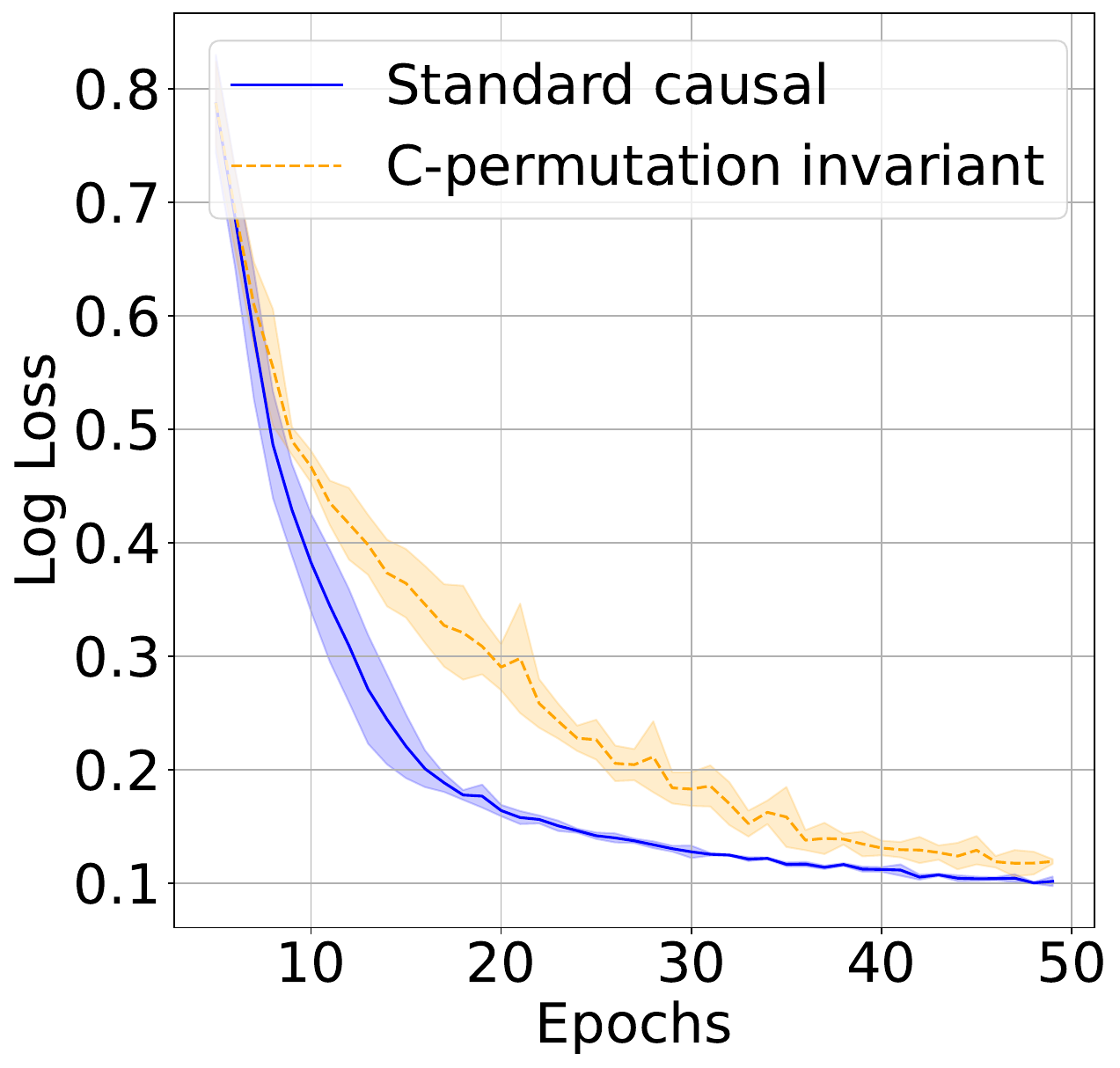}
  \caption*{\textbf{(a)} Noise: 0.05}
\end{minipage}
\hfill
\begin{minipage}[b]{0.32\textwidth}
\centering \includegraphics[width=0.9\textwidth,height=4cm] {figures/figures_ablation_error_bars/X_1_Y_1_Horizon_100_Noise_0_1.pdf}
  \caption*{\textbf{(b)} Noise: 0.1}
\end{minipage}
\hfill
\begin{minipage}[b]{0.32\textwidth}
\centering \includegraphics[width=0.9\textwidth,height=4cm] {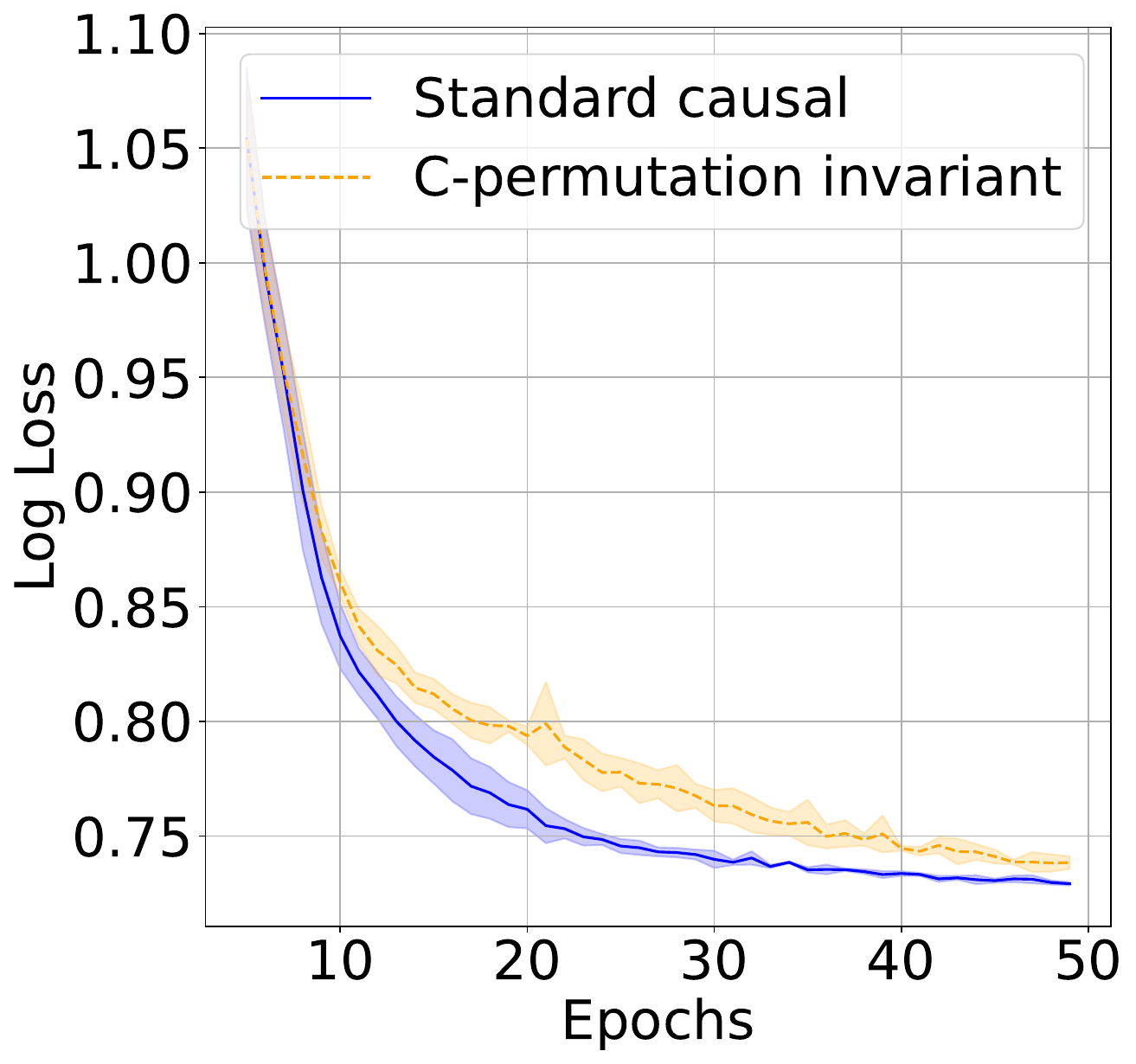}
  \caption*{\textbf{(c)} Noise: 0.2}
\end{minipage}
\hfill
\vspace{10mm}

\begin{minipage}[b]{0.32\textwidth}
\centering \includegraphics[width=0.9\textwidth,height=4cm] {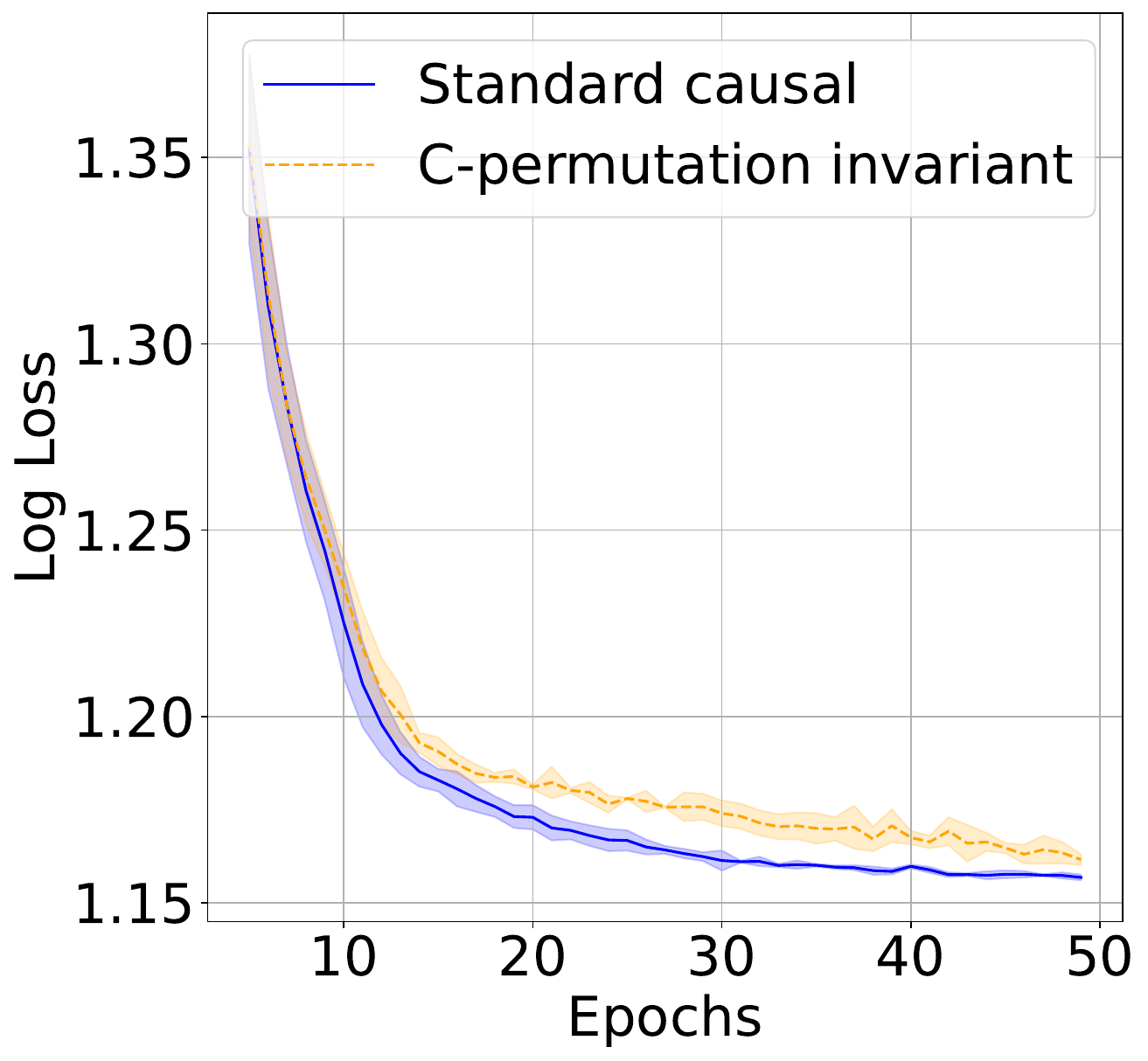}
  \caption*{\textbf{(d)} Noise: 0.5}
\end{minipage}
\hfill
\begin{minipage}[b]{0.32\textwidth}
\centering \includegraphics[width=0.9\textwidth,height=4cm] {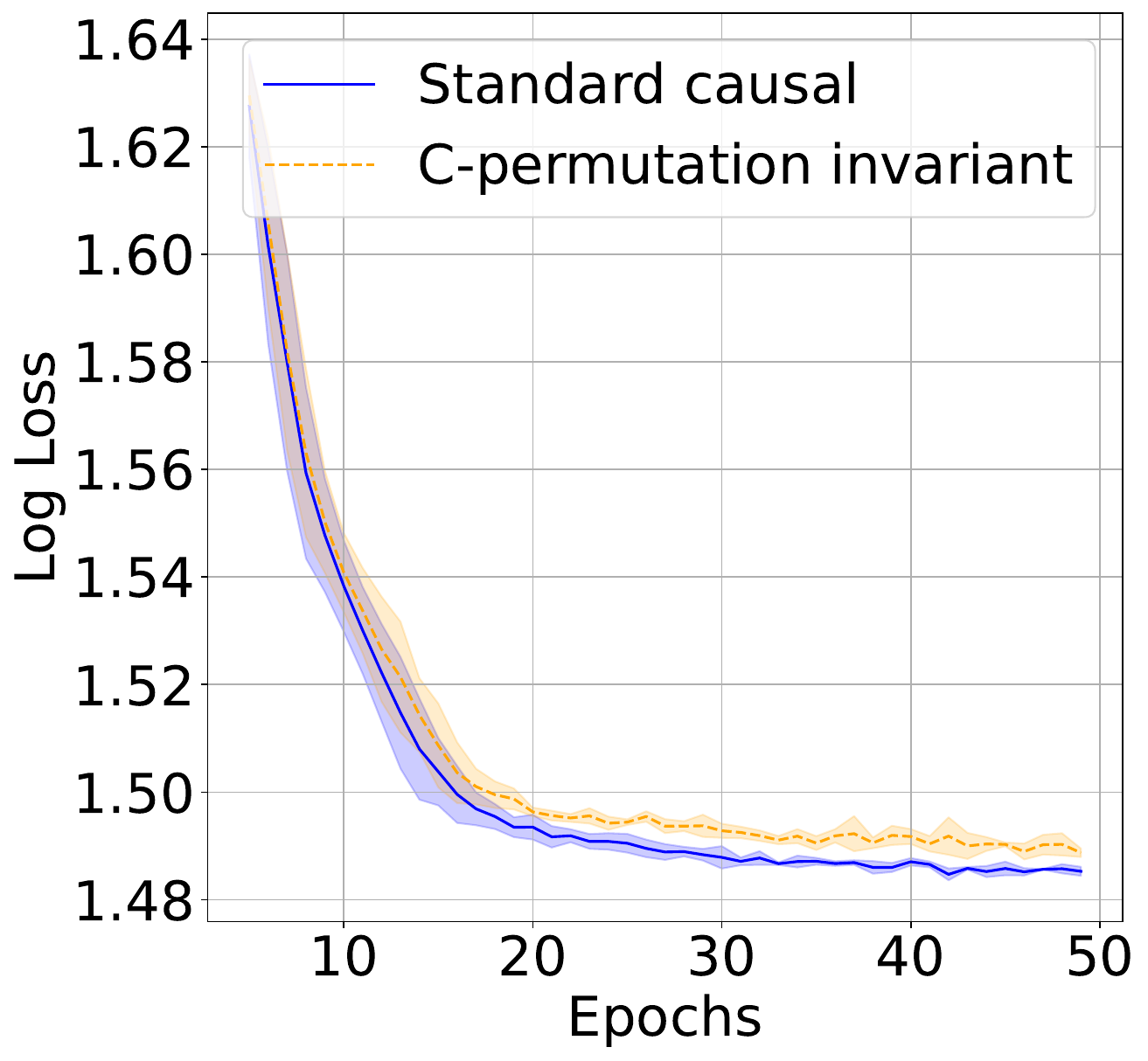}
  \caption*{\textbf{(e)} Noise: 1.0}
\end{minipage}
\hfill
\begin{minipage}[b]{0.32\textwidth}
\centering \includegraphics[width=0.9\textwidth,height=4cm] {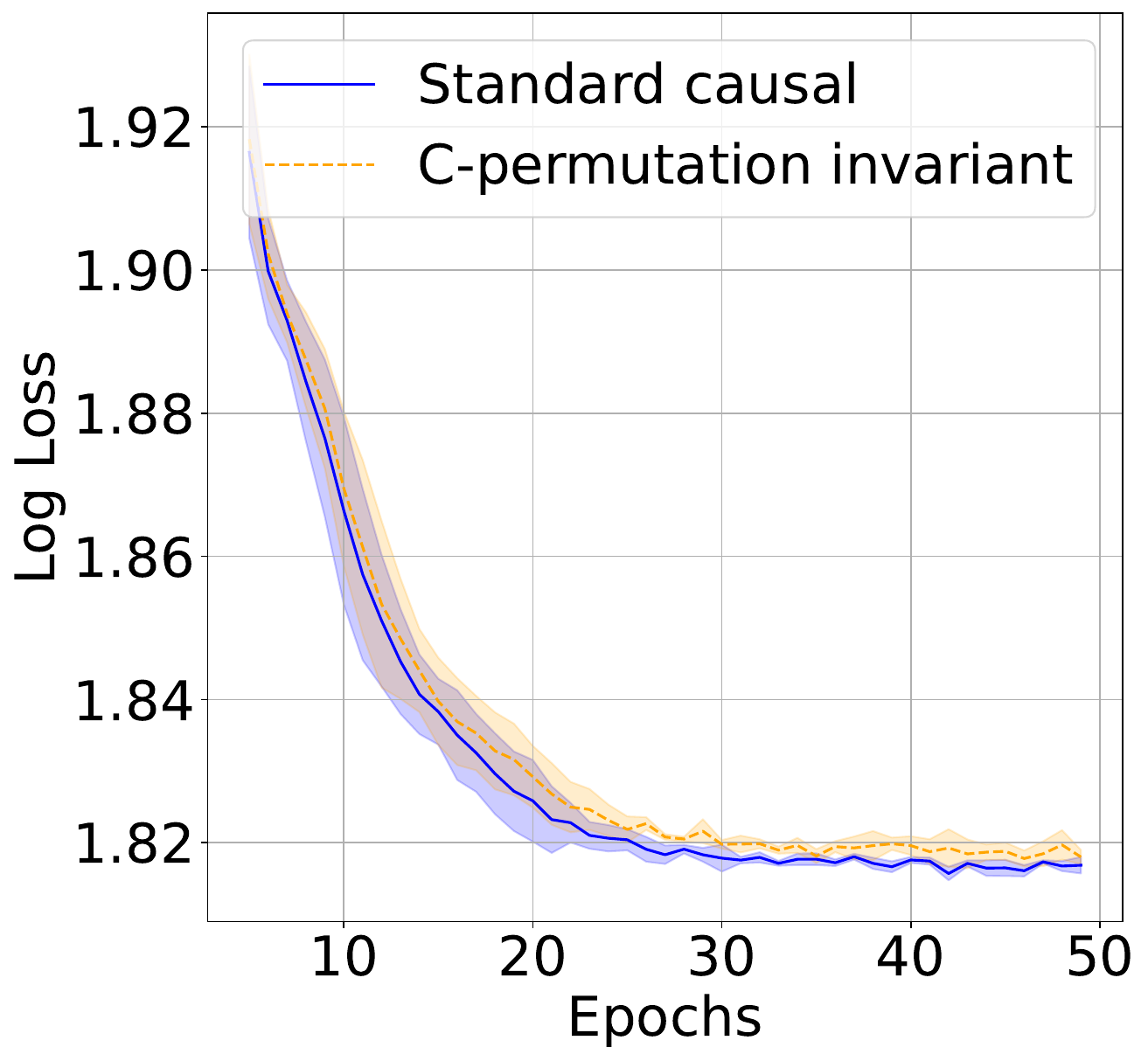}
  \caption*{\textbf{(f)} Noise: 2.0}
\end{minipage}
 \caption{\textbf{Ablation on noise (Training/Data efficiency):} Comparing two architectures [Training horizon: 100, Metric: Multi-step log-loss, Target length: 10].}
 \label{fig:ablation_noise_data_efficiency}
\end{figure*}

\noindent\textit{Training/Data efficiency:}
Figures \ref{fig:ablation_noise_data_efficiency} shows the training and data efficiency across different levels of observation noise. The results align with our earlier findings.

\begin{figure*}[t]
\centering
\begin{minipage}[b]{0.32\textwidth}
\centering \includegraphics[width=0.9\textwidth,height=4cm]
{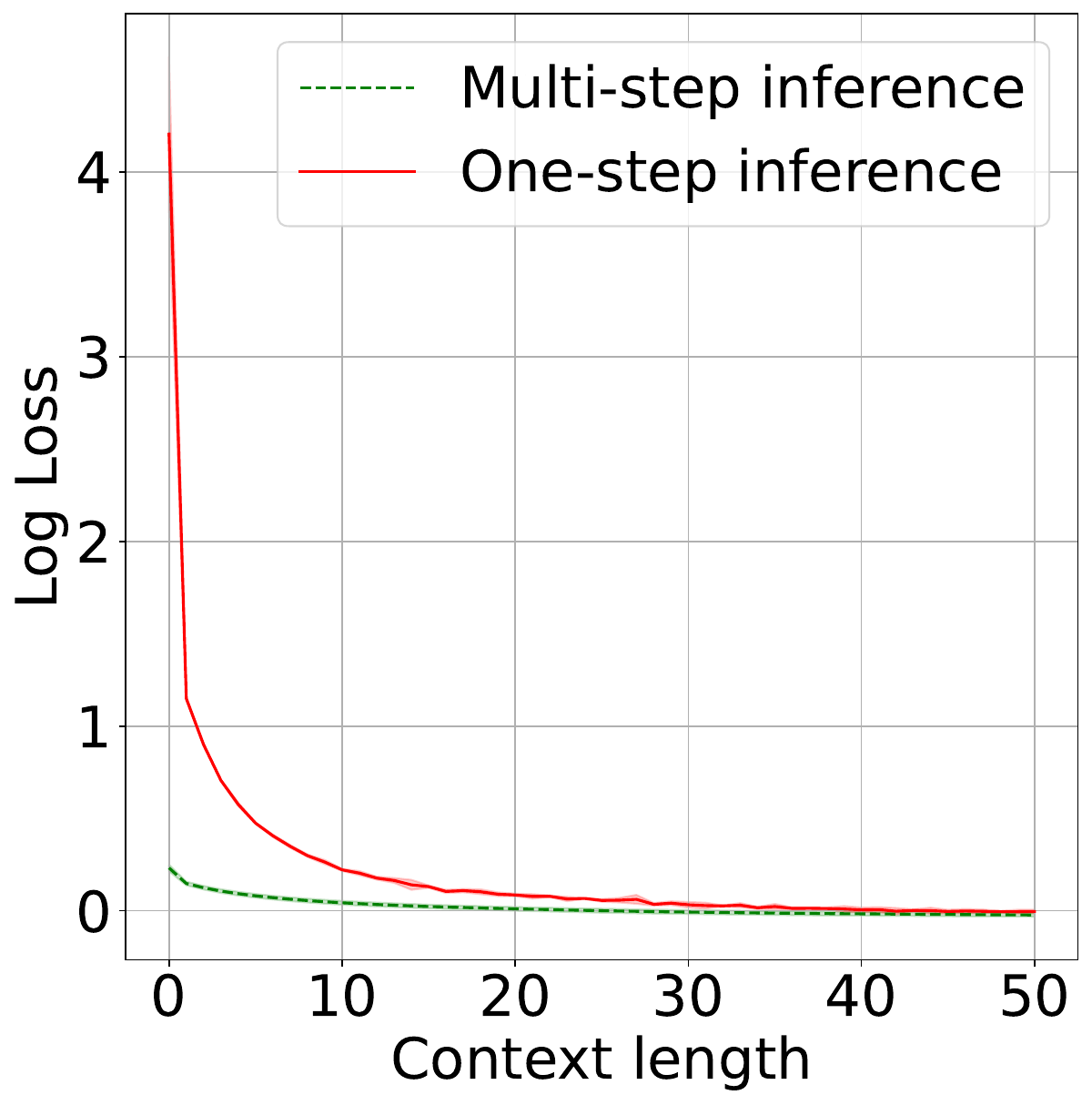}
  \caption*{\textbf{(a)} Noise: 0.05}
\end{minipage}
\hfill
\begin{minipage}[b]{0.32\textwidth}
\centering \includegraphics[width=0.9\textwidth,height=4cm] {figures/figures_ablation_error_bars/joint_x_axis_context_multi_one_step_Horizon_100_epoch_400_dim_1_target_len_50_noise_0.1.pdf}
  \caption*{\textbf{(b)} Noise: 0.1}
\end{minipage}
\hfill
\begin{minipage}[b]{0.32\textwidth}
\centering \includegraphics[width=0.9\textwidth,height=4cm] {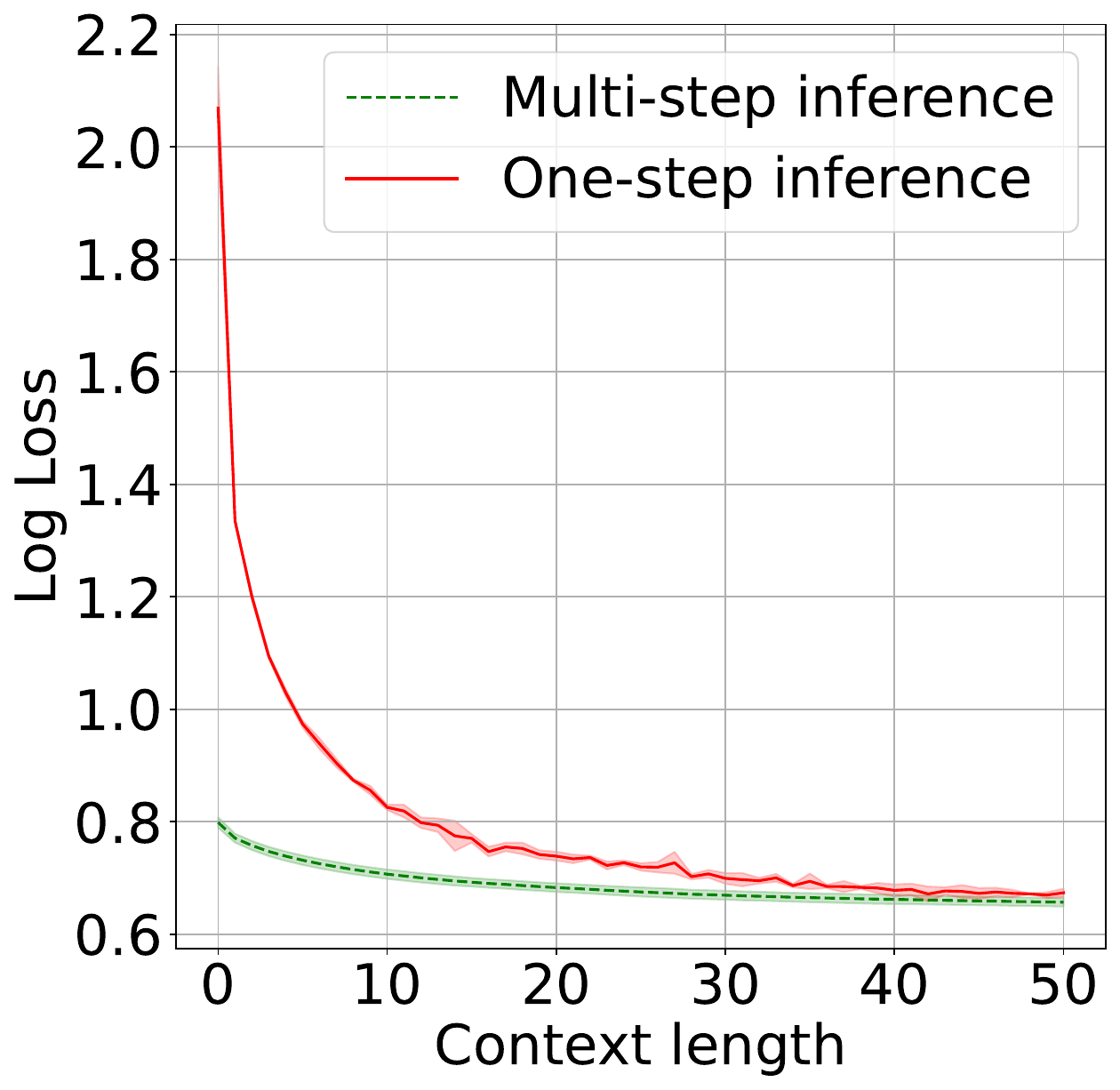}
  \caption*{\textbf{(c)} Noise: 0.2}
\end{minipage}
\hfill
\vspace{10mm}

\begin{minipage}[b]{0.32\textwidth}
\centering \includegraphics[width=0.9\textwidth,height=4cm]
{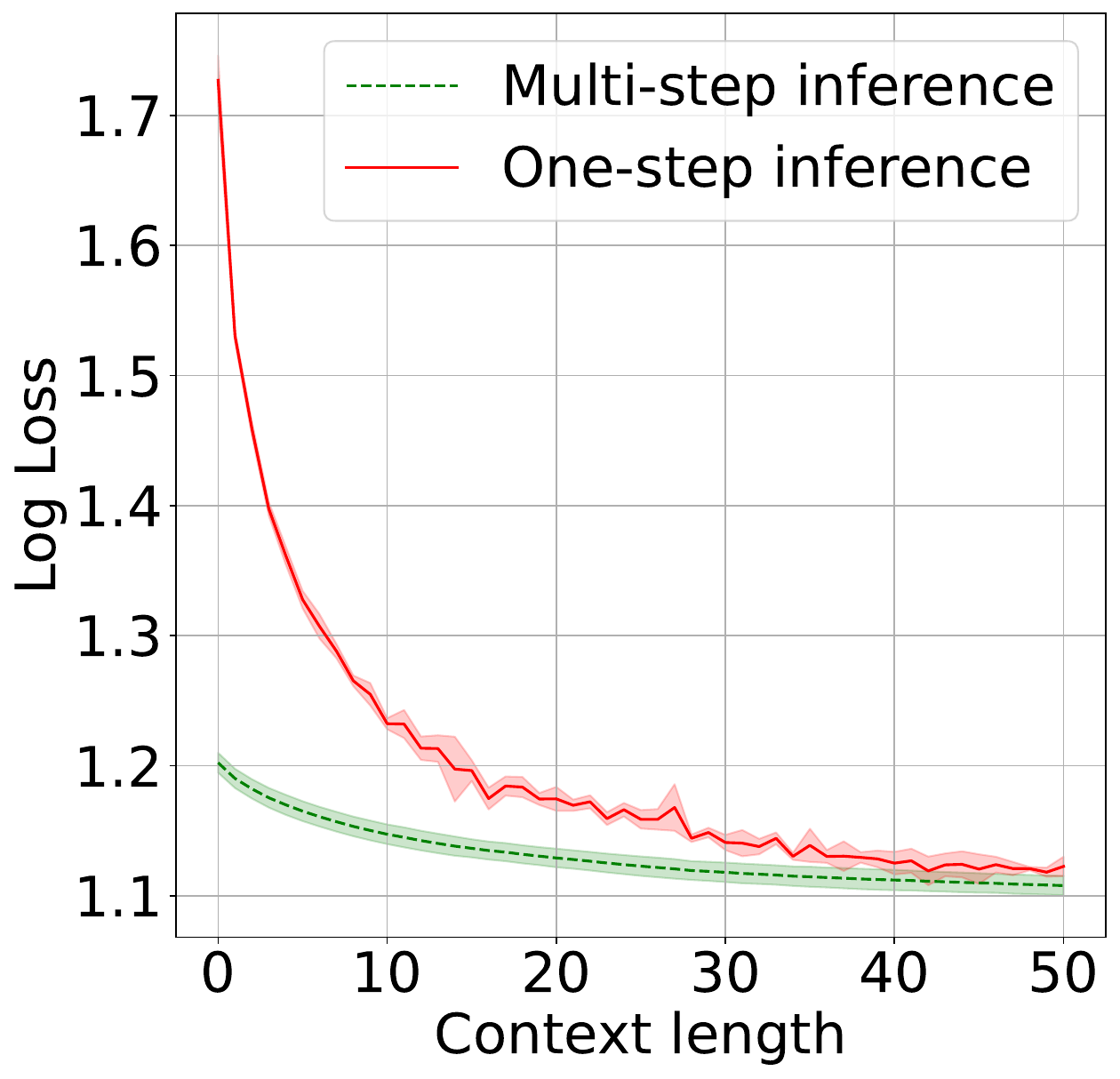}
  \caption*{\textbf{(d)} Noise: 0.5}
\end{minipage}
\hfill
\begin{minipage}[b]{0.32\textwidth}
\centering \includegraphics[width=0.9\textwidth,height=4cm] {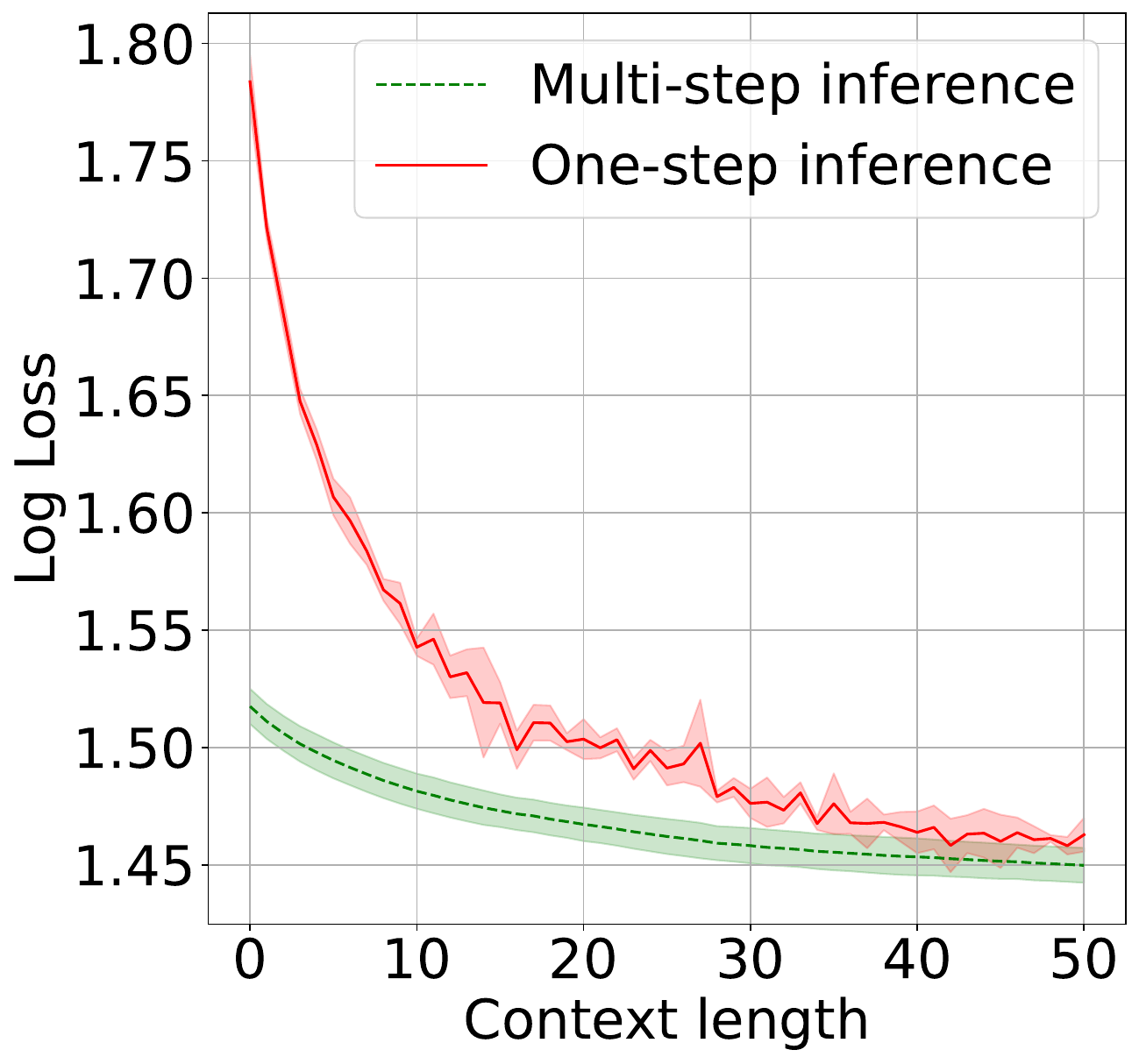}
  \caption*{\textbf{(e)} Noise: 1.0}
\end{minipage}
\hfill
\begin{minipage}[b]{0.32\textwidth}
\centering \includegraphics[width=0.9\textwidth,height=4cm] {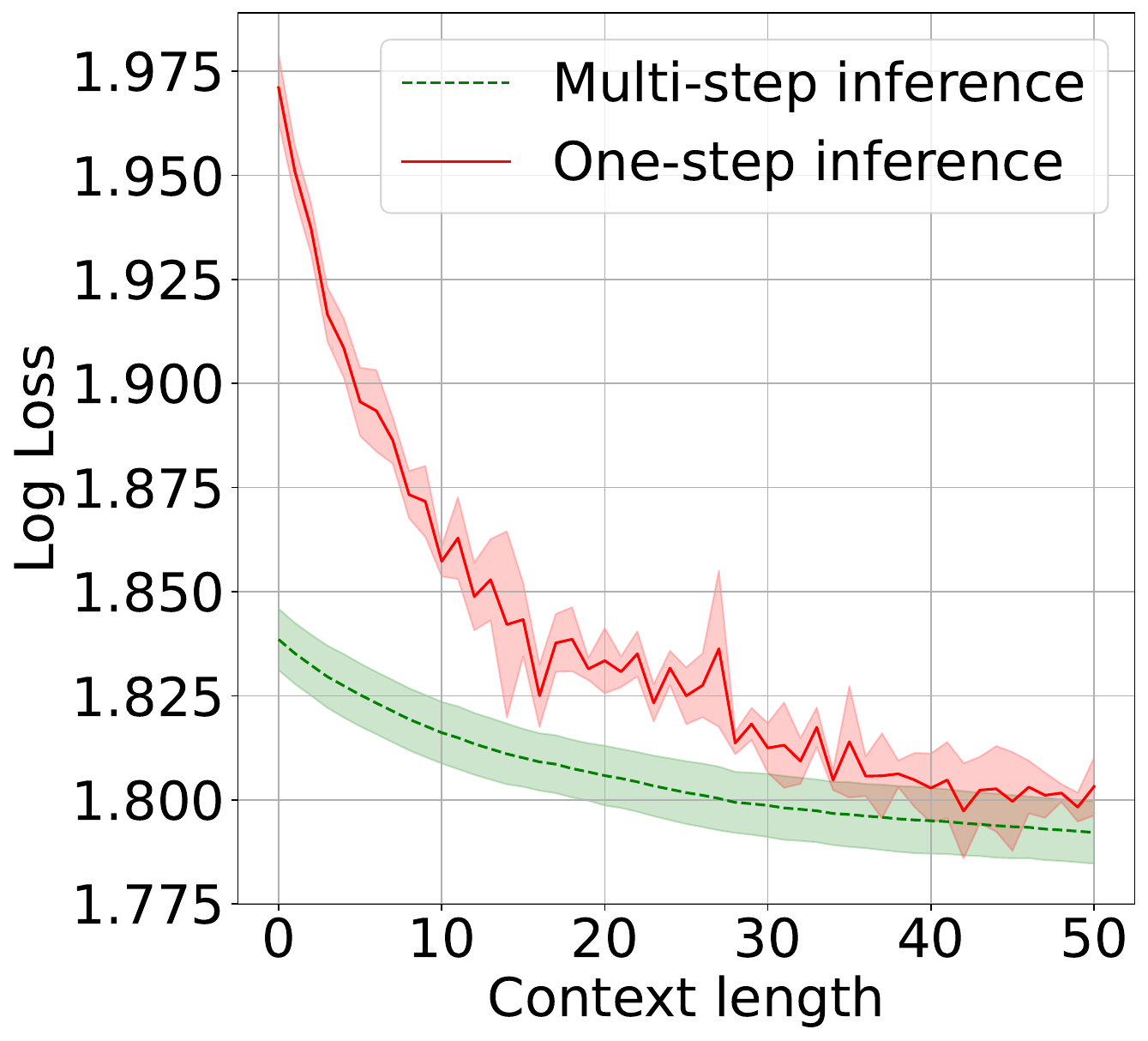}
  \caption*{\textbf{(f)} Noise: 2.0}
\end{minipage}
\caption{\textbf{Ablation on noise (Uncertainty Quantification):} Comparing One-step inference and multi-step inference [Train horizon: 100, Metric: Multi-step log-loss, Target Length: $50$].}
\label{fig:ablation_noise_uq_one_multi}
\end{figure*}

\noindent\textit{Multi-step v/s One-step inference:}
Figures \ref{fig:ablation_noise_uq_one_multi} present a comparison of multi-step and one-step inference under different observation noise levels.

\subsection{Downstream performance of the two architectures}

\begin{figure}
    \centering \includegraphics[width=0.3\textwidth, height=4cm]{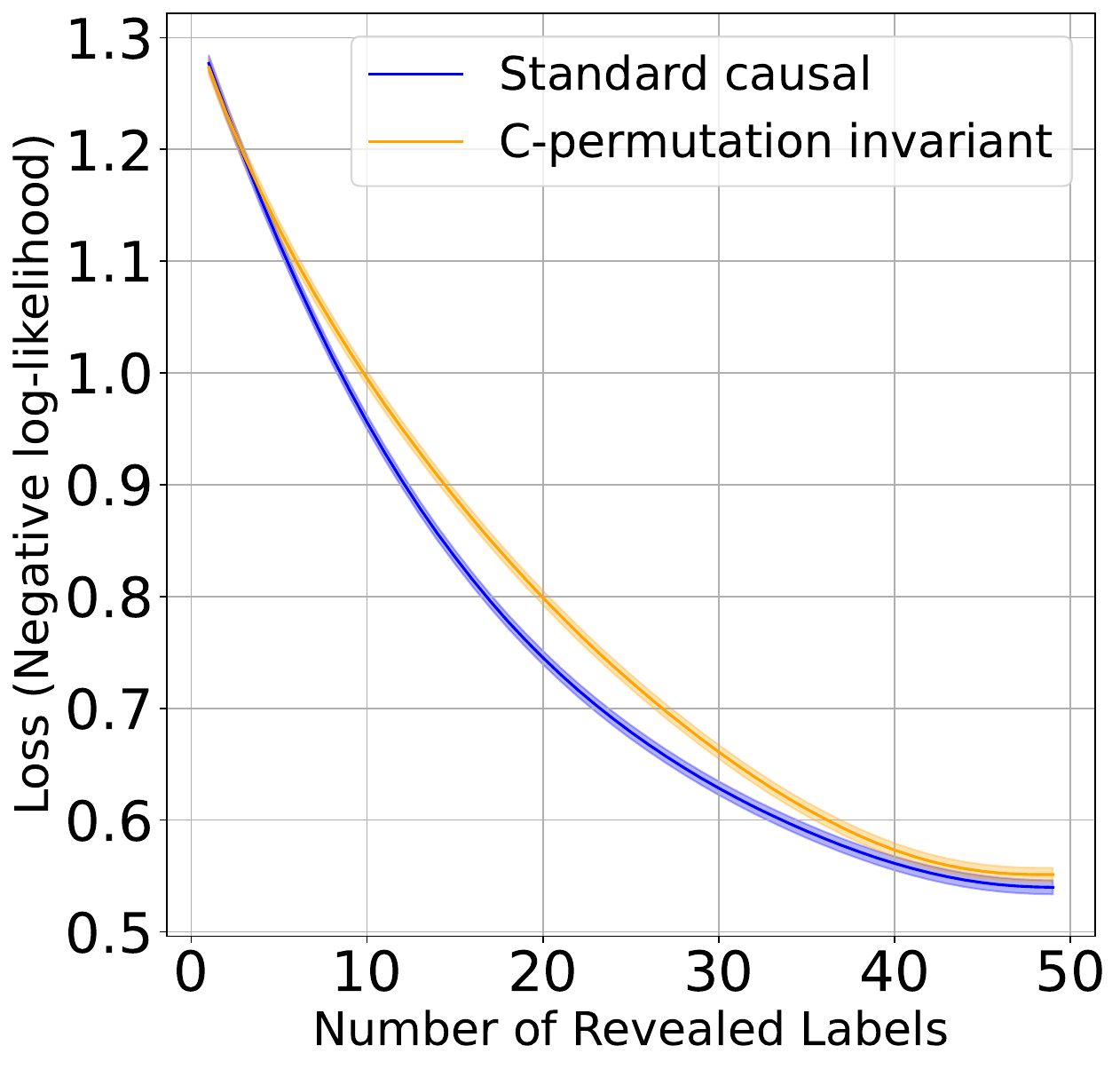}
  \caption{\textbf{Active Learning:} Comparing C-permutation invariant and  standard causal architecture for the active learning setting.}
\label{fig:comparing_autoreg_exchg_active_learning}
\end{figure}

Figure \ref{fig:comparing_autoreg_exchg_active_learning} compares the performance of the conditionally permutation-invariant architecture and the standard causal architecture in an active learning setting. Our results indicate that the standard causal architecture outperforms the conditionally permutation-invariant architecture.

\end{document}